\documentclass[10pt,journal,compsoc]{IEEEtran}

\ifCLASSOPTIONcompsoc
  \usepackage[nocompress]{cite}
\else
  \usepackage{cite}
\fi
\ifCLASSINFOpdf
\else
\fi
\usepackage{times}
\usepackage{epsfig}
\usepackage{graphicx}
\usepackage{amsmath}
\usepackage{amssymb}
\usepackage{enumitem}
\usepackage{multirow}
\usepackage{xcolor}
\usepackage{adjustbox}
\newcommand{\widthscale}{0.135} 
\usepackage[toc,page]{appendix}
\usepackage{caption}
\usepackage{comment}
\usepackage{subcaption}
\usepackage{cite}
\usepackage[colorlinks]{hyperref}

\usepackage{xspace}

\makeatletter
\DeclareRobustCommand\onedot{\futurelet\@let@token\@onedot}
\def\@onedot{\ifx\@let@token.\else.\null\fi\xspace}

\def\eg{\emph{e.g}\onedot}

\def\etc{\emph{etc}\onedot} 
 
\def\etal{\emph{et al}\onedot}
\makeatother

\usepackage{url}
\begin{document}
%
\title{Zooming SlowMo: An Efficient One-Stage Framework for Space-Time Video Super-Resolution}
%
%
%
%

\author{Xiaoyu Xiang$^{\ast}$,~\IEEEmembership{Member,~IEEE,}
        Yapeng Tian$^{\ast}$,~\IEEEmembership{Member,~IEEE,}
        Yulun Zhang$^{\ast}$,~\IEEEmembership{Member,~IEEE,}\\
        Yun Fu,~\IEEEmembership{Fellow,~IEEE,}
        Jan P. Allebach$^{\dagger}$,~\IEEEmembership{Life Fellow,~IEEE,}
        and~Chenliang~Xu$^{\dagger}$,~\IEEEmembership{Member,~IEEE}
\IEEEcompsocitemizethanks{\IEEEcompsocthanksitem $^{*}$Equal contribution; $^{\dagger}$\text{Equal advising.}

\IEEEcompsocthanksitem X. Xiang and J. Allebach are with Department of ECE, Purdue University, West Lafayette,
IN 47907. E-mail: \{xiang43, allebach\}@purdue.edu
\IEEEcompsocthanksitem Y. Tian and C. Xu are with Department of Computer Science, University of Rochester,
Rochester, NY 14627. E-Mail: \{yapengtian, chenliang.xu\}@rochester.edu
\IEEEcompsocthanksitem Y. Zhang is with Department of ECE, Northeastern University, Boston, MA 02115. E-mail: yulun100@gmail.com
\IEEEcompsocthanksitem Y. Fu is with Department of ECE and Khoury College of Computer Science, Northeastern University, Boston, MA 02115. E-Mail: yunfu@ece.neu.edu

}
}

%
%

\markboth{Journal of \LaTeX\ Class Files,~Vol.~14, No.~8, August~2015}%
{Shell \MakeLowercase{\textit{et al.}}: Bare Advanced Demo of IEEEtran.cls for IEEE Computer Society Journals}
%



\IEEEtitleabstractindextext{%
\begin{abstract}
In this paper,{\let\thefootnote\relax\footnote{{$^{*}$Equal contribution; $^{\dagger}$\text{Equal advising}.}}} we address the space-time video super-resolution, which aims at generating a high-resolution (HR) slow-motion video from a low-resolution (LR) and low frame rate (LFR) video sequence. A naive method is to decompose it into two sub-tasks: video frame interpolation (VFI) and video super-resolution (VSR). Nevertheless, temporal interpolation and spatial upscaling are intra-related in this problem. Two-stage approaches cannot fully make use of this natural property. Besides, state-of-the-art VFI or VSR deep networks usually have a large frame reconstruction module in order to obtain high-quality photo-realistic video frames, which makes the two-stage approaches have large models and thus be relatively time-consuming. To overcome the issues, we present a one-stage space-time video super-resolution framework, which can directly reconstruct an HR slow-motion video sequence from an input LR and LFR video. Instead of reconstructing missing LR intermediate frames as VFI models do, we temporally interpolate LR frame features of the missing LR frames capturing local temporal contexts by a feature temporal interpolation module. Extensive experiments on widely used benchmarks demonstrate that the proposed framework not only achieves better qualitative and quantitative performance on both clean and noisy LR frames but also is several times faster than recent state-of-the-art two-stage networks. \textit{The source code is released in \url{https://github.com/Mukosame/Zooming-Slow-Mo-CVPR-2020}.}
\end{abstract}

}

\maketitle

\IEEEdisplaynontitleabstractindextext

%
\IEEEpeerreviewmaketitle

\ifCLASSOPTIONcompsoc
\IEEEraisesectionheading{\section{Introduction}\label{sec:introduction}}
\else
\section{Introduction}
\label{sec:introduction}
\fi

%
%
%
%

\begin{figure*}
    \centering
    \includegraphics[width=\textwidth]{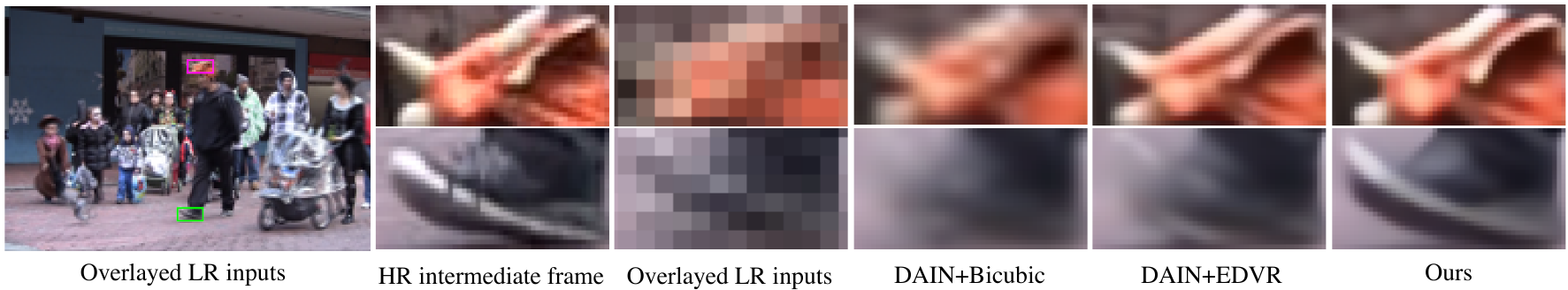}
    \caption{\textbf{Example of space-time video super-resolution.} We propose a one-stage STVSR network to directly predict high frame rate (HFR) and high-resolution (HR) frames from the corresponding low-resolution (LR) and low frame rate (LFR) frames without explicitly interpolating intermediate LR frames. A HR intermediate frame $t$ and its neighboring LR frames: $t-1$ and $t+1$ as an overlayed image are shown.  Comparing with a SOTA two-stage method: DAIN~\cite{bao2019depth}$+$EDVR~\cite{wang2019edvr} on the HR intermediate frame $t$, our model is more effective in handling video motions and therefore restores more accurate visual structures and sharper edges. Furthermore, our network is more than $3$ times faster on inference speed with a $4$ times smaller model size than the DAIN$+$EDVR.}
    \label{fig:examples}
\vspace{-4mm}
\end{figure*}
\IEEEPARstart{S}{pace-Time} Video Super-Resolution (STVSR)~\cite{shechtman2005space} aims to automatically generate a photo-realistic video sequence with high-resolution (HR) and high frame rate (HFR) from a low space-time resolution input video. Containing clear motion dynamics and fine image details, HR slow-motion videos are more visually appealing. They are desired in various applications, such as character animation, film making, and high-definition television. However, the reconstruction of such kind of videos is an ill-posed problem.



To address this problem, most existing research works~\cite{shechtman2005space,mudenagudi2010space,takeda2010spatiotemporal,shahar2011space,faramarzi2012space,li2015space} often introduce some prior knowledge, make strong assumptions, and adopt hand-crafted regularization. For instance, Shechtman \textit{et al.}~\cite{shechtman2005space} adopted space-time directional smoothness prior. An assumption made in~\cite{mudenagudi2010space} shows that the illumination change for the static pixels is not significant. However, these strong constraints hinder the capacity of the methods to model diverse and various space-time visual patterns. Moreover, it is computationally expensive (\eg, $\sim$ $60$ frames per hour in~\cite{mudenagudi2010space}) for these methods in terms of optimization.

Recently, deep convolutional neural networks (CNNs) have shown excellent performance in various video restoration applications, such as video deblurring~\cite{su2017deep}, video frame interpolation (VFI)~\cite{niklaus2017adavonv}, video super-resolution (VSR)~\cite{caballero2017real}, and even space-time video super-resolution (STVSR). To design a network for such an STVSR task, one straightforward way is to sequentially combine a VFI method (\eg, SepConv~\cite{niklaus2017adsconv}, ToFlow~\cite{xue2019video}, DAIN~\cite{bao2019depth} \etc) and a VSR method (\eg, DUF~\cite{jo2018deep}, RBPN~\cite{haris2019recurrent}, EDVR~\cite{wang2019edvr} \etc) in a two-stage manner. Specifically, it firstly interpolates missing intermediate low-resolution (LR) video frames with VFI and then super-resolves all LR frames to the desired resolution with VSR. However, it is not independent but intra-related for temporal interpolation and spatial super-resolution in STVSR. By splitting them into two individual procedures, the two-stage methods fail to make better usage of this natural property. Moreover, to predict high-quality video frames, both state-of-the-art (SOTA) VFI and VSR networks require a large frame reconstruction module. Consequently, the composed two-stage STVSR network will inevitably contain a huge number of parameters, resulting in high computational complexity. 

To alleviate the above issues, we propose Zooming SlowMo (ZSM), a unified one-stage STVSR framework, to simultaneously conduct temporal interpolation and spatial super-resolution. Specifically, instead of synthesizing pixel-wise LR frames as in two-stage methods, we propose to adaptively learn a deformable feature interpolation function to temporally interpolate intermediate LR frame features. The interpolation function contains learnable offsets, which can not only aggregate useful local temporal contexts but also help the temporal interpolation handle complex visual motions. Furthermore, we introduce a new deformable ConvLSTM model to effectively leverage global contexts with simultaneous temporal alignment and aggregation. We then reconstruct the HR video frames from the aggregated LR features with a deep SR network. To this end, the one-stage network can learn end-to-end to map an LR, LFR video sequence to its HR, HFR counterpart in a sequence-to-sequence manner. Moreover, we propose a cyclic interpolation loss to guide the frame feature interpolation and further improve STVSR performance. Extensive experiments show that our proposed one-stage STVSR framework outperforms SOTA two-stage methods even with much fewer parameters and is capable of handling multiple degradations. We illustrate an STVSR example in Figure~\ref{fig:examples}. 



The main contributions of this work are three-fold:
\begin{itemize}
    \item We propose Zooming SlowMo (ZSM), a one-stage STVSR network, to conduct temporal interpolation and spatial SR simultaneously. Our ZSM is more effective than two-stage methods by taking advantage of the intra-relatedness between the two sub-problems. It is also more computationally efficient because it only requires one frame reconstruction module rather than two large networks as in two-stage methods. 
    \item We propose a frame feature temporal interpolation network, which leverages local temporal contexts based on deformable sampling for intermediate LR frames. We further devise a novel deformable ConvLSTM to explicitly enhance temporal alignment capacity and exploit global temporal contexts to handle large motions in videos. 
    \item Our one-stage ZSM method achieves SOTA performance on commonly used benchmarks (i.e., Vid4 \cite{liu2011bayesian} and Vimeo~\cite{xue2019video}). It is 3$\times$ faster than the two-stage network: DAIN~\cite{bao2019depth} + EDVR~\cite{wang2019edvr} and has a nearly 4$\times$ reduction in model size. 
\end{itemize}
A preliminary version of this manuscript has been presented in~\cite{xiang2020zooming}. In the current work, we incorporate additional contents to improve our method and address the space-time video super-resolution in more challenging conditions:
\begin{itemize}
    \item We improve model performance via integrating the guided feature interpolation learning into our one-stage framework.
    \item We investigate space-time video super-resolution under even noisy conditions, in which random noises or JPEG compression artifacts corrupt input LR video frames. Such applications allow us to explore the flexibility and potential breath of our Zooming SlowMo (ZSM) method.
    \item Additional and extensive experimental results can demonstrate the effectiveness of the proposed guided interpolation learning, and further show the superiority of our one-stage network on tackling more challenging noisy STVSR tasks. 
\end{itemize}

\begin{figure*}
    \centering
    \includegraphics[width=\textwidth]{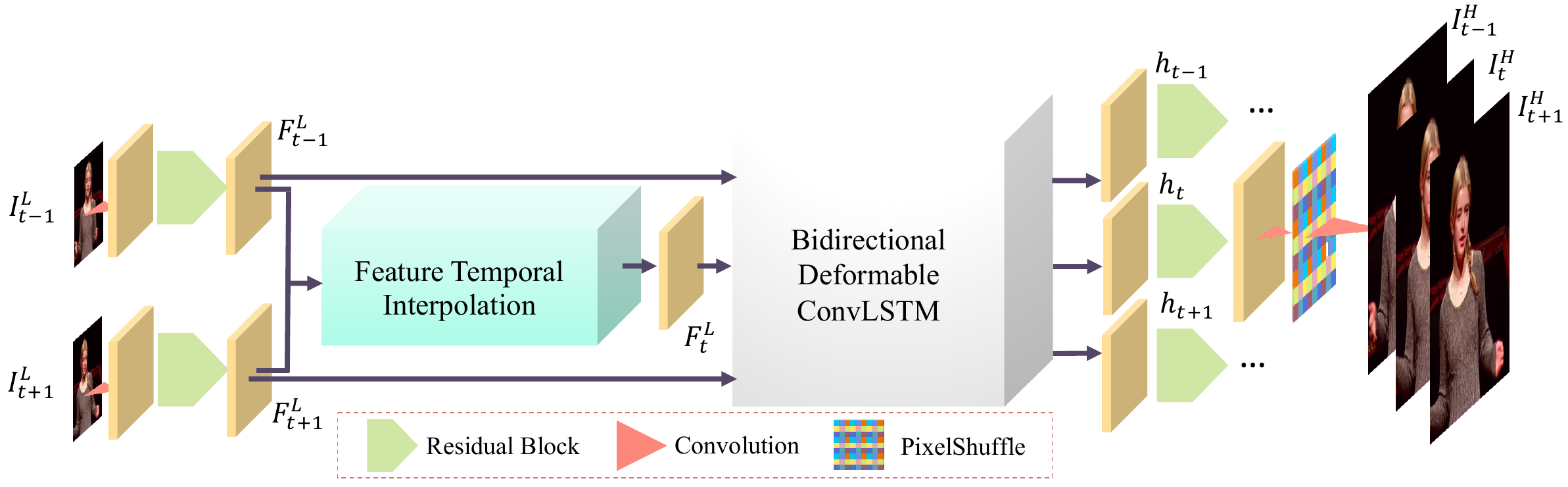}
    \caption{Overview of our proposed one-stage STVSR framework ZSM. Without synthesizing intermediate LR frames $I_t^L$, our ZSM directly predicts temporally consecutive HR frames. We utilized frame feature temporal interpolation and deformable ConvLSTM to leverage local and global temporal contexts, which contribute to exploit temporal information and handle large motions. It should be noted that we only show two LR input frames from a long video sequence in the figure for better understanding.}
    \label{fig:structure}
    \vspace{-3mm}
\end{figure*}

\vspace{-2mm}
\section{Related Work}
In this section, we discuss works on related topics: video frame interpolation (VFI), video super-resolution (VSR), and space-time video super-resolution (STVSR), and image degradation.

\subsection{Video Frame Interpolation}
Video frame interpolation (VFI) aims to synthesize non-existent intermediate frames between the original frames. 
Meyer~\etal~\cite{meyer2015phase} introduced a phase-based frame interpolation method, which utilizes per-pixel phase modification to generate intermediate frames. Long \etal \cite{long2016learning} applied an encoder-decoder CNN to predict intermediate frames directly. Regarding the frame interpolation as a local convolution over the two input frames, Niklaus~\etal~\cite{niklaus2017adavonv,niklaus2017adsconv} utilized a CNN to learn a spatially-adaptive convolution kernel for each pixel. Choi~\etal~\cite{choi2020channel} introduced a feature reshaping operation with Pixelshuffle \cite{shi2016real} and a channel attention module for motion estimation. Lee~\etal~\cite{lee2020adacof} proposed adaptive collaboration of flows as a new warping module to deal with complex motions.
To handle motions more explicitly, many flow-based video interpolation approaches~\cite{jiang2018super,liu2017video,niklaus2018context,bao2019memc,bao2019depth,niklaus2020softmax,park2020bmbc} have also been proposed. However, these methods usually have inherent issues with inaccuracies and missing information from optical flow results. 
Cycle consistency constraints were utilized in the time domain to supervise the intermediate frame generation~\cite{reda2019unsupervised,liu2019deep,shen2020blurry}.
Recently, \cite{pan2019bringing,pollak2020across,shen2020blurry,linlearning} addressed the motion blur and motion aliasing of complex scenes in temporal frame interpolation. Choi~\etal~\cite{choi2020scene} proposed a novel scene adaptation framework to further improve the frame interpolation models via meta-learning.
Instead of synthesizing the intermediate LR frames as current VFI methods do, our one-stage framework interpolates features from two neighboring LR frames to directly synthesize LR feature maps for missing frames without explicit supervision.

\subsection{Video Super-Resolution}
Video super-resolution (VSR) aims to reconstruct an HR video frame from the corresponding LR frame (reference frame) and its neighboring LR frames (supporting frames). How to temporally align the LR supporting frames with the reference frame is a key problem for VSR. Various VSR approaches~\cite{caballero2017real,tao2017detail,sajjadi2018frame,wang2018learning,xue2019video} used optical flow for explicit temporal alignment. Such a temporal alignment first estimates motions between the reference frame and each supporting frame with optical flow and then warps the supporting frame using the predicted motion map. In recent years, RBPN proposed to incorporate the single image and multi-frame SR methods for VSR, where LR video frames are directly concatenated with flow maps. However, there exist some problems. For example, it is difficult to obtain accurate flow. Flow warping also introduces artifacts into the aligned frames. 

To alleviate this problem, DUF~\cite{jo2018deep} with dynamic filters and TDAN~\cite{tian2018tdan} with deformable alignment was proposed for implicit temporal alignment without motion estimation. Wang~\etal proposed EDVR~\cite{wang2019edvr} by extending the deformable alignment in TDAN with multiscale information. MuCAN~\cite{li2020mucan} applies a temporal multi-correspondence aggregation method to utilize similar patches in both intra- and inter-frames. However, most of those methods are many-to-one architectures, where a batch of LR frames are needed to predict only one HR frame. Such a process makes the methods computationally inefficient. On the other hand, recurrent neural networks (RNNs), such as convolutional LSTMs~\cite{xingjian2015convolutional} (ConvLSTM), can ease sequence-to-sequence (S2S) learning. VSR methods~\cite{lim2017deep,huang2017video,isobe2020video} widely adopted them to leverage temporal information. However, when processing videos with large and complex motions, the RNN-based VSR networks would suffer from obvious performance drop for the lack of explicit temporal alignment. 

To achieve efficient yet effective modeling, unlike existing methods, we propose a novel ConvLSTM structure ZSM embedded with an explicit state updating cell for STVSR.
Instead of simply combining VFI and VSR networks to solve the STVSR problem, we propose a more efficient and effective one-stage framework ZSM. Our ZSM simultaneously learns the temporal feature interpolation and the spatial super-resolution without synthesizing LR intermediate frames during inference.

\subsection{Space-Time Video Super-Resolution}
Space-time video super-resolution (STVSR), as a more difficult video restoration task, aims to super-resolve the space-time resolution according to the LR and LFR video. Shechtman \etal~\cite{shechtman2002increasing} firstly proposed to extend SR to the space-time domain. STVSR is a highly ill-posed inverse problem, because pixels are missing in LR frames, and even several entire LR frames are unavailable. To increase video resolution both in time and space, Shechtman \etal~\cite{shechtman2002increasing} combined information from multiple video sequences of dynamic scenes, which can be obtained at sub-pixel and sub-frame misalignments with a directional space-time smoothness regularization to constrain the ill-posed problem. Mudenagudi~\cite{mudenagudi2010space} posed STVSR as a reconstruction problem, which utilized the Maximum a posteriori-Markov Random Field~\cite{geman1984stochastic} and solved the optimization problem via graph-cuts~\cite{boykov2001fast}. Local orientation and local motion were exploited in~\cite{takeda2010spatiotemporal} to steer spatio-temporal regression kernels. Shahar~\etal~\cite{shahar2011space} proposed to exploit a space-time patch recurrence prior within natural videos for STVSR. However, these methods have limited capacity to model rich and complex space-time visual patterns. Plus, the optimization in these methods is usually computationally expensive.

Very recently, \cite{kim2020fisr} proposed a multi-scale temporal loss. \cite{haris2020space} concatenated LR images and pre-computed optical flow for intermediate feature estimation and refinement. \cite{kang2020upsample} used optical flow to explicitly warp the features. \cite{xiao2020space} introduced a news perspective to exploit the STVSR problem from the temporal profile. Unlike previous methods, we propose a one-stage framework ZSM to directly learn the mapping between partial LR observations and HR video frames, achieving accurate and fast STVSR. 
\subsection{Image Degradation}

Besides downsampling, captured images might also be degraded by other corruptions, such as noise, blur, and compression artifacts. To restore high-quality images from the corresponding degraded inputs, many image restoration methods have been developed. A pioneer CNN-based image restoration method is from~\cite{jain2009natural}, in which Jain and Seung firstly propose a CNN architecture with stacked convolutional layers for natural image denoising. Dong~\etal introduce shallow plain CNN networks to address image super-resolution in~\cite{dong2014learning} and handle JPEG compression artifacts~\cite{dong2015compression}. Furthermore, more advanced and deeper models are proposed in ~\cite{wang2016d3,guo2016building,zhang2017beyond,mao2016image,tai2017memnet,zhang2017learning,zhang2018ffdnet,zhang2019residual}. 

To further demonstrate the robustness of the proposed approach to different image corruptions, we explore STVSR under noisy conditions. We introduce random noise and JPEG compression artifacts into LR inputs, respectively, such that our STVSR needs to hallucinate missing HR frames from noisy LR frames.

\vspace{2mm}
\section{Space-Time Video Super-Resolution}
\label{STVSR}

In this section, we first give an overview of the proposed space-time video super-resolution framework in Sec.~\ref{sec:overview}. Built upon this framework, we then propose a novel frame feature temporal interpolation network in Sec.~\ref{dcn}; deformable ConvLSTM in Sec.~\ref{dconvlstm}; frame reconstruction module in Sec.~\ref{sec:framerec}; guided feature interpolation learning in Sec.~\ref{sec:cyclic}. Finally, we provide details about our implementation in Sec.~\ref{sec:implement}.

\subsection{Overview}
\label{sec:overview}

Given an LFR and LR video sequence: $\mathcal{I}^{L}=\{I_{2t-1}^L\}_{t=1}^{n+1}$, we aim to reconstruct the corresponding slow-motion high-resolution video sequence: $\mathcal{I}^{H}=\{I_{t}^H\}_{t=1}^{2n+1}$. For intermediate HR video frames $\{I_{2t}^H\}_{t=1}^{n}$, there are no LR counterparts in the input video sequence. To efficiently and effectively increase video resolution in both time and space domains, we present a one-stage space-time video super-resolution framework: Zooming SlowMo (ZSM) as shown in Figure~\ref{fig:structure}. The framework mainly includes four parts: \textit{feature extractor}, \textit{frame feature temporal interpolation module}, \textit{deformable ConvLSTM network}, \textit{HR frame 
reconstruction module}, and \textit{guided feature interpolation learning module}.

We first utilize a feature extractor with one convolutional layer and $k_1$ residual units to extract visual feature maps: $\{F_{2t-1}^L\}_{t=1}^{n+1}$ from each input video frame. Taking the extracted feature maps as input, we then synthesize feature maps: $\{F_{2t}^L\}_{t=1}^{n}$ of intermediate LR frames with a feature temporal interpolation module. In addition, to better exploit temporal information, we propose a deformable ConvLSTM to process the temporally consecutive feature maps: $\{F_{t}^L\}_{t=1}^{2n+1}$. Different from vanilla ConvLSTM, our deformable ConvLSTM can simultaneously employ temporal alignment and aggregation. Finally, we restore the HR slow-mo video frames from the aggregated feature maps. 
Since the features for reconstructing intermediate frames are synthesized, there will be feature synthesis errors that will be propagated into restored HR frames, making the predicted video sequence suffer from jitters. To alleviate this problem, we further propose a guided feature interpolation learning mechanism. 


\begin{figure}
    \centering
    \includegraphics[width=1.0\columnwidth]{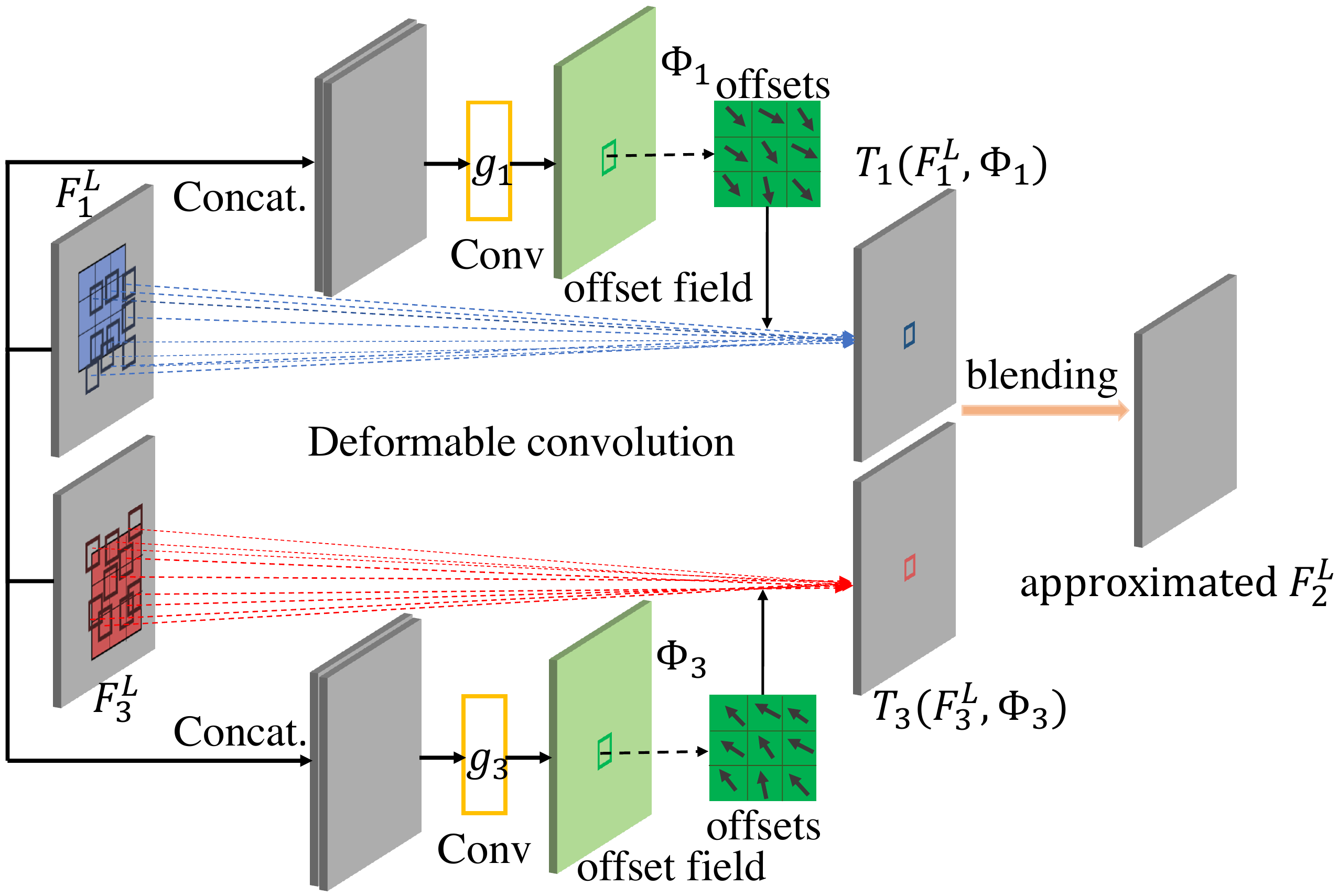}
    \caption{Frame feature temporal interpolation based on deformable sampling. Since approximated $F_2^L$ is utilized to reconstruct the corresponding HR video frame, it will implicitly enforce learnable convolution offsets to capture temporally dynamic local contexts and be motion-aware.}
    \label{fig:interp_module}
    \vspace{-3mm}
\end{figure}

\subsection{Frame Feature Temporal Interpolation}
\label{dcn}

Given extracted feature maps: $F_{1}^L$ and $F_{3}^L$ from input LR frames: $I_{1}^L$ and $I_{3}^L$, we aim to synthesize the feature map $F_{2}^L$ corresponding to the unavailable intermediate LR video frame $I_{2}^L$. Previous video frame interpolation models usually employ temporal interpolation on pixel-wise frames, which leads to a two-stage space-time super-resolution design. Different from previous approaches, we propose to learn a feature-wise temporal interpolation function $f(\cdot)$ to directly synthesize the intermediate LR feature map $F_{2}^L$ (see Figure~\ref{fig:interp_module}). A general form of the proposed temporal interpolation function can be formulated as:
\begin{equation}
    F_{2}^L = f(F_{1}^L, F_{3}^L) = H(T_{1}(F_{1}^L, \Phi_1), T_{3}(F_{3}^L, \Phi_3))
    \enspace,
\end{equation}
where $T_1(\cdot)$ and $T_3(\cdot)$ are two feature sampling functions and $\Phi_1$ and $\Phi_3$ are sampling parameters; $H(\cdot)$ is a weighted blending function to temporally aggregate sampled features.

For obtaining accurate $F_{2}^L$, $T_1(\cdot)$ should be able to capture forward motion information between $F_{1}^L$ and $F_{2}^L$, and $T_3(\cdot)$ should be able to capture backward motion information between $F_{3}^L$ and $F_{2}^L$. However, $F_{2}^L$ is unavailable for directly computing forward and backward temporally dynamic information in STVSR. 

To mitigate this issue, we utilize temporal information between $F_{1}^L$ and $F_{3}^L$ to approximate the forward and backward motion. Motivated by deformable alignment in~\cite{tian2018tdan} for video super-resolution, we propose to use deformable sampling to implicitly capture dynamic motion information for feature temporal interpolation. Exploiting rich locally temporal contexts by deformable convolutions in the sampling functions, the proposed frame feature temporal interpolation is capable of handling large video motions.

The two temporal sampling functions share the same architecture but have different parameters. For simplicity, we take the $T_1(\cdot)$ as an example. It uses LR feature maps $F_{1}^L$ and $F_{3}^L$ as the input to predict a temporally dynamic offset for sampling the $F_{1}^L$:
\begin{equation}
    \Delta p_{1} = g_1([F_{1}^L, F_{3}^L])
    \enspace,
\end{equation}
where $\Delta p_1$ is a learnable convolution offset and also denotes the sampling parameter: $\Phi_1$; $g_1$ refers to a general function of multiple convolution layers; $[ ,]$ is a channel-wise concatenation operator. Using the learned offset, the sampling function can be implemented with a deformable convolution~\cite{dai2017deformable, zhu2019deformable}: 
\begin{equation}
  T_1(F_1^L, \Phi_1) = DConv(F_{1}^L, \Delta p_{1})
  \enspace.  
\end{equation}
Similarly, we can predict an offset: $\Delta p_{3} = g_3([F_{3}^L, F_{1}^L])$ as the sampling parameter: $\Phi_3$ and then produce sampled features $T_3(F_3^L, \Phi_3)$ via a deformable convolution.

To blend the sampled frame features, we adopt a simple linear weighted blending function $H(\cdot)$:
\begin{equation}
   F_{2}^{L}  = \alpha * T_{1}(F_{1}^L, \Phi_1) + \beta * T_{3}(F_{3}^L, \Phi_3)
   \enspace,
\end{equation}
where $\alpha$ and $\beta$ are two $1\times1$ convolution kernels and $*$ denotes a convolution operator. Since the synthesized $F_2^L$ will be used to predict the intermediate HR video frame $I_2^H$, it will enforce the $F_2^L$ to be close to the real intermediate LR frame feature map. Therefore, the offsets: $\Delta p_{1}$ and $\Delta p_{3}$ will implicitly learn to capture the forward and backward dynamic information, respectively.

Applying the devised deformable temporal interpolation function to LR feature maps: $\{F_{2t-1}^L\}_{t=1}^{n+1}$, we can predict a sequence of intermediate LR frame feature maps: $\{F_{2t}^L\}_{t=1}^{n}$.


\subsection{Deformable ConvLSTM}
\label{dconvlstm}

Given the consecutive LR frame feature maps: $\{F_{t}^L\}_{t=1}^{2n+1}$, we will use a sequence-to-sequence mapping to reconstruct the corresponding HR frames. It has been widely proved in video restoration tasks~\cite{xue2019video, tao2017detail, wang2019edvr} that temporal information is essential. Therefore, instead of generating HR frames from individual feature maps, we adaptively aggregate temporal contexts from neighboring video frames. ConvLSTM \cite{xingjian2015convolutional} is a popular 2D sequential data modeling approach, and we can utilize it for temporal aggregation. At the time step $t$, the ConvLSTM will update hidden state $h_{t}$ and cell state $c_{t}$ with:
\begin{equation}
h_t, c_t = ConvLSTM(h_{t-1},c_{t-1},{F}_t^{L})
\enspace.
\end{equation}
From its internal state updating mechanism \cite{xingjian2015convolutional}, we can observe that the ConvLSTM can only implicitly capture temporal information between previous states: $h_{t-1}$ and $c_{t-1}$ and the current input frame feature map with small 2D convolution receptive fields. So, ConvLSTM has a relatively limited ability to deal with large motions in natural videos. If there are large motions in a video, a severe temporal mismatch between previous states and $F_t^L$ will occur. With the state updating, $h_{t-1}$ and $c_{t-1}$ will propagate temporally mismatched ``noisy'' content but not useful global temporal contexts into $h_t$. Consequently, the generated HR video frame $I_t^H$ from $h_t$ will suffer from annoying visual artifacts.  
 
\begin{figure}
    \centering
    \includegraphics[width=1.0\columnwidth]{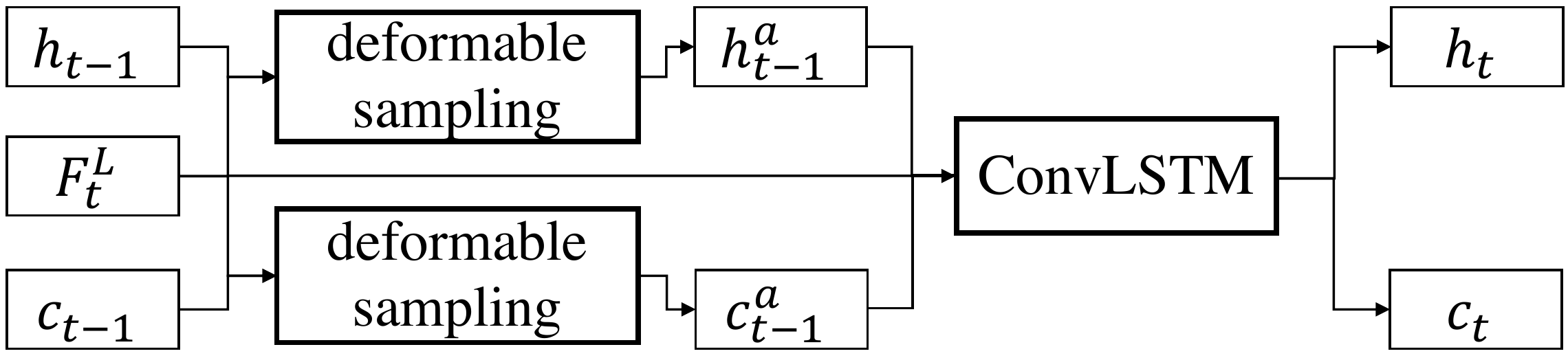}
    \caption{Deformable ConvLSTM for better leveraging global temporal contexts and dealing with fast motion videos. At time step $t$, we learn deformable sampling to adaptively align hidden state $h_{t-1}$ and cell state $c_{t-1}$ with current input feature map: $F_t^L$.}
    \label{fig:conv_lstm}
\end{figure}

To address the large motion issue and effectively leverage global temporal contexts, we embed state-updating cells with deformable alignment into vanilla ConvLSTM (see Figure~\ref{fig:conv_lstm}): 
 \begin{equation}
\begin{split}
& \Delta p^h_{t} = g^h([h_{t-1},F_t^L ]) \enspace,\\
& \Delta p^c_{t} = g^c([c_{t-1},F_t^L ]) \enspace,\\
& h^a_{t-1} = DConv(h_{t-1}, \Delta p^h_{t}) \enspace,\\
& c^a_{t-1} = DConv(c_{t-1}, \Delta p^c_{t}) \enspace,\\
& h_t, c_t = ConvLSTM(h^a_{t-1}, c^a_{t-1},F_t^L) \enspace,
\end{split}
\end{equation}
where $g^h$ and $g^c$ refer to general functions of convolution layers, $\Delta p^h_{t}$ and $\Delta p^c_{t}$ are learned offsets, and $h^a_{t-1}$ and $c^a_{t-1}$ are temporally aligned hidden and cell states, respectively.
Different from the vanilla ConvLSTM, we explicitly enforce $h_{t-1}$ and $c_{t-1}$ to align with the current input feature map: $F_t^L$ in our deformable ConvLSTM, which makes it more effective in handling video motions. Besides, to make full use of temporal information, we perform Deformable ConvLSTM in a bidirectional fashion~\cite{schuster1997bidirectional}. We take temporally reversed feature maps as inputs for the same Deformable ConvLSTM and concatenate hidden states from forward and backward passes as the final output hidden state $h_t$\footnote{We use $h_t$ to refer to the final hidden state, but it will denote a concatenated hidden state in the Bidirectional Deformable ConvLSTM.} for HR video frame reconstruction.

\subsection{Frame Reconstruction}
\label{sec:framerec}
To reconstruct HR frames, we use a shared frame synthesis network, which takes individual hidden state $h_t$ as input and predicts the corresponding HR frame. It has $k_2$ stacked residual units~\cite{lim2017enhanced} for learning deep frame features and adopts a sub-pixel upscaling network with PixelShuffle as in \cite{shi2016real} to obtain HR frames $\{I_{t}^H\}_{t=1}^{2n+1}$. To optimize the network, we utilize a reconstruction loss function:
\begin{equation}
\label{eq:rec_loss}
l_{rec} = \sqrt{||I^{GT}_{t} - I_{t}^H ||^2 + \epsilon^2}
\enspace,
\end{equation}
where $I^{GT}_{t}$ denotes the $t$-th ground-truth HR frame, Charbonnier penalty function~\cite{lai2017deep} is used as the objective function, and $\epsilon$ is empirically set to $1\times10^{-3}$. 
Since spatial and temporal SR problems are intra-related in our STVSR task, the proposed model can simultaneously learn this spatio-temporal interpolation using only supervision from HR frames in an end-to-end manner. 
\begin{table*}[ht]
\caption{Quantitative comparison of two-stage VFI and VSR methods and our results on Vid4~\cite{liu2011bayesian} dataset. The best two results are highlighted in \textcolor{red}{red} and \textcolor{blue}{blue} colors, respectively. We measure the total runtime on the entire Vid4 dataset \cite{liu2011bayesian}. Note that we omit the baseline methods with Bicubic when comparing in terms of runtime.}
\centering
\resizebox{\textwidth}{!}{
\begin{tabular}{cc|ccccccc}
\hline
\multirow{2}{*}{\begin{tabular}[c]{@{}c@{}}VFI\\   Method\end{tabular}} & \multirow{2}{*}{\begin{tabular}[c]{@{}c@{}}SR\\   Method\end{tabular}} & \multicolumn{2}{c}{Vid4} & \multirow{2}{*}{\begin{tabular}[c]{@{}c@{}}Parameters\\   (Million)\end{tabular}} & \multirow{2}{*}{\begin{tabular}[c]{@{}c@{}}Runtime-VFI\\   (s)\end{tabular}} & \multirow{2}{*}{\begin{tabular}[c]{@{}c@{}}Runtime-SR\\   (s)\end{tabular}} & \multirow{2}{*}{\begin{tabular}[c]{@{}c@{}}Total\\   Runtime (s)\end{tabular}} & \multirow{2}{*}{\begin{tabular}[c]{@{}c@{}}Average\\   Runtime (s/frame)\end{tabular}} \\ 
                                                                        &                                                                        & PSNR       & SSIM       \\ \hline \hline
SuperSloMo \cite{jiang2018super}                                                                 & Bicubic                                                                   &   22.84 &   	0.5772 &      	19.8 &   	0.28 &   	- &   	- &   	-                                                                               \\
SuperSloMo \cite{jiang2018super}                                                                & RCAN \cite{zhang2018image}                                                                  & 23.80      & 0.6397     & 19.8+16.0                                                                         & 0.28                                                                         & 68.15                                                                       & 68.43                                                                          & 0.4002                                                                                 \\
SuperSloMo \cite{jiang2018super}                                                                & RBPN \cite{haris2019recurrent}                                                                  & 23.76      & 0.6362  & 	\textcolor{blue}{19.8+12.7} & 	0.28 & 	82.62 & 	82.90 & 	0.4848        \\
SuperSloMo \cite{jiang2018super}                                                                & EDVR \cite{wang2019edvr}                                                                  & 24.40  & 	0.6706	   & 	19.8+20.7  & 	0.28  & 	24.65  & 	\textcolor{blue}{24.93}  & 	\textcolor{blue}{0.1458}                                                                        \\ \hline
SepConv \cite{niklaus2017adsconv}                                                                 & Bicubic                                                                   &   23.51 &   	0.6273 &    	21.7 &   	2.24 &   	- &   	- &   	-                                                           \\
SepConv \cite{niklaus2017adsconv}                                                                & RCAN \cite{zhang2018image}                                                                  & 24.92      & 0.7236      & 21.7+16.0                                                                         & 2.24                                                                         & 68.15                                                                       & 70.39                                                                          & 0.4116                                                                                 \\
SepConv \cite{niklaus2017adsconv}                                                                & RBPN \cite{haris2019recurrent}                                                                  & 26.08      & 0.7751             & 21.7+12.7                                                                         & 2.24                                                                    & 82.62                                                                       & 84.86                                                                          & 0.4963                                                                                 \\
SepConv \cite{niklaus2017adsconv}                                                                & EDVR \cite{wang2019edvr}                                                                  & 25.93      & 0.7792       & 21.7+20.7                                                                         & 2.24                                                                         & 24.65                                                                       & 26.89                                                                          & 0.1572                                                                                 \\
\hline
DAIN \cite{bao2019depth}                                                                   & Bicubic                                                                &    23.55  &   	0.6268  &      	24.0	  &   8.23  &   	-  &   	-  &   	-                         \\
DAIN  \cite{bao2019depth}                                                                  & RCAN \cite{zhang2018image}                                                                  & 25.03      & 0.7261           & 24.0+16.0                                                                         & 8.23                                                                         &           68.15                                                                  & 76.38                                                                           & 0.4467                                                                                 \\
DAIN  \cite{bao2019depth}                                                                  & RBPN \cite{haris2019recurrent}                                                                  & 25.96      & 0.7784        & 24.0+12.7                                                                         & 8.23                                                                         &           82.62                                                                  & 90.85                                                                           & 0.5313                                                                                 \\
DAIN \cite{bao2019depth}                                                                   & EDVR \cite{wang2019edvr}                                                                  & \textcolor{blue}{26.12}      & \textcolor{blue}{0.7836}                 & 24.0+20.7                                                                         & 8.23                                                                         & 24.65                                                                       & 32.88                                                                          & 0.1923                                                                                 \\ \hline
 
\multicolumn{2}{c|}{ZSM (Ours)}                                                                                                                         & \textcolor{red}{26.49}      &      \textcolor{red}{0.8028}              & \textcolor{red}{11.10}                                                                             & -                                                                            & -                                                                           & \textcolor{red}{10.36}                                                                          & \textcolor{red}{0.0606}                                                                                
 
\\ \hline
\end{tabular}
}
\label{tab:result_vid}
\end{table*}

\subsection{Guided Feature Interpolation Learning}
\label{sec:cyclic}
\begin{figure}[t]
    \centering
    \includegraphics[width=1.0\columnwidth]{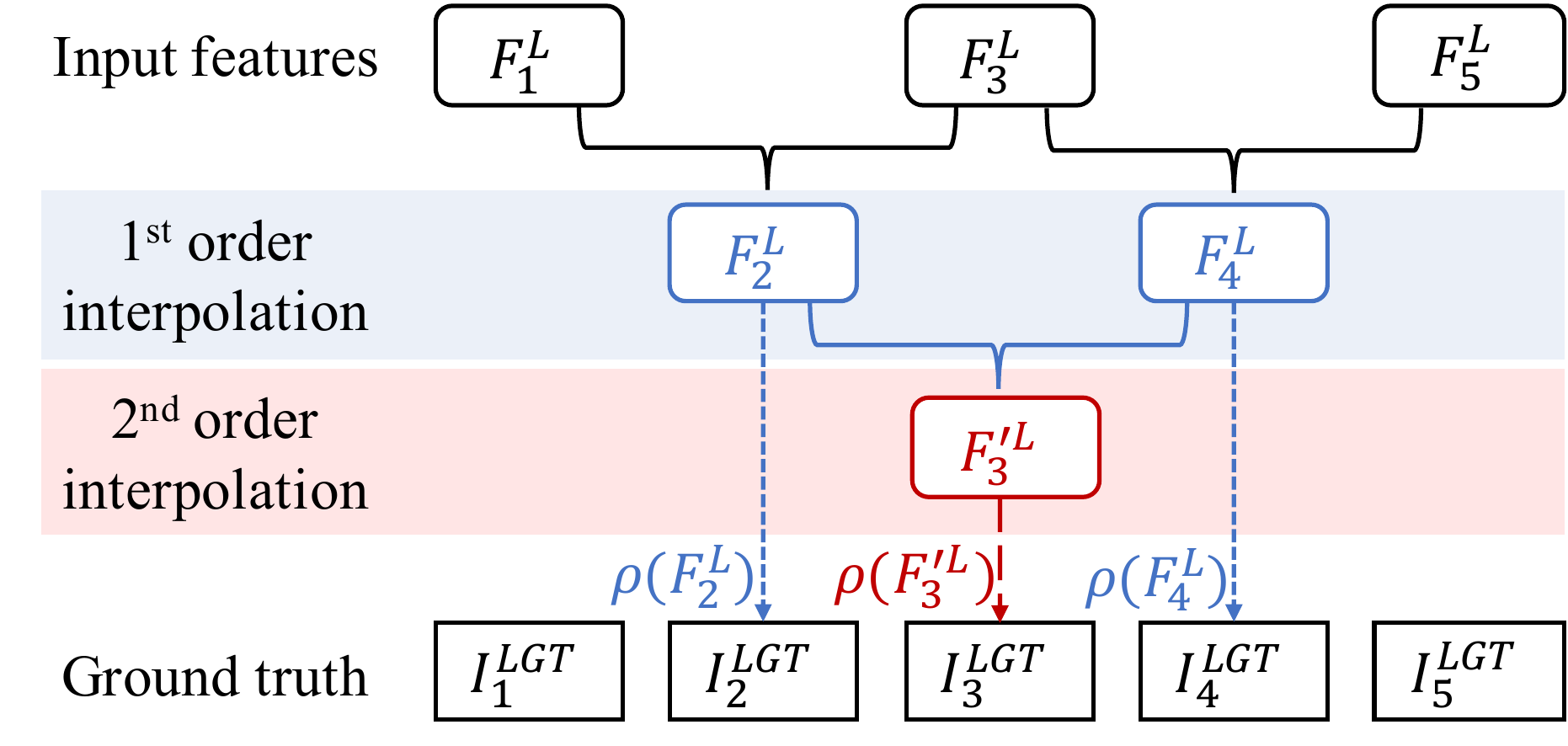}
    \caption{Feature interpolation learning guided by LR frames. The cyclic interpolation loss is computed between the ground truth LR frames and the 1st-order and 2nd-order interpolated LR frames. By minimizing the difference of LR frames and their corresponding interpolation acquired at each order, our temporal interpolation module can be self-supervised with the natural temporal coherence.}
    \label{fig:cyclic_fea_interp}
\end{figure}

In addition, we employ a cyclic interpolation loss to guide the learning of frame feature interpolation with LR frames. It utilizes the inherent temporal coherence in natural video (see Figure \ref{fig:cyclic_fea_interp}).  

Given a sequence of LR, LFR inputs $\{I^L_{2t-1}\}_{t=1}^{n+1}$, we can obtain the extracted frame feature maps $\{F^L_{2t-1} \}^{n+1}_{t=1}$, and the interpolated intermediate frame feature maps $\{F_{2t}^L \}_{t=1}^n$. During the training phase, we have a a set of LR ground truth $\{I^{LGT}_t\}_{t=1}^{2n+1}$.  The first-order interpolation loss is defined as:
\begin{equation}
\label{eq:1st_cyc_loss}
    l^1_i = ||I^{LGT}_{2t} - \rho(F^L_{2t})||_c,
\end{equation}
where $||\cdot||_c$ stands for the Charbonnier penalty function as defined in Equation \eqref{eq:rec_loss}, and $\rho$ represents the LR synthesis module that turns feature maps into the corresponding LR frames. We apply $k_3$ stacked residual blocks~\cite{lim2017enhanced} and a convolution layer, which is similar to the the design in Section \ref{sec:framerec}, to predict the LR frames.

If we conduct feature interpolation on the acquired intermediate frame feature maps $\{F_{2t}^L \}_{t=1}^n$, we can get a sequence of re-interpolated feature maps $\{F_{2t+1}^{'L} \}_{t=1}^{n-1}$. Similar to Equation \eqref{eq:1st_cyc_loss}, the second-order cyclic interpolation loss is defined as:
\begin{equation}
\label{eq:2nd_cyc_loss}
    l^2_i = ||I^{LGT}_{2t+1} - \rho(F^{'L}_{2t+1})||_c.
\end{equation}

The overall training loss is the weighted summation of the reconstruction loss, and the 1st- and 2nd-order cyclic interpolation losses:
\begin{equation}
    L = \lambda_1 l_{rec} + \lambda_2 l_i^1 + \lambda_3 l_i^2.
\end{equation}

\begin{figure*}[htbp]
	\scriptsize
	\centering
	\begin{tabular}{cc}
		\hspace{-0.43cm}
        \begin{adjustbox}{valign=t}
			\begin{tabular}{c}
				\includegraphics[width=0.273\textwidth]{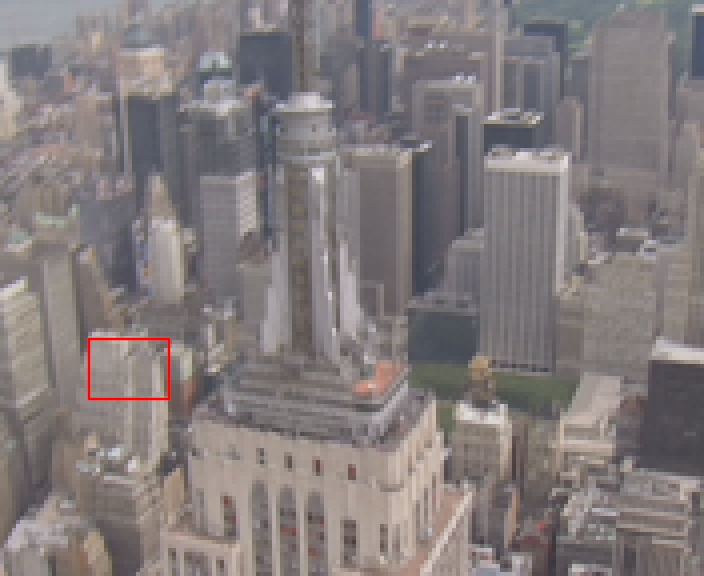}
				\\
				Overlayed LR frames
			
			\end{tabular}
		\end{adjustbox}
		\hspace{-4.3mm}
		\begin{adjustbox}{valign=t}
			\begin{tabular}{ccccc}
				\includegraphics[width=\widthscale \textwidth]{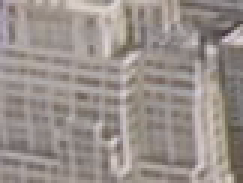} \hspace{-4mm} &
				\includegraphics[width=\widthscale \textwidth]{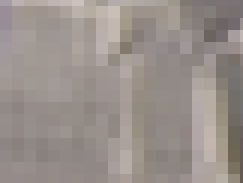} \hspace{-4mm} &
				\includegraphics[width=\widthscale \textwidth]{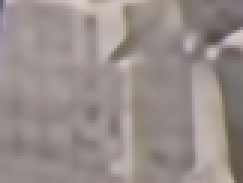} \hspace{-4mm} &
				\includegraphics[width=\widthscale \textwidth]{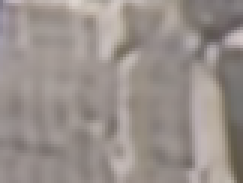}\hspace{-3.5mm} &
				\includegraphics[width=\widthscale \textwidth]{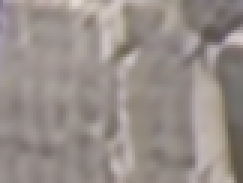}
				\\
				HR \hspace{-4mm} &
    			Overlayed LR \hspace{-4mm} &
				SepConv+RCAN \hspace{-4mm} &
				SepConv+RBPN\hspace{-4mm} &
				SepConv+EDVR

				\\
				\includegraphics[width=\widthscale \textwidth]{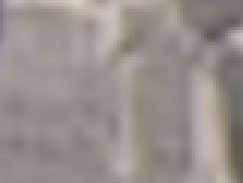} \hspace{-4mm} &
				\includegraphics[width=\widthscale \textwidth]{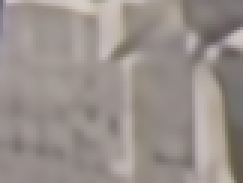} \hspace{-4mm} &
				\includegraphics[width=\widthscale \textwidth]{Figs/fig2/city/city_004_dain_rbpn.png} \hspace{-4mm} &
				\includegraphics[width=\widthscale \textwidth]{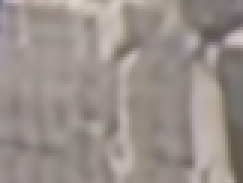}\hspace{-3.5mm} &
				\includegraphics[width=\widthscale \textwidth]{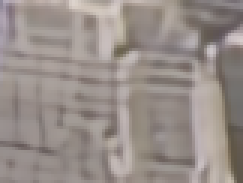}
				\\
				DAIN+Bicubic \hspace{-4mm} &
				DAIN+RCAN \hspace{-4mm} &
				DAIN+RPBN\hspace{-4mm} &
				DAIN+EDVR\hspace{-4mm} &
				\textbf{ZSM (Ours)}
				\\
			\end{tabular}
			\end{adjustbox}
			\vspace{1mm}
         
         \\
    \hspace{-0.43cm}
    \begin{adjustbox}{valign=t}
			\begin{tabular}{c}
				\includegraphics[width=0.273\textwidth]{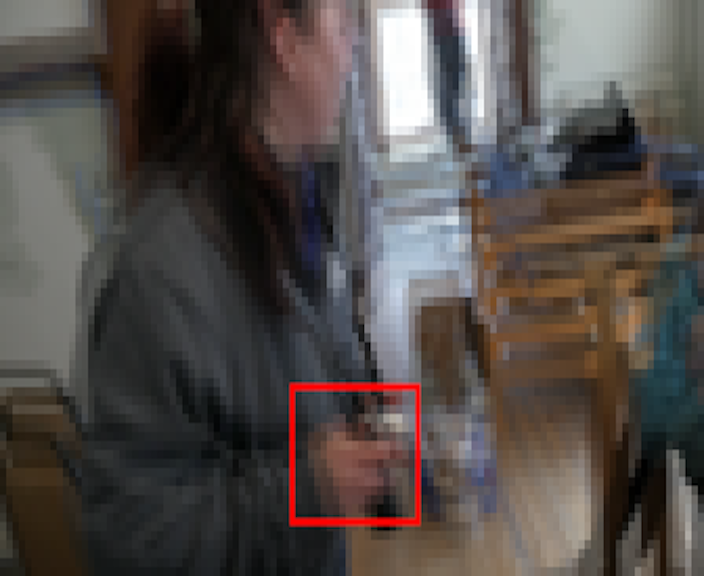}
				\\
				Overlayed LR frames
			
			\end{tabular}
		\end{adjustbox}
		\hspace{-4.3mm}
		\begin{adjustbox}{valign=t}
			\begin{tabular}{ccccc}
				\includegraphics[width=\widthscale \textwidth]{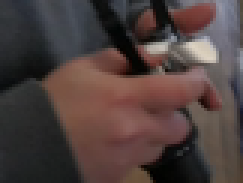} \hspace{-4mm} &
				\includegraphics[width=\widthscale \textwidth]{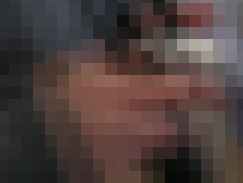} \hspace{-4mm} &
				\includegraphics[width=\widthscale \textwidth]{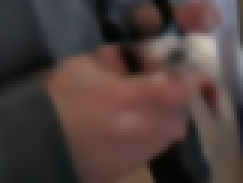} \hspace{-4mm} &
				\includegraphics[width=\widthscale \textwidth]{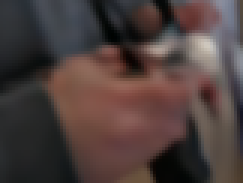}\hspace{-3.5mm} &
				\includegraphics[width=\widthscale \textwidth]{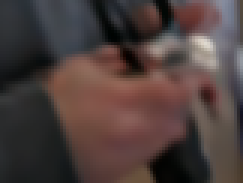}
				\\
			    HR \hspace{-4mm} &
    			Overlayed LR \hspace{-4mm} &
				SepConv+RCAN \hspace{-4mm} &
				SepConv+RBPN\hspace{-4mm} &
				SepConv+EDVR

				\\
				\includegraphics[width=\widthscale \textwidth]{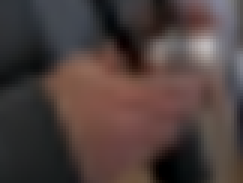} \hspace{-4mm} &
				\includegraphics[width=\widthscale \textwidth]{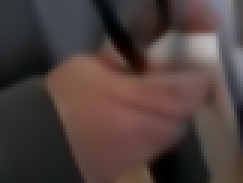} \hspace{-4mm} &
				\includegraphics[width=\widthscale \textwidth]{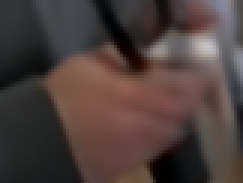} \hspace{-4mm} &
				\includegraphics[width=\widthscale \textwidth]{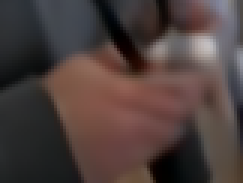}\hspace{-3.5mm} &
				\includegraphics[width=\widthscale \textwidth]{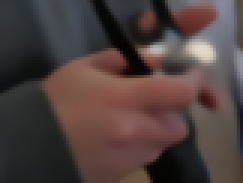}
				\\
			    DAIN+Bicubic \hspace{-4mm} &
				DAIN+RCAN \hspace{-4mm} &
				DAIN+RPBN\hspace{-4mm} &
				DAIN+EDVR\hspace{-4mm} &
				\textbf{ZSM (Ours)}
				\\
			\end{tabular}
			\end{adjustbox}
			\vspace{1mm}
					      \\
	\hspace{-0.43cm}
    \begin{adjustbox}{valign=t}
			\begin{tabular}{c}
				\includegraphics[width=0.273\textwidth]{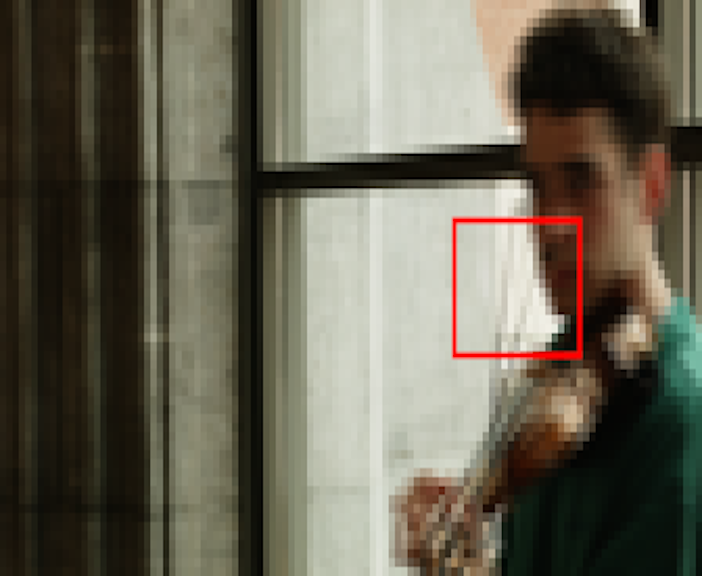}
				\\
				Overlayed LR frames
			
			\end{tabular}
		\end{adjustbox}
		\hspace{-4.3mm}
		\begin{adjustbox}{valign=t}
			\begin{tabular}{ccccc}
				\includegraphics[width=\widthscale \textwidth]{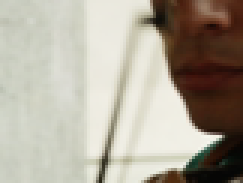} \hspace{-4mm} &
				\includegraphics[width=\widthscale \textwidth]{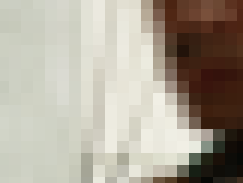} \hspace{-4mm} &
				\includegraphics[width=\widthscale \textwidth]{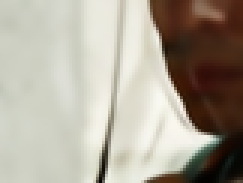} \hspace{-4mm} &
				\includegraphics[width=\widthscale \textwidth]{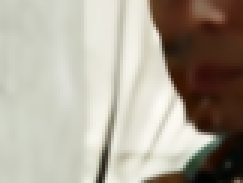}\hspace{-3.5mm} &
				\includegraphics[width=\widthscale \textwidth]{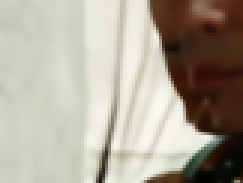}
				\\
			    HR \hspace{-4mm} &
    			Overlayed LR \hspace{-4mm} &
				SepConv+RCAN \hspace{-4mm} &
				SepConv+RBPN\hspace{-4mm} &
				SepConv+EDVR

				\\
				\includegraphics[width=\widthscale \textwidth]{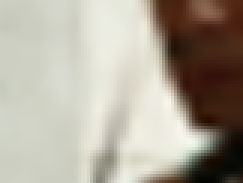} \hspace{-4mm} &
				\includegraphics[width=\widthscale \textwidth]{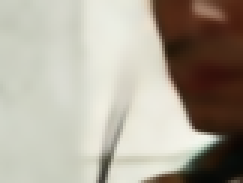} \hspace{-4mm} &
				\includegraphics[width=\widthscale \textwidth]{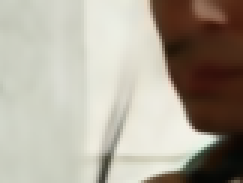} \hspace{-4mm} &
				\includegraphics[width=\widthscale \textwidth]{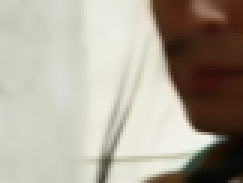}\hspace{-3.5mm} &
				\includegraphics[width=\widthscale \textwidth]{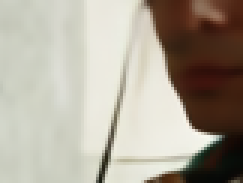}
				\\
			    DAIN+Bicubic \hspace{-4mm} &
				DAIN+RCAN \hspace{-4mm} &
				DAIN+RPBN\hspace{-4mm} &
				DAIN+EDVR\hspace{-4mm} &
				\textbf{ZSM (Ours)}
				\\
			\end{tabular}
			\end{adjustbox}
			\vspace{1mm}
							      \\
    \hspace{-0.43cm}
    \begin{adjustbox}{valign=t}
			\begin{tabular}{c}
				\includegraphics[width=0.273\textwidth]{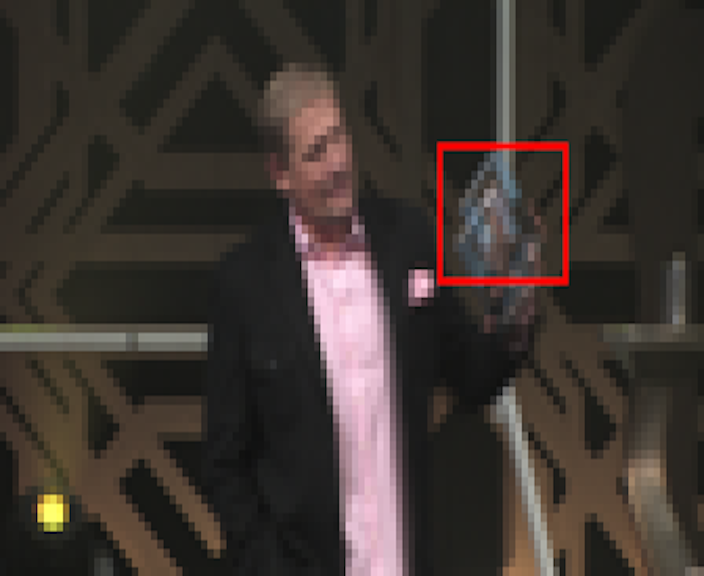}
				\\
				Overlayed LR frames
			
			\end{tabular}
		\end{adjustbox}
		\hspace{-4.3mm}
		\begin{adjustbox}{valign=t}
			\begin{tabular}{ccccc}
				\includegraphics[width=\widthscale \textwidth]{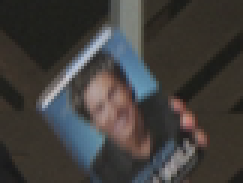} \hspace{-4mm} &
				\includegraphics[width=\widthscale \textwidth]{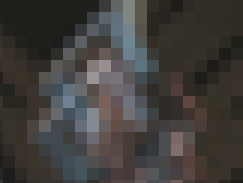} \hspace{-4mm} &
				\includegraphics[width=\widthscale \textwidth]{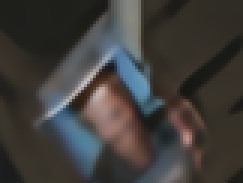} \hspace{-4mm} &
				\includegraphics[width=\widthscale \textwidth]{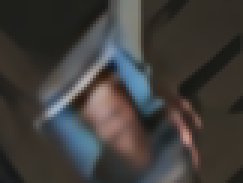}\hspace{-3.5mm} &
				\includegraphics[width=\widthscale \textwidth]{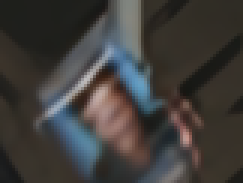}
				\\
			    HR \hspace{-4mm} &
    			Overlayed LR \hspace{-4mm} &
				SepConv+RCAN \hspace{-4mm} &
				SepConv+RBPN\hspace{-4mm} &
				SepConv+EDVR

				\\
				\includegraphics[width=\widthscale \textwidth]{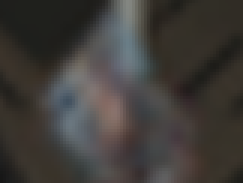} \hspace{-4mm} &
				\includegraphics[width=\widthscale \textwidth]{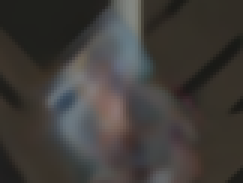} \hspace{-4mm} &
				\includegraphics[width=\widthscale \textwidth]{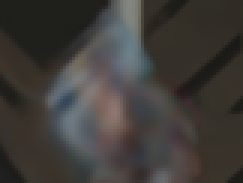} \hspace{-4mm} &
				\includegraphics[width=\widthscale \textwidth]{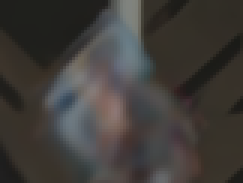}\hspace{-3.5mm} &
				\includegraphics[width=\widthscale \textwidth]{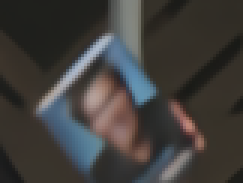}
				\\
			    DAIN+Bicubic \hspace{-4mm} &
				DAIN+RCAN \hspace{-4mm} &
				DAIN+RPBN\hspace{-4mm} &
				DAIN+EDVR\hspace{-4mm} &
				\textbf{ZSM (Ours)}
				\\
			\end{tabular}
			\end{adjustbox}
  
	\end{tabular}
	\caption{Visual comparisons of different methods on Vid4 and Vimeo datasets. Our one-stage Zooming SlowMo model (ZSM) can generate more visually appealing HR video frames with fewer blurring artifacts and more accurate image structures. }
	\label{fig:stvsrresult}
	\vspace{-2mm}
\end{figure*}

\begin{figure*}[t]
\captionsetup[subfigure]{labelformat=empty}
\begin{center}

\begin{subfigure}[b]{0.195\linewidth}
     \includegraphics[width=\linewidth]{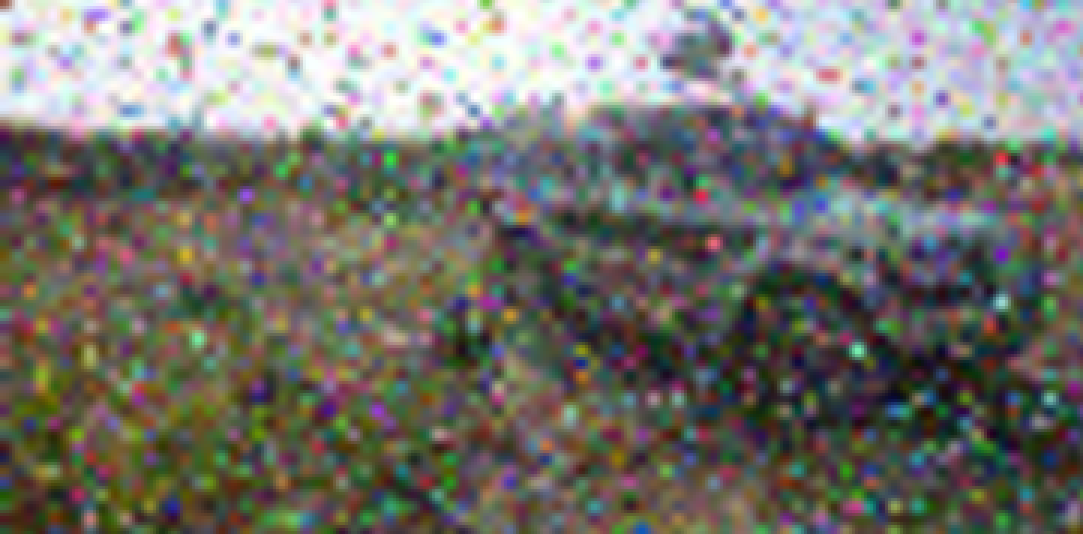}
  \end{subfigure}
  \begin{subfigure}[b]{0.195\linewidth}
  \includegraphics[width=\linewidth]{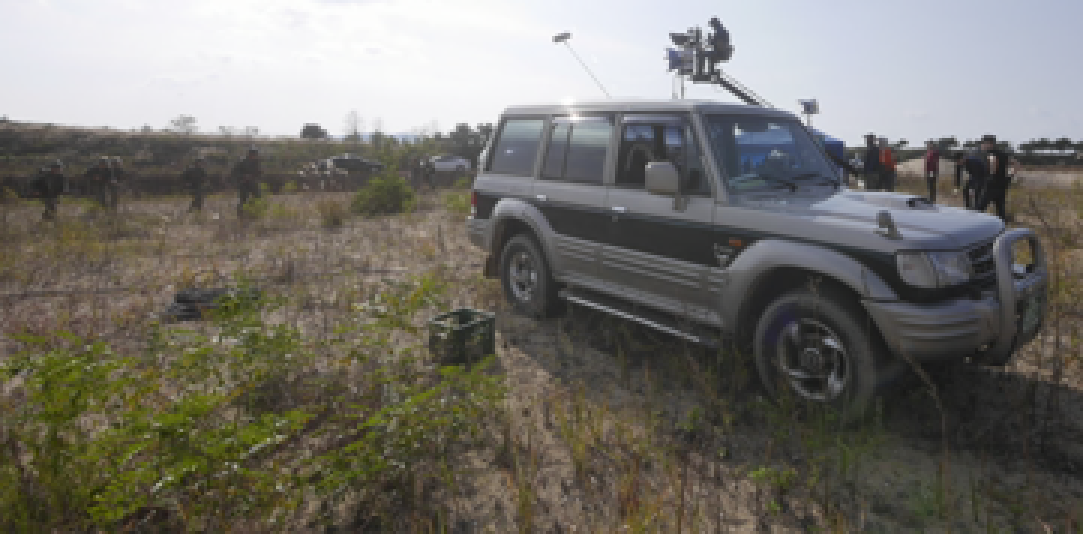}
  \end{subfigure}
  \begin{subfigure}[b]{0.195\linewidth}
     \includegraphics[width=\linewidth]{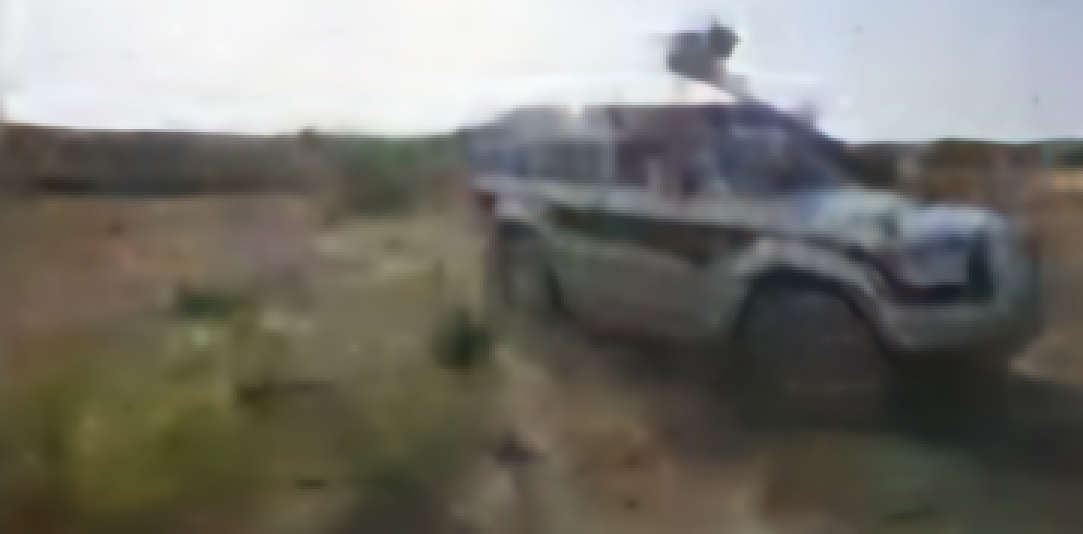}
  \end{subfigure}
  \begin{subfigure}[b]{0.195\linewidth}
  \includegraphics[width=\linewidth]{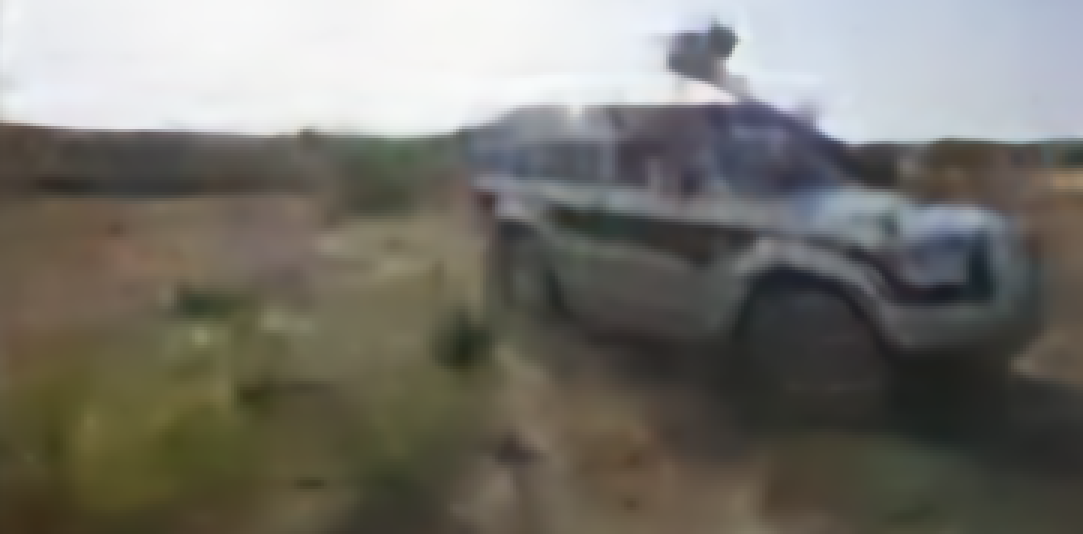}
  \end{subfigure}
  \begin{subfigure}[b]{0.195\linewidth}
  \includegraphics[width=\linewidth]{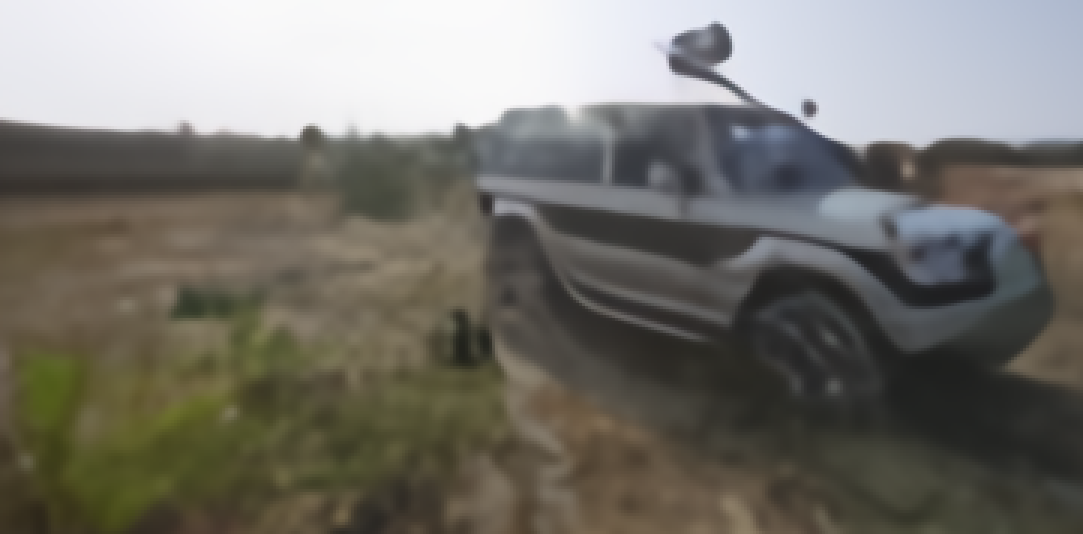}
  \end{subfigure}\vspace{1mm}

   \begin{subfigure}[b]{0.195\linewidth}
     \includegraphics[width=\linewidth]{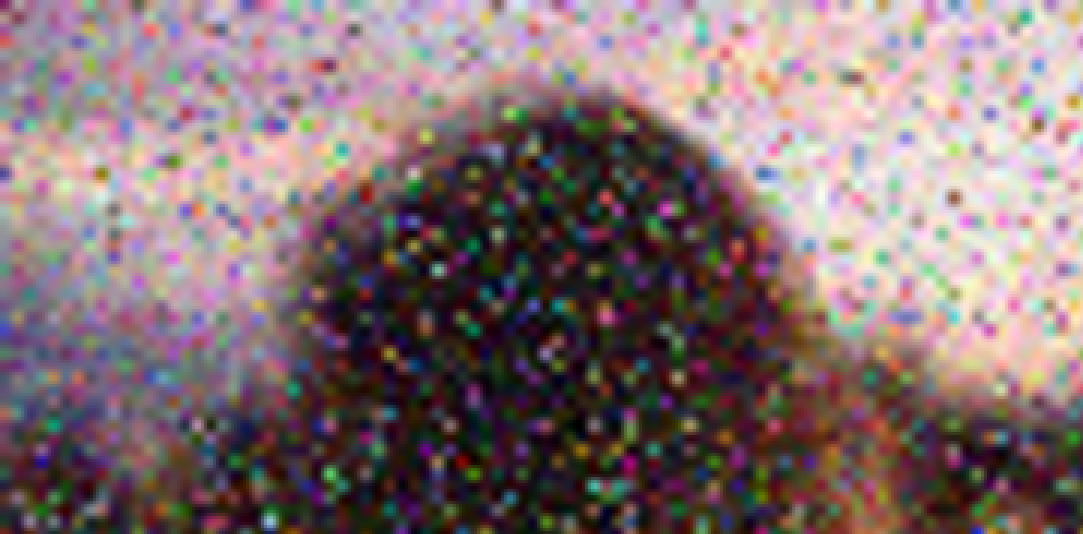}
  \end{subfigure}
  \begin{subfigure}[b]{0.195\linewidth}
  \includegraphics[width=\linewidth]{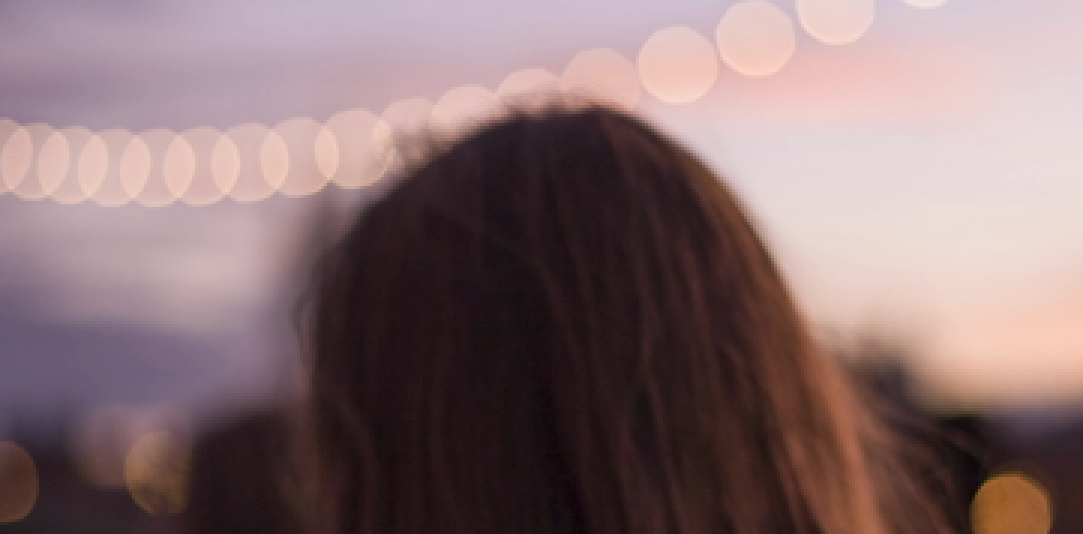}
  \end{subfigure}
  \begin{subfigure}[b]{0.195\linewidth}
     \includegraphics[width=\linewidth]{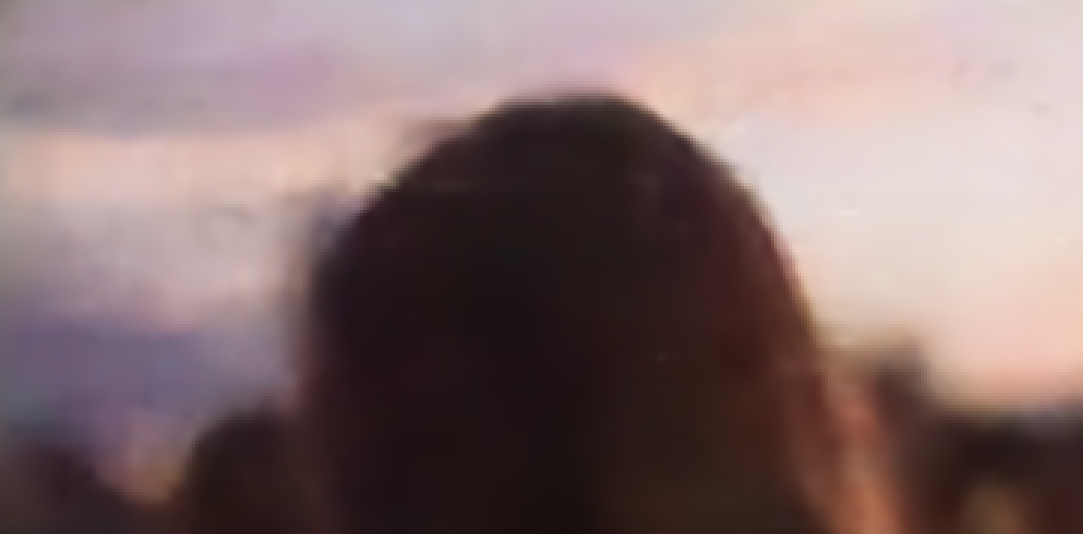}
  \end{subfigure}
  \begin{subfigure}[b]{0.195\linewidth}
  \includegraphics[width=\linewidth]{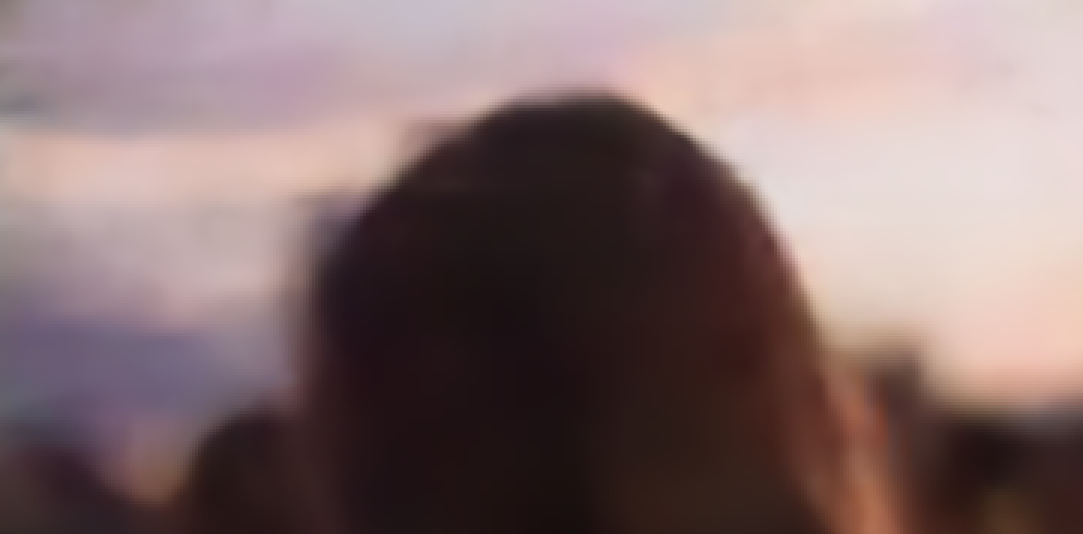}
  \end{subfigure}
  \begin{subfigure}[b]{0.195\linewidth}
  \includegraphics[width=\linewidth]{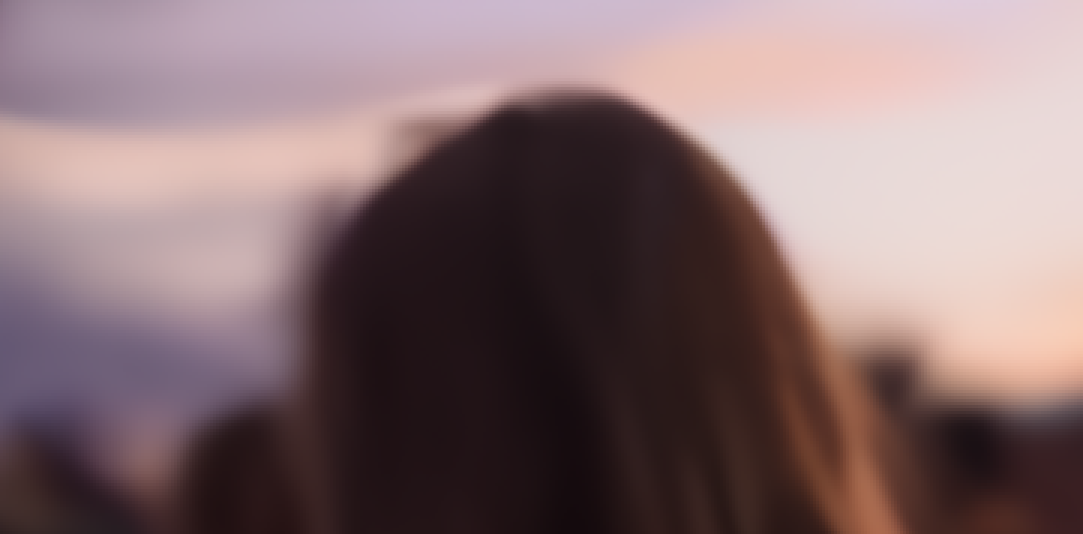}
  \end{subfigure}\vspace{1mm}
 
  \begin{subfigure}[b]{0.195\linewidth}
     \includegraphics[width=\linewidth]{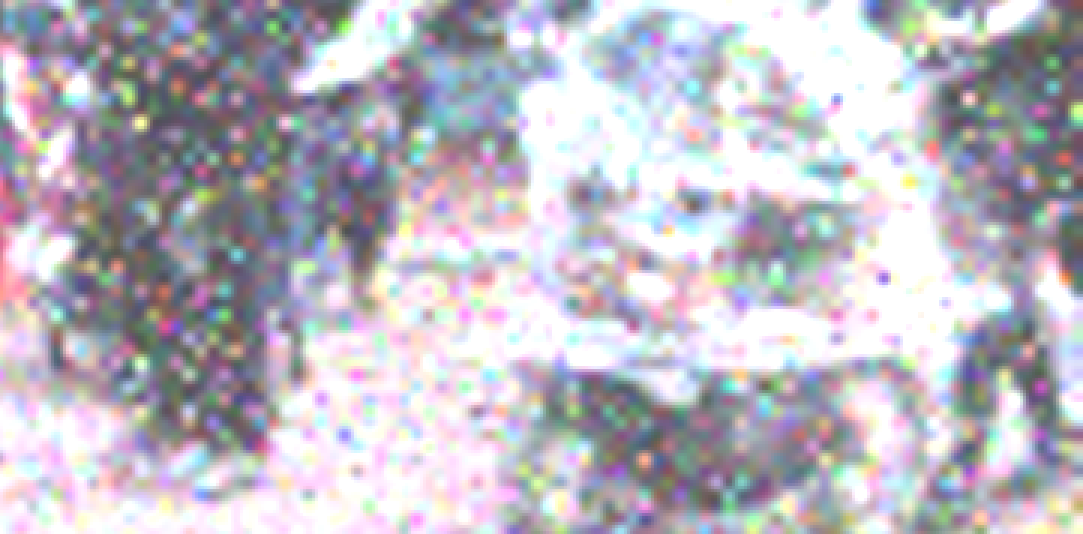}
     \subcaption{Overlayed LR}
  \end{subfigure}
  \begin{subfigure}[b]{0.195\linewidth}
  \includegraphics[width=\linewidth]{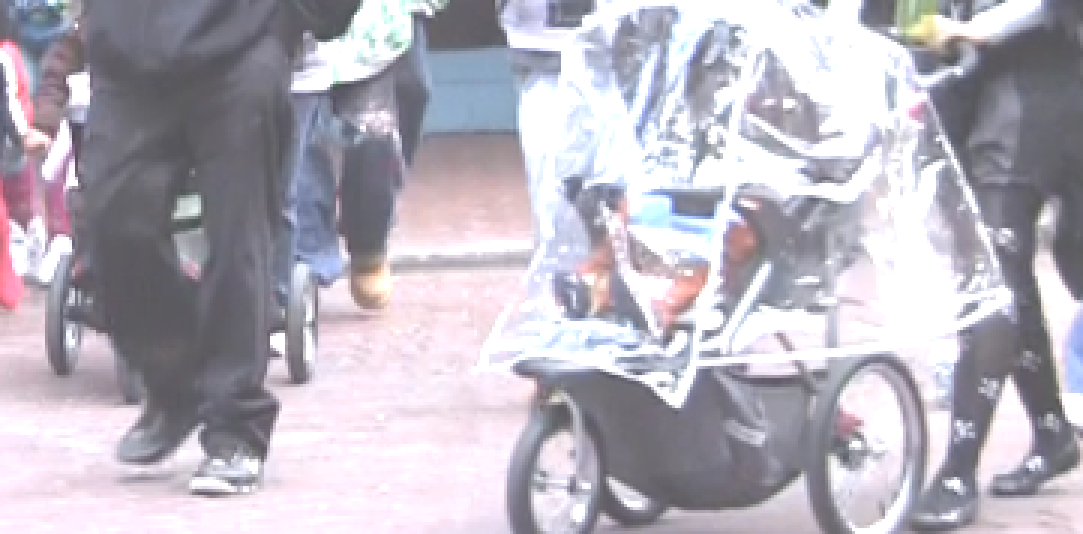}
  \subcaption{HR}
  \end{subfigure}
  \begin{subfigure}[b]{0.195\linewidth}
     \includegraphics[width=\linewidth]{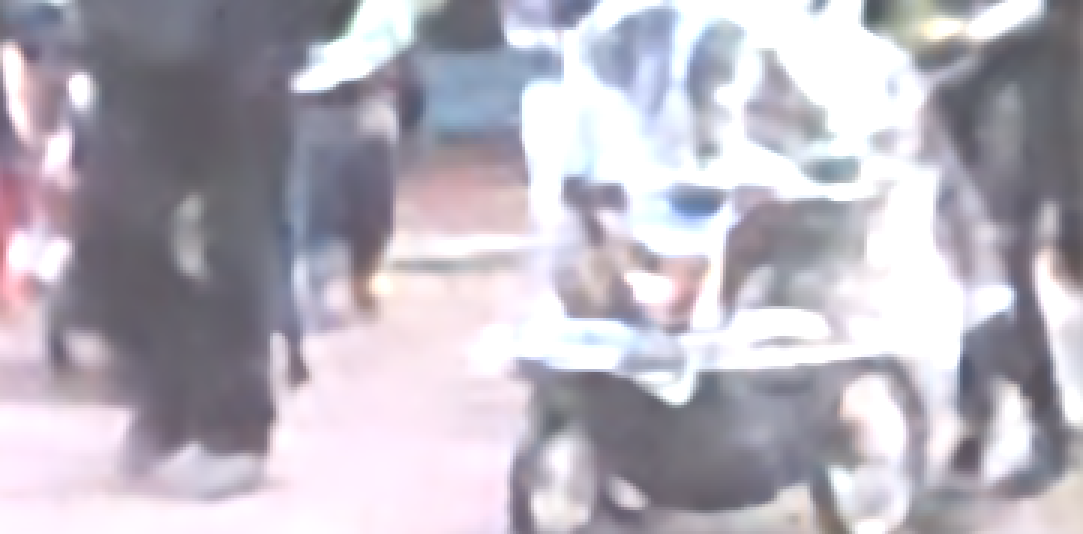}
     \subcaption{Toflow+Sepconv+EDVR}
  \end{subfigure}
  \begin{subfigure}[b]{0.195\linewidth}
  \includegraphics[width=\linewidth]{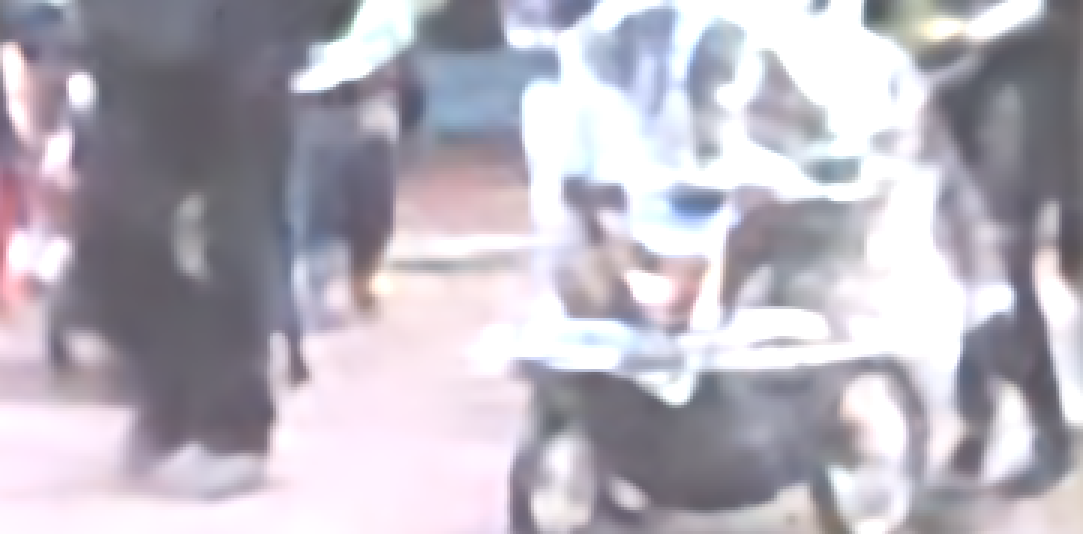}
  \subcaption{Toflow+DAIN+EDVR}
  \end{subfigure}
  \begin{subfigure}[b]{0.195\linewidth}
  \includegraphics[width=\linewidth]{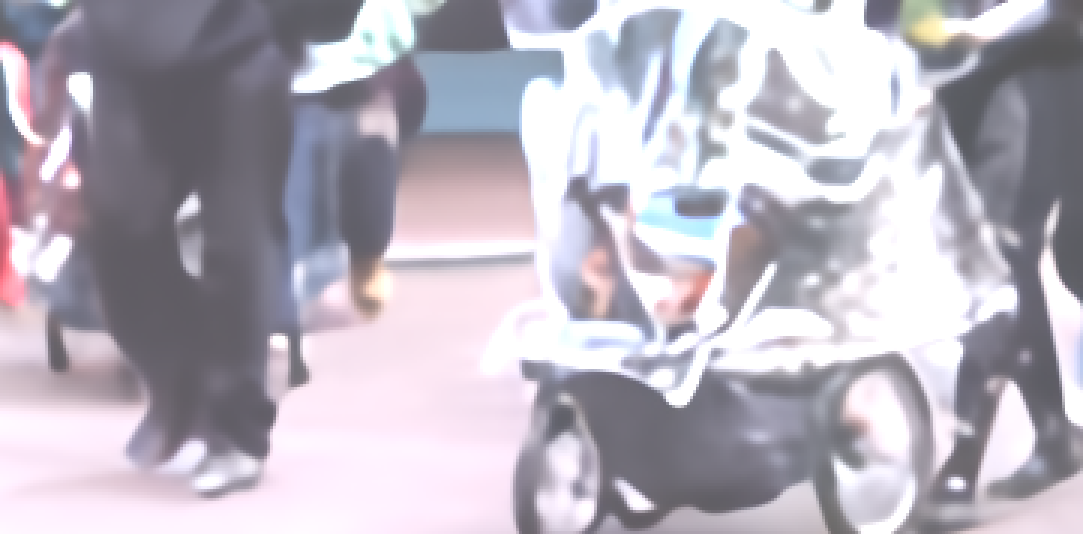}
  \subcaption{\textbf{ZSM (Ours)}}
  \end{subfigure}
\end{center}
\vspace{-3mm}
   \caption{Visual comparisons of different methods on noisy input video frames. Our one-stage Zooming SlowMo model (ZSM) can effectively restore clean missing HR frames from noisy LR frames.
   }
 \label{fig:noise}
\end{figure*}
\subsection{Implementation Details}
\label{sec:implement}
In our implementation, $k_1 = 5$, $k_2 = 40$, and $k_3$ residual blocks are used in feature extraction, HR frame reconstruction, and LR frame reconstruction modules respectively. We randomly crop a sequence of down-sampled image patches with the size of $32\times32$ and take out the odd-indexed 4 frames as LFR and LR inputs, and the corresponding consecutive 7-frame sequence of $4\times$\footnote{Considering recent state-of-the-art methods (\eg, EDVR~\cite{wang2019edvr} and RBPN~\cite{haris2019recurrent}) use only 4 as the upscaling factor, we adopt the same practice.} size as supervision. Besides, we perform data augmentation by randomly rotating $90^{\circ}$, $180^{\circ}$, and $270^{\circ}$, and horizontal-flipping. We adopt a Pyramid, Cascading and Deformable (PCD) structure in~\cite{wang2019edvr} to employ deformable alignment and apply Adam \cite{kingma2014adam} optimizer, where we decay the learning rate with a cosine annealing for each batch \cite{loshchilov2016sgdr} from 4$\times10^{-4}$ to 1$\times10^{-7}$. We set batch size as 24 and train the network on $2$ Nvidia Titan XP GPUs.

\begin{figure*}[t]
\captionsetup[subfigure]{labelformat=empty}
\begin{center}
\begin{subfigure}[b]{0.195\linewidth}
     \includegraphics[width=\linewidth]{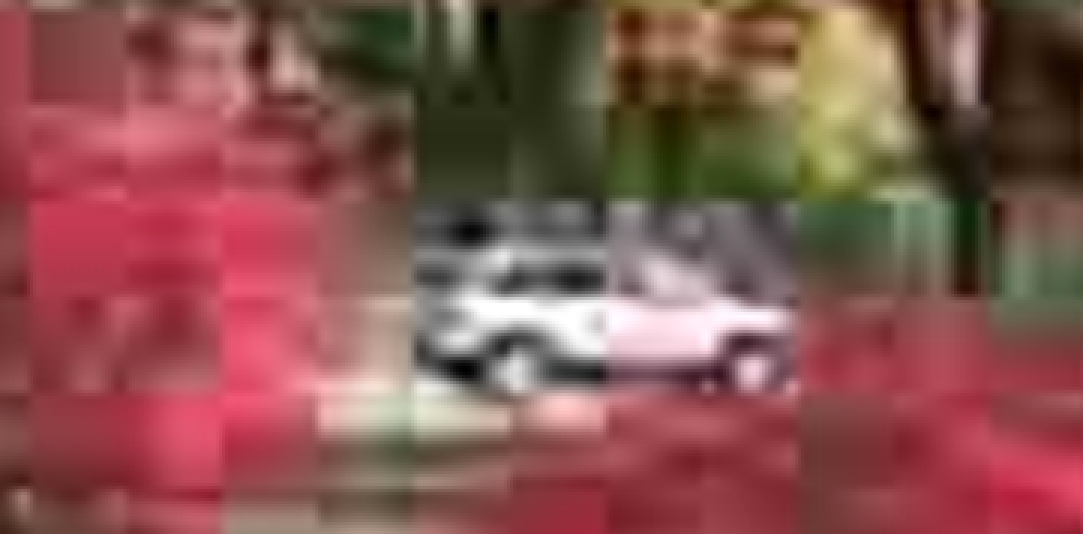}
  \end{subfigure}
  \begin{subfigure}[b]{0.195\linewidth}
  \includegraphics[width=\linewidth]{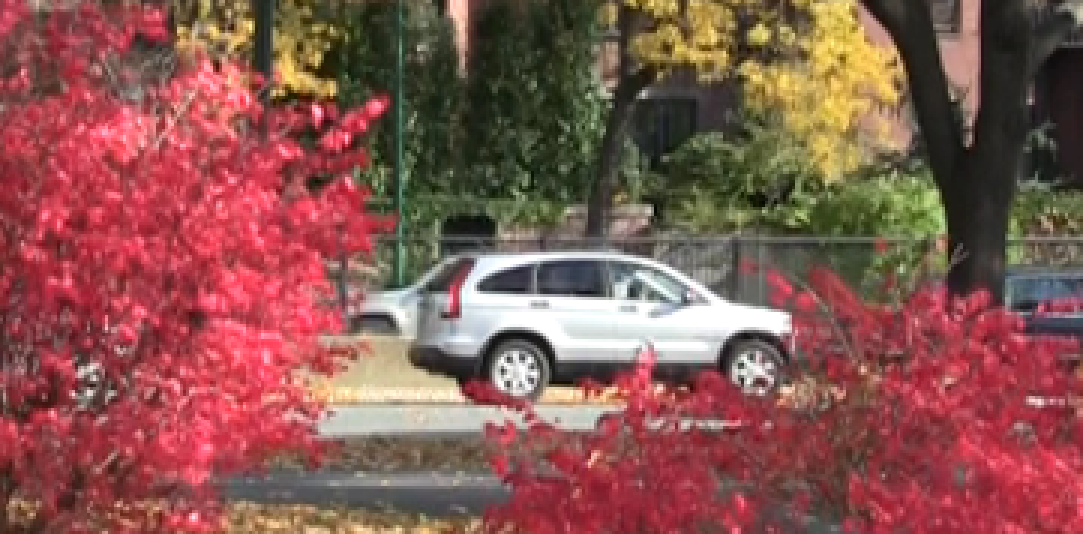}
  \end{subfigure}
  \begin{subfigure}[b]{0.195\linewidth}
     \includegraphics[width=\linewidth]{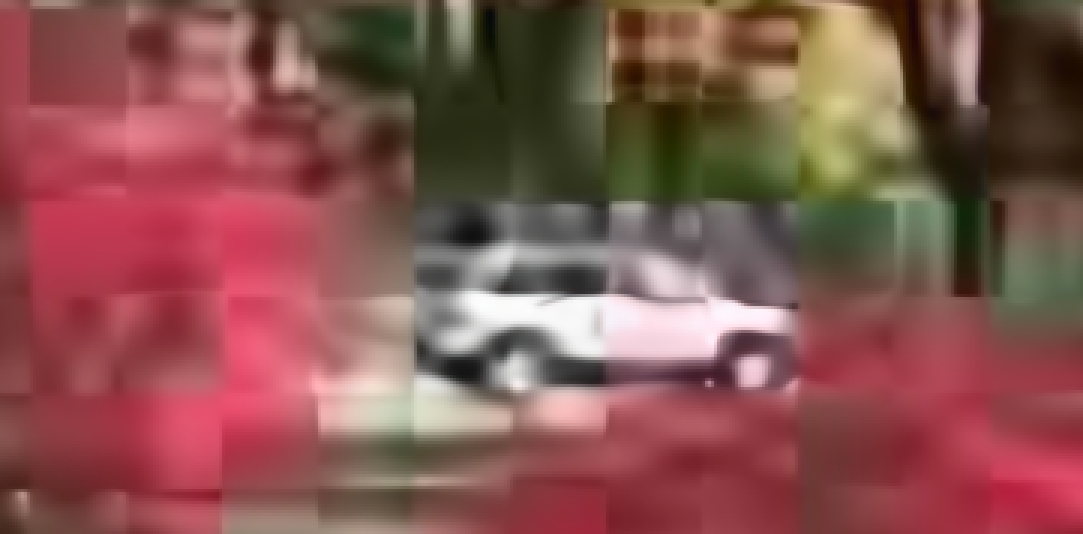}
  \end{subfigure}
  \begin{subfigure}[b]{0.195\linewidth}
  \includegraphics[width=\linewidth]{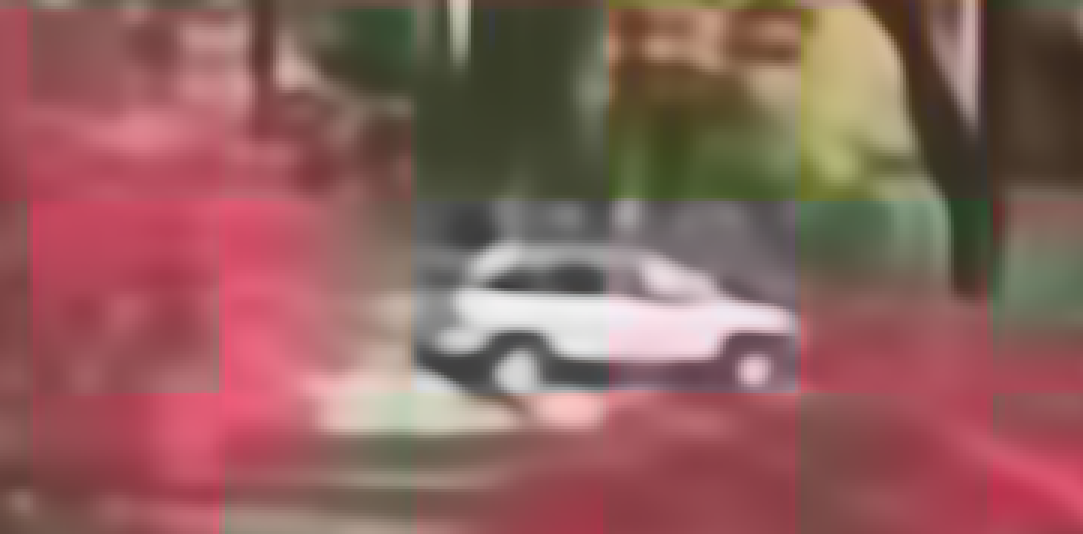}
  \end{subfigure}
  \begin{subfigure}[b]{0.195\linewidth}
  \includegraphics[width=\linewidth]{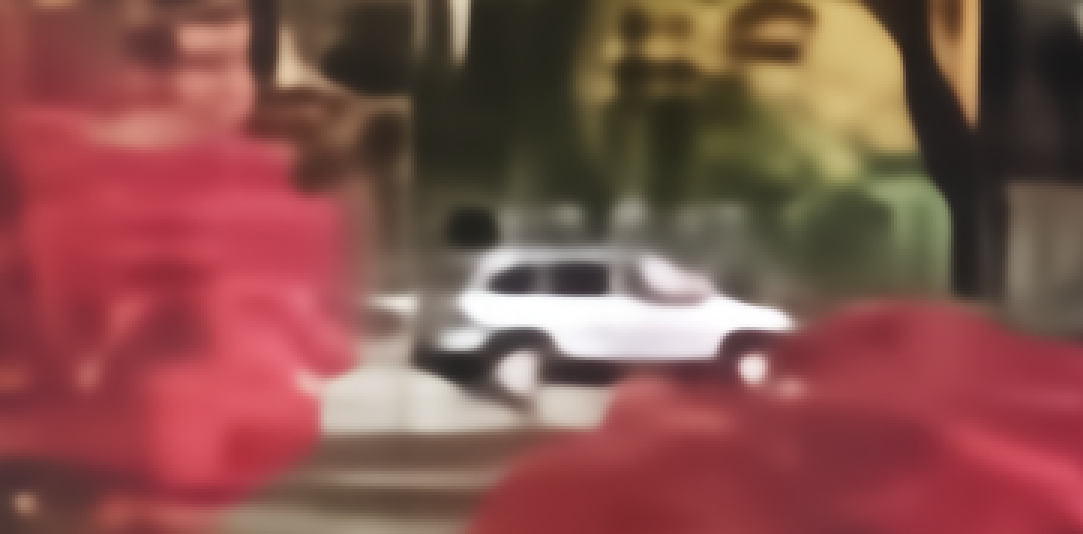}
  \end{subfigure}\vspace{1mm}

  

   \begin{subfigure}[b]{0.195\linewidth}
     \includegraphics[width=\linewidth]{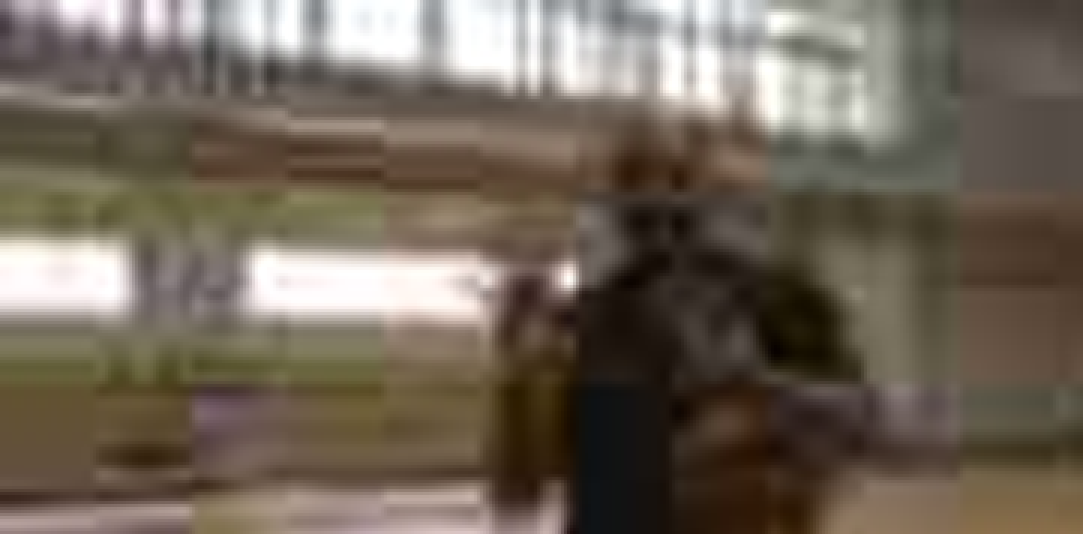}
  \end{subfigure}
  \begin{subfigure}[b]{0.195\linewidth}
  \includegraphics[width=\linewidth]{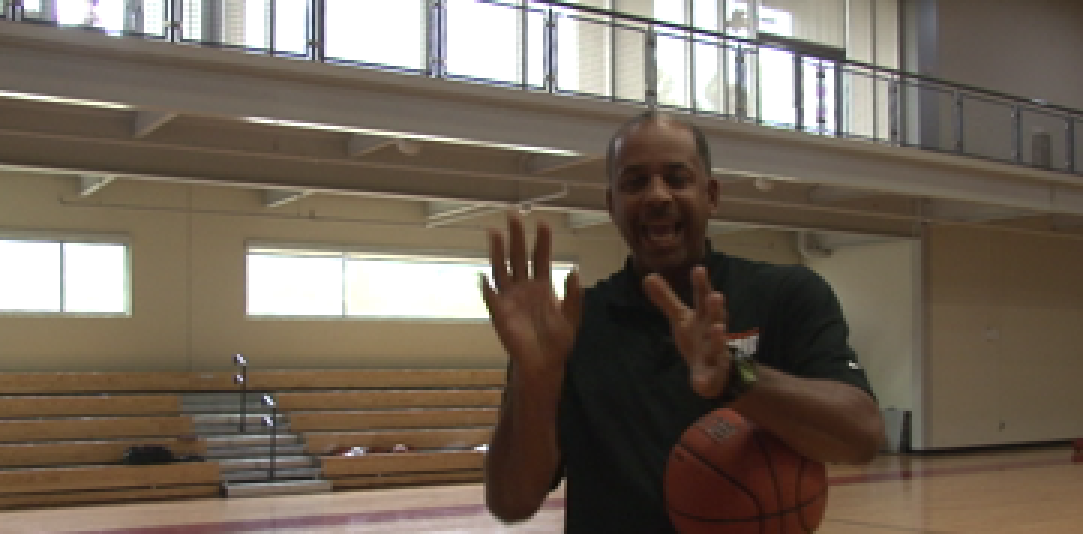}
  \end{subfigure}
  \begin{subfigure}[b]{0.195\linewidth}
     \includegraphics[width=\linewidth]{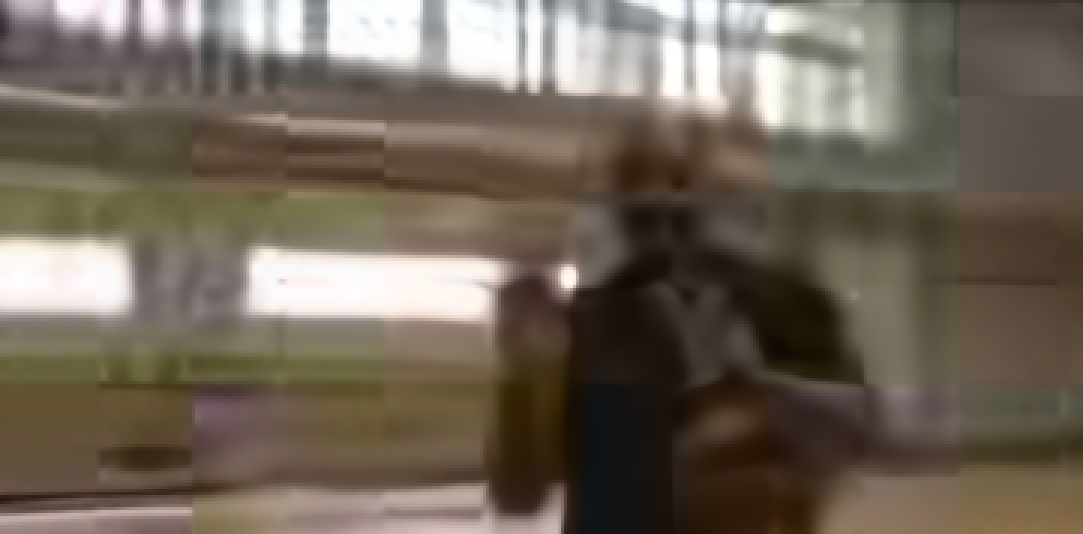}
  \end{subfigure}
  \begin{subfigure}[b]{0.195\linewidth}
  \includegraphics[width=\linewidth]{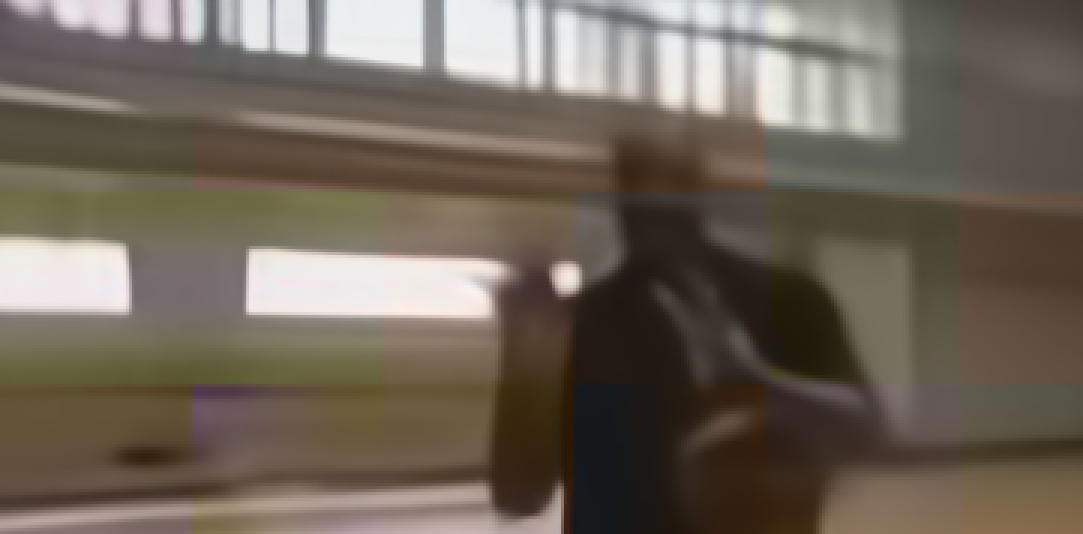}
  \end{subfigure}
  \begin{subfigure}[b]{0.195\linewidth}
  \includegraphics[width=\linewidth]{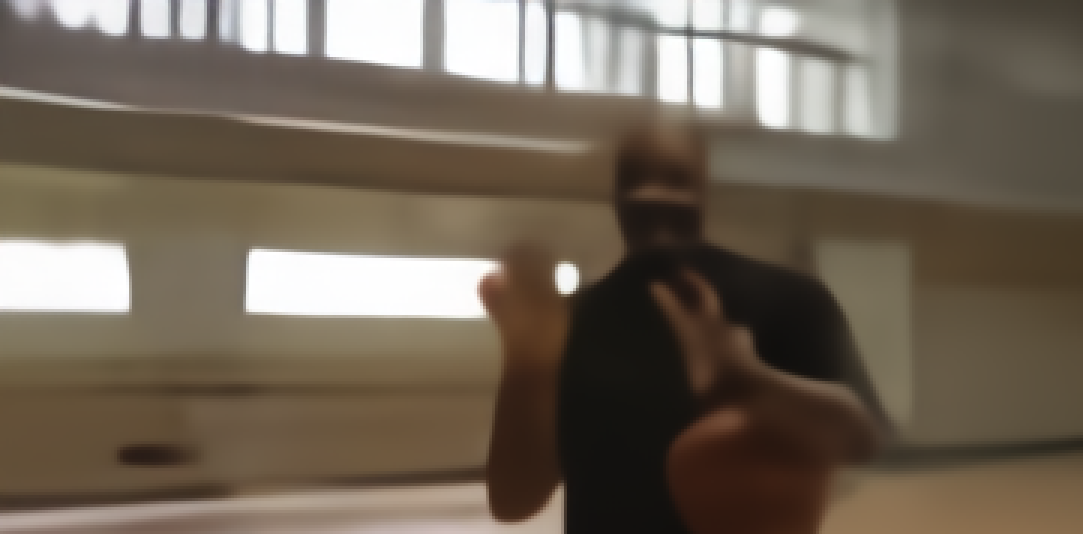}
  \end{subfigure}\vspace{1mm}
  
  
      \begin{subfigure}[b]{0.195\linewidth}
     \includegraphics[width=\linewidth]{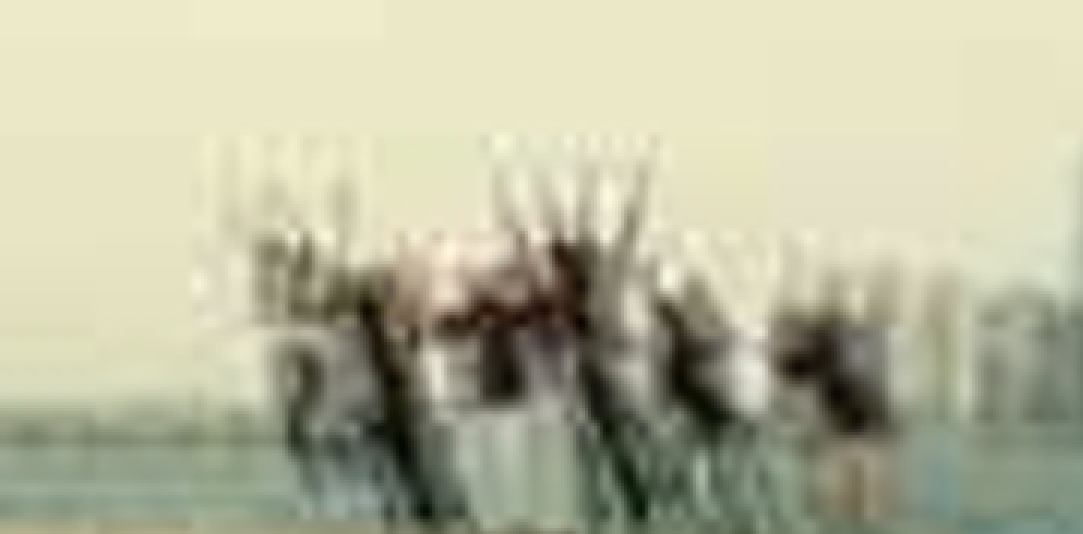}
  \end{subfigure}
  \begin{subfigure}[b]{0.195\linewidth}
  \includegraphics[width=\linewidth]{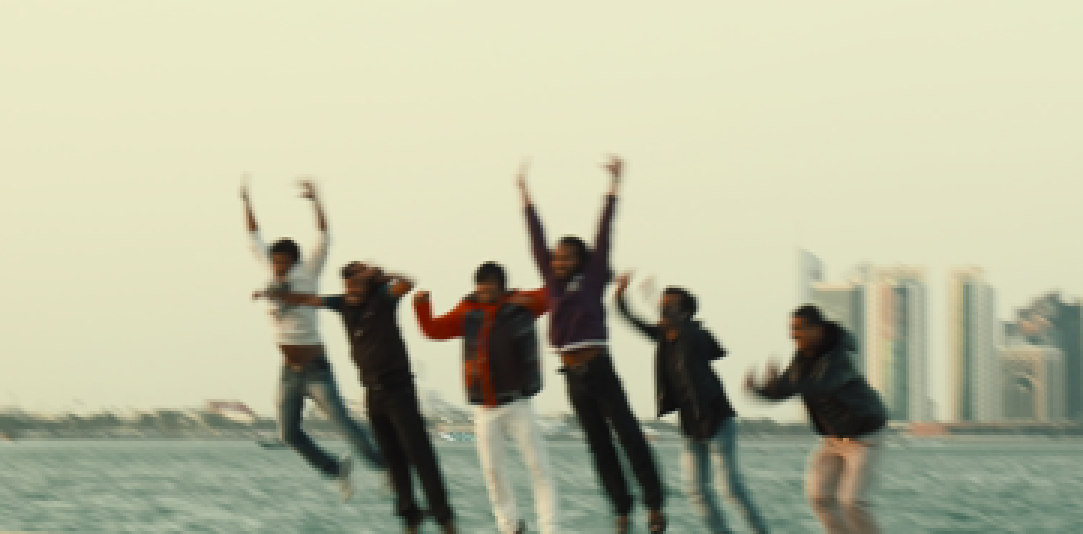}
  \end{subfigure}
  \begin{subfigure}[b]{0.195\linewidth}
     \includegraphics[width=\linewidth]{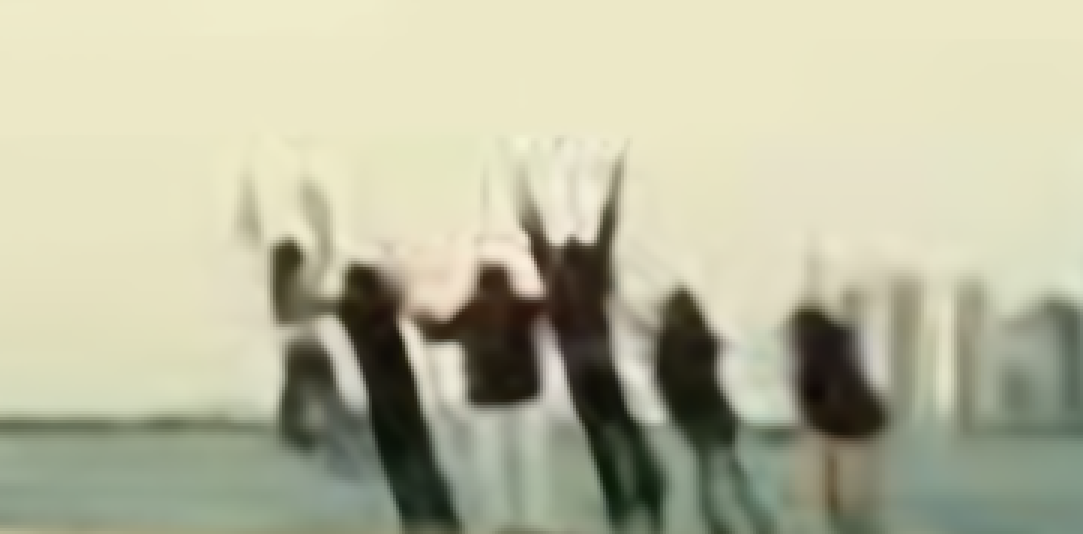}
  \end{subfigure}
  \begin{subfigure}[b]{0.195\linewidth}
  \includegraphics[width=\linewidth]{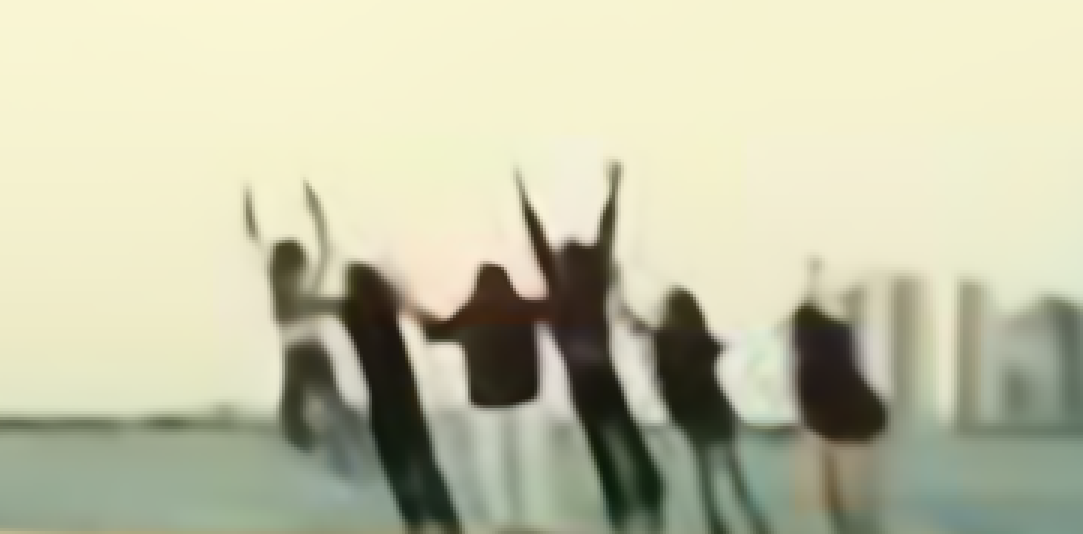}
  \end{subfigure}
  \begin{subfigure}[b]{0.195\linewidth}
  \includegraphics[width=\linewidth]{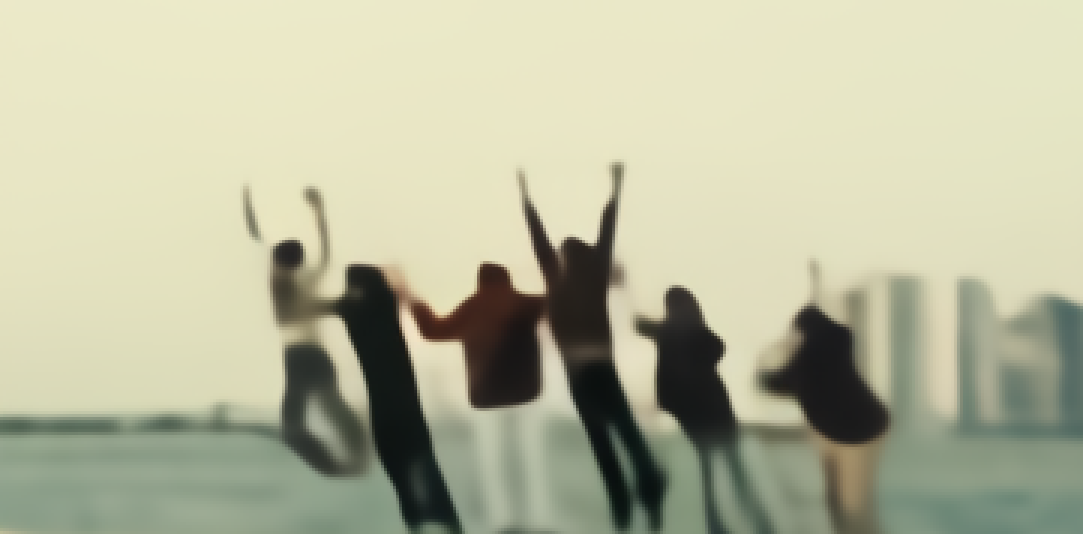}
  \end{subfigure}\vspace{1mm}
  
  
  \begin{subfigure}[b]{0.195\linewidth}
     \includegraphics[width=\linewidth]{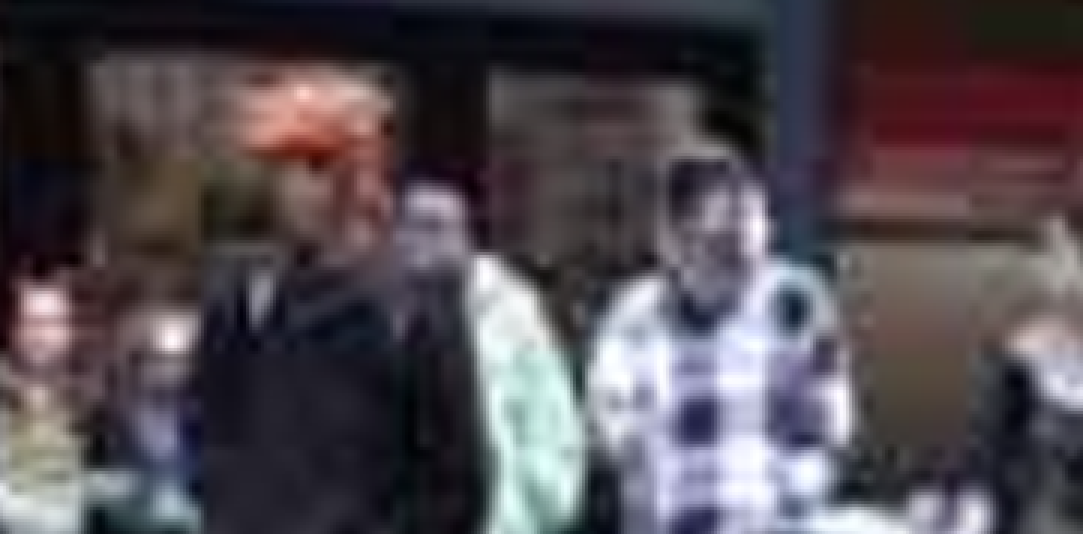}
     \caption{Overlayed LR}
  \end{subfigure}
  \begin{subfigure}[b]{0.195\linewidth}
  \includegraphics[width=\linewidth]{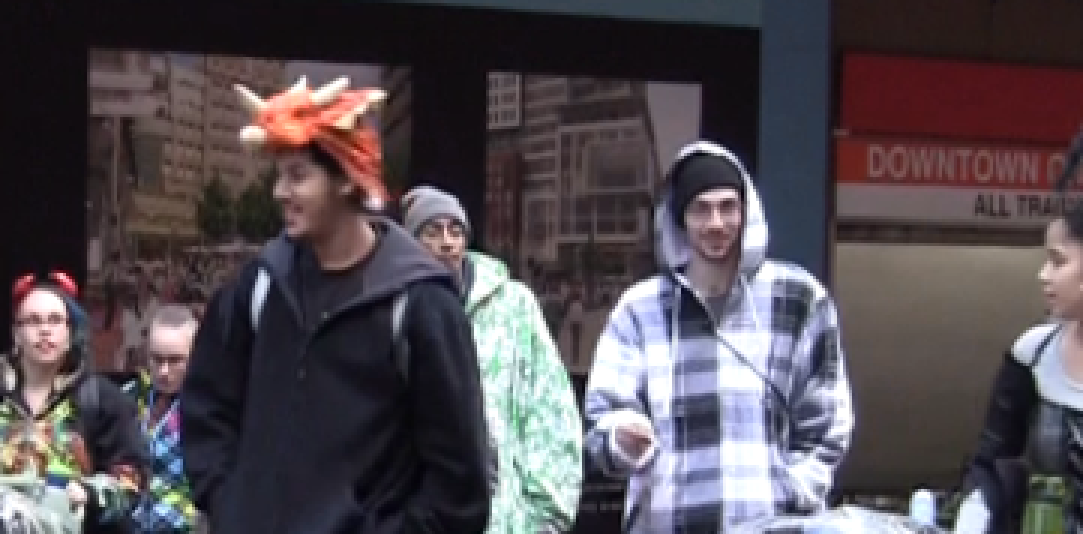}
  \caption{HR}
  \end{subfigure}
  \begin{subfigure}[b]{0.195\linewidth}
     \includegraphics[width=\linewidth]{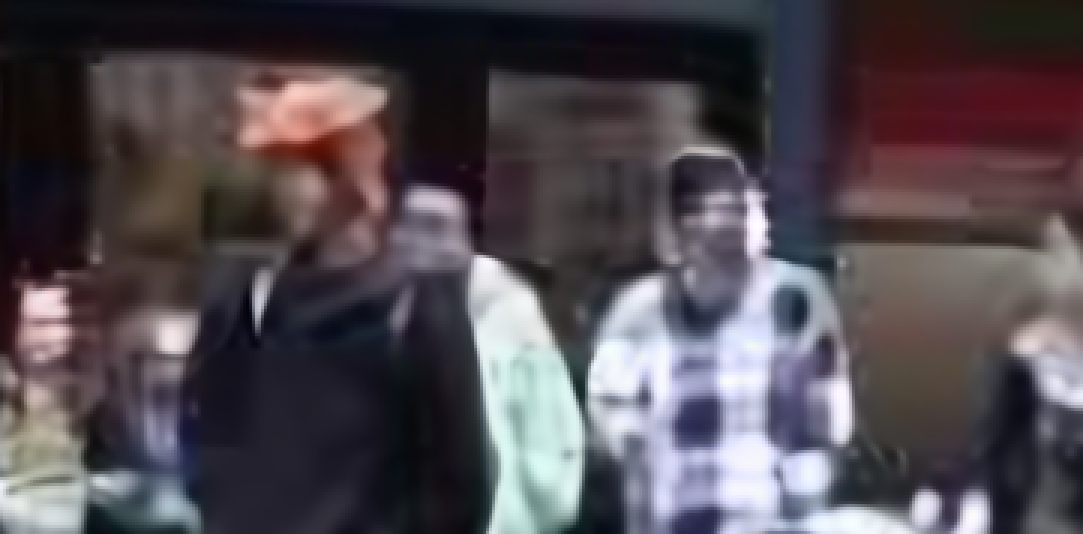}
     \caption{DNCNN+DAIN+RBPN}
  \end{subfigure}
  \begin{subfigure}[b]{0.195\linewidth}
  \includegraphics[width=\linewidth]{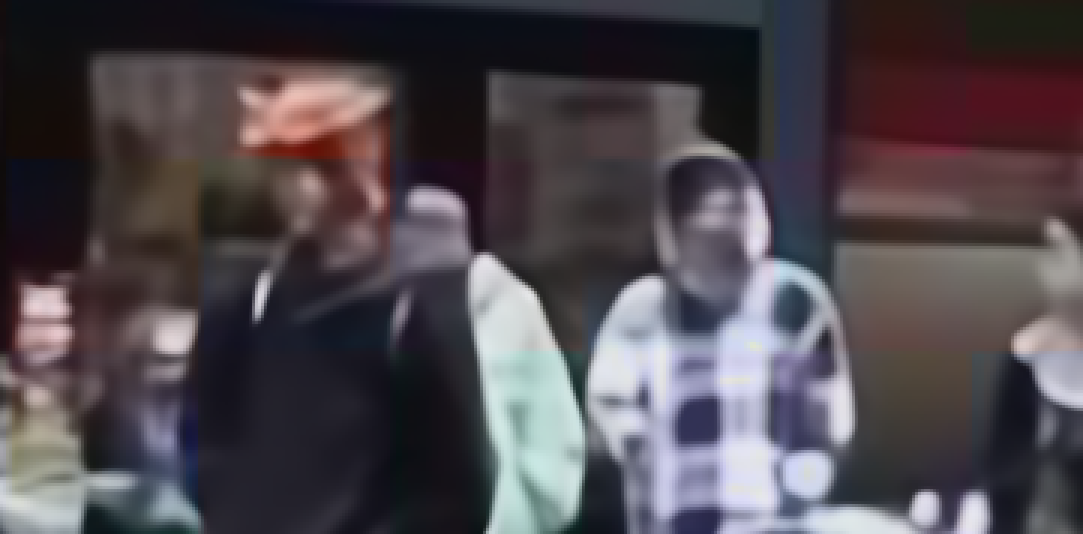}
  \caption{RNAN+DAIN+RBPN}
  \end{subfigure}
  \begin{subfigure}[b]{0.195\linewidth}
  \includegraphics[width=\linewidth]{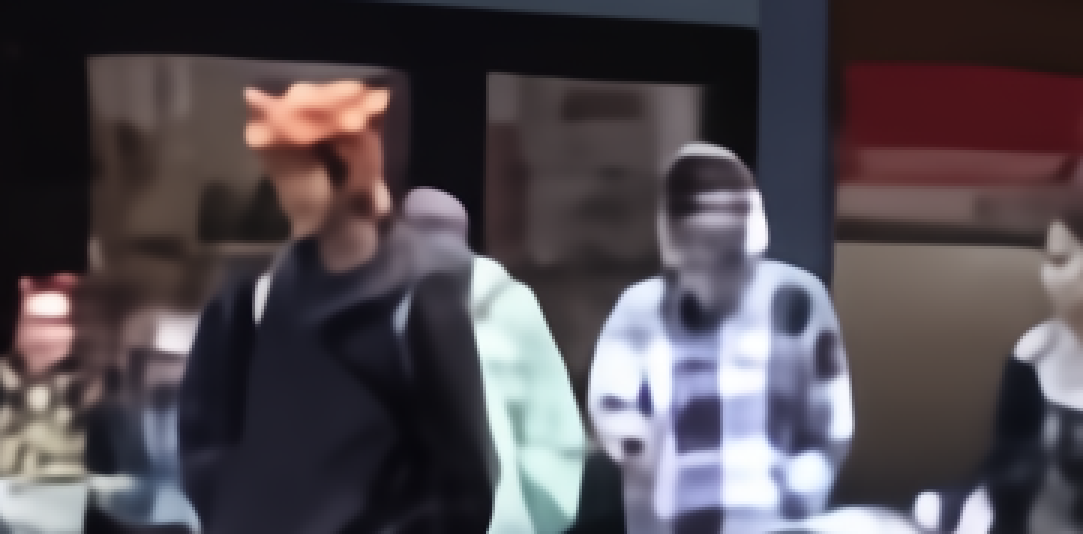}
  \caption{\textbf{ZSM (Ours)}}
  \end{subfigure}
\end{center}
\vspace{-3mm}
   \caption{Visual comparisons of different methods on compressed LR frames. The first, second, third, and fourth rows are results for QR = 10, 20, 30, and 40, respectively.
   }
 \label{fig:car}
\vspace{-5mm}
\end{figure*}

\section{Experiments and Analysis}
\subsection{Experimental Setup}


\begin{description}[style=unboxed,leftmargin=0cm]
\item[Datasets]
Vimeo-90K is used as the training set \cite{xue2019video}, including over 60,000 7-frame training video sequences. Vimeo-90K is widely used in previous VFI and VSR works \cite{bao2019memc, bao2019depth, tian2018tdan, haris2019recurrent, wang2019edvr}. Besides, Vid4~\cite{liu2011bayesian} and Vimeo testset \cite{xue2019video} are used as the evaluate datasets. To compare the performance of different methods under different motion conditions, we split the Vimeo testset into fast motion, medium motion, and slow motion sets as in ~\cite{haris2019recurrent}, including $1225$, $4977$ and $1613$ video clips, respectively. We remove $5$ videos from the original medium motion set and $3$ videos from the slow motion set, which include consecutively all-black frames that will lead to infinite values when calculating PSNR.
We generate LR frames by bicubic downsampling with factor$=4$ and use odd-indexed LR frames as inputs for predicting the corresponding consecutive HR and HFR video frames.

\vspace{1mm}
\item[Evaluation Metrics] We adopt Peak Signal-to-Noise Ratio (PSNR) and Structural Similarity Index (SSIM) \cite{wang2004image} to evaluate STVSR performance. To measure the efficiency of different methods, we also compare the model parameters and inference time on the entire Vid4 \cite{liu2011bayesian} dataset using an Nvidia Titan XP GPU.


\end{description}


\begin{table}
\caption{Quantitative comparison of our one-stage ZSM and two-stage VFI and VSR methods on Vimeo-90K~\cite{xue2019video} testset. The best two results are highlighted in \textcolor{red}{red} and \textcolor{blue}{blue} colors, respectively.}
\centering
\resizebox{1\columnwidth}{!}{
\setlength{\tabcolsep}{0.7mm}
\begin{tabular}{cc|cccccc}
\hline
\multirow{2}{*}{\begin{tabular}[c]{@{}c@{}}VFI\\   Method\end{tabular}} & \multirow{2}{*}{\begin{tabular}[c]{@{}c@{}}SR\\   Method\end{tabular}} & \multicolumn{2}{c}{Vimeo-Fast} & \multicolumn{2}{c}{Vimeo-Medium} & \multicolumn{2}{c}{Vimeo-Slow}  \\ 
                                                                        &                                                                               & PSNR          & SSIM           & PSNR           & SSIM            & PSNR          & SSIM                                                                                                \\ \hline \hline
SuperSloMo \cite{jiang2018super}                                                                 & Bicubic                                                                  &   	31.88 &   	0.8793 &   	29.94 &   	0.8477 &   	28.37 &   	0.8102                                                                      \\
SuperSloMo \cite{jiang2018super}                                                                & RCAN \cite{zhang2018image}                                                                       & 34.52         & 0.9076         & 32.50          & 0.8884          & 30.69         & 0.8624                                                                                       \\
SuperSloMo \cite{jiang2018super}                                                                & RBPN \cite{haris2019recurrent}                                                                      & 34.73         & 0.9108    &      32.79 & 	0.8930	 & 30.48 & 	0.8584         \\
SuperSloMo \cite{jiang2018super}                                                                & EDVR \cite{wang2019edvr}                                                                   & 35.05  & 	0.9136  & 	33.85  & 	0.8967  & 	30.99  & 	0.8673                                                                       \\ \hline
SepConv \cite{niklaus2017adsconv}                                                                 & Bicubic                                                                  &   	32.27 &   	0.8890 &   	30.61 &   	0.8633 &   	29.04 &   	0.8290                                                    \\
SepConv \cite{niklaus2017adsconv}                                                                & RCAN \cite{zhang2018image}                                                                       & 34.97         & 0.9195         & 33.59          & 0.9125          & 32.13         & 0.8967                                                                               \\
SepConv \cite{niklaus2017adsconv}                                                                & RBPN \cite{haris2019recurrent}                                                            & 35.07         & 0.9238         &       34.09         &      0.9229           & 32.77         & 0.9090                                                                            \\
SepConv \cite{niklaus2017adsconv}                                                                & EDVR \cite{wang2019edvr}                                                                       & 35.23         &       0.9252         & 34.22          &     0.9240            & 32.96         &          0.9112                                                                            \\
\hline
DAIN \cite{bao2019depth}                                                                   & Bicubic                                                               &   	32.41  &   	0.8910  &   	30.67  &   	0.8636  &   	29.06	  &   0.8289            \\
DAIN  \cite{bao2019depth}                                                                  & RCAN \cite{zhang2018image}                                                     & 35.27         & 0.9242         & 33.82          & 0.9146          & 32.26         & 0.8974                                                                                 \\
DAIN  \cite{bao2019depth}                                                                  & RBPN \cite{haris2019recurrent}                                                            & 35.55         & 0.9300         & 34.45          & 0.9262          & 32.92       & 0.9097                                                                                     \\
DAIN \cite{bao2019depth}                                                                   & EDVR \cite{wang2019edvr}                                                        & \textcolor{blue}{35.81}         &     \textcolor{blue}{0.9323}       & \textcolor{blue}{34.66}          &     \textcolor{blue}{0.9281}            & \textcolor{blue}{33.11}         &      \textcolor{blue}{0.9119}                                                                                         \\ \hline
 
\multicolumn{2}{c|}{ZSM (Ours)}                                                                                                                                & \textcolor{red}{36.96}         &      \textcolor{red}{0.9444}          & \textcolor{red}{35.56}          &      \textcolor{red}{0.9385}           & \textcolor{red}{33.50}         &      \textcolor{red}{0.9166}                                                                                     
 
\\ \hline
\end{tabular}
}
\vspace{-5mm}
\label{tab:result_vimeo}
\end{table}

\subsection{Space-Time Video Super-resolution}

We compare the performance of our one-stage Zooming SlowMo (ZSM) network to other two-stage methods that are composed of state-of-the-art (SOTA) VFI and VSR networks.
Three recent SOTA VFI approaches, SepConv\footnote{\url{https://github.com/sniklaus/sepconv-slomo}} \cite{niklaus2017adsconv}, Super-SloMo\footnote{Since there is no official code released, we used an unofficial PyTorch \cite{pytorch_nips} implementation from  \url{https://github.com/avinashpaliwal/Super-SloMo}.}~\cite{jiang2018super}, and DAIN\footnote{\url{https://github.com/baowenbo/DAIN}} \cite{bao2019depth}, are compared. Besides, three SOTA SR models, including a single-image SR model, RCAN\footnote{\url{https://github.com/yulunzhang/RCAN}} \cite{zhang2018image}, and two recent VSR models, RBPN\footnote{\url{https://github.com/alterzero/RBPN-PyTorch}} \cite{haris2019recurrent} and EDVR\footnote{\url{https://github.com/xinntao/EDVR}}  \cite{wang2019edvr}, are adopted to generate HR frames from both original and interpolated LR frames. 

Quantitative results on Vid4 and Vimeo testsets are shown in Tables \ref{tab:result_vid} and \ref{tab:result_vimeo}. From these tables, we can observe the following facts: (1) DAIN+EDVR is the best performing two-stage method among the compared 12 approaches; (2) VFI model matters, especially for videos with fast motion. Although RBPN and EDVR perform much better than RCAN for SR, however, when equipped with more recent SOTA VFI network DAIN, DAIN+RCAN can achieve a comparable or even better performance than SepConv+RBPN and SepConv+EDVR on the Vimeo-Fast testset; (3) VSR model also matters. For example, equipped with the same VFI network DAIN, EDVR keeps achieving better STVSR performance than other SR methods. Moreover, we can observe that our ZSM outperforms the DAIN+EDVR by $0.19$dB on Vid4, $0.25$dB on Vimeo-Slow, $0.75$dB on Vimeo-Medium, and $1.00$dB on Vimeo-Fast in terms of PSNR. Such significant improvements obtained on fast-motion videos demonstrate that our one-stage approach with simultaneously leveraging local and global temporal contexts can better handle diverse space-time patterns, including challenging large motions than two-stage methods.

\begin{table*}[htbp]
\caption{Quantitative comparisons of our results and three-stage denoising, VFI and VSR methods on video frames with noise. The best two results for each quality factor(QF) are highlighted in \textcolor{red}{red} and \textcolor{blue}{blue} colors, respectively.}
\centering
\resizebox{0.9\textwidth}{!}{
\begin{tabular}{ccccccccccc}
\hline
\multirow{2}{*}{DN Method} & \multirow{2}{*}{VFI Method} & \multirow{2}{*}{SR Method} & \multicolumn{2}{c}{Vid4} & \multicolumn{2}{c}{Vimeo-Fast} & \multicolumn{2}{c}{Vimeo-Medium} & \multicolumn{2}{c}{Vimeo-Slow}  \\
                                &                             &                            & PSNR  & SSIM             & PSNR  & SSIM                   & PSNR  & SSIM                     & PSNR  & SSIM                    \\ \hline 
                                \hline
\multirow{4}{*}{ToFlow~\cite{xue2019video}}         & SepConv \cite{niklaus2017adsconv}                     & RBPN \cite{haris2019recurrent}             & 22.78 & 0.5692           & 29.03 & 0.8376                 & 28.46 & 0.8140                   & 27.54 & 0.7826                  \\
                                & SepConv \cite{niklaus2017adsconv}                     &  EDVR \cite{wang2019edvr}                       & 22.79 & 0.5692           & 29.02 & 0.8370                 & 28.44 & 0.8131                   & 27.52 & 0.7819                  \\
                                & DAIN \cite{bao2019depth}                         & RBPN \cite{haris2019recurrent}             & 22.79 & 0.5687           & \textcolor{blue}{29.08} & \textcolor{blue}{0.8388}                 & \textcolor{blue}{28.50} & 0.8144                   & \textcolor{blue}{27.56} & \textcolor{blue}{0.7830}                  \\
                                & DAIN \cite{bao2019depth}                         &  EDVR \cite{wang2019edvr}                       & \textcolor{blue}{22.80} & \textcolor{blue}{0.5693}           & 29.08 & 0.8387                 & 28.49 & \textcolor{blue}{0.8144}                   & 27.55 & 0.7825                  \\
                                \hline
\multicolumn{3}{c}{ZSM (Ours)}                                                             & \textcolor{red}{23.91} & \textcolor{red}{0.6514}           & \textcolor{red}{31.49} & \textcolor{red}{0.8446}                 & \textcolor{red}{30.49} & \textcolor{red}{0.8594}                   & \textcolor{red}{29.36} & \textcolor{red}{0.8321}                  \\
\hline
\end{tabular}
}
\label{tbl:noise}
\end{table*}

Furthermore, we also compare the efficiency of different networks and show their model sizes and runtime in Table \ref{tab:result_vid}. To synthesize high-quality frames, SOTA VFI and VSR networks usually come with very large frame reconstruction modules. As a result, the composed two-stage SOTA STVSR networks will have a large number of parameters. Our one-stage model is with only one frame reconstruction module, thus has much fewer parameters than the SOTA two-stage networks. Table \ref{tab:result_vid} shows that our ZSM is more than $4\times$ and $3\times$ smaller than the DAIN+EDVR and DAIN+RBPN, respectively. In terms of runtime, the smaller model size makes our network more than $8\times$ faster than DAIN+RBPN and $3\times$ faster than the DAIN+EDVR. Our method is still more than $2\times$ faster compared to two-stage methods with a fast VFI network: SuperSlowMo. These results can validate the superiority of our one-stage ZSM model in terms of efficiency.



Qualitative results of these different methods are illustrated in Figure~\ref{fig:stvsrresult}. Our method demonstrates noticeably perceptual improvements over other compared two-stage methods. Clearly, our proposed network can synthesize visually appealing HR video frames with more accurate structures, more fine details, and fewer blurry artifacts, even for challenging video sequences with large motions. We can also observe that the SOTA VFI methods: SepConv and DAIN fail to handle fast motions. Consequently, the composed two-stage methods tend to generate severe motion blurs in output frames. In the proposed one-stage framework, we simultaneously learn temporal and spatial SR by exploring the intra-relatedness within natural videos. Thus, our proposed framework: ZSM can well address the large motion issue in temporal SR even with a much smaller model. 

\begin{table*}[htbp]
\caption{Quantitative comparison of our results and three-stage CAR, VFI and VSR methods on compressed testsets (QF = 10, 20, 30, and 40). The best two results are highlighted in \textcolor{red}{red} and \textcolor{blue}{blue} colors, respectively. }
\centering
\resizebox{0.9\textwidth}{!}{
\begin{tabular}{c|ccccccccccc}
\hline
\multirow{2}{*}{QF} & \multirow{2}{*}{CAR Method} & \multirow{2}{*}{VFI Method} & \multirow{2}{*}{SR Method} & \multicolumn{2}{c}{Vid4} & \multicolumn{2}{c}{Vimeo-Fast} & \multicolumn{2}{c}{Vimeo-Medium} & \multicolumn{2}{c}{Vimeo-Slow}  \\
                                &                            &                 &           & PSNR  & SSIM             & PSNR  & SSIM                   & PSNR  & SSIM                     & PSNR  & SSIM                    \\ \hline
                                \hline
\multirow{9}{*}{10} & \multirow{4}{*}{DNCNN~\cite{zhang2017beyond}}          & SepConv \cite{niklaus2017adsconv}                     & RBPN \cite{haris2019recurrent}             & 21.33 & 0.4659           & 27.84 & 0.7917                 & 26.59 & 0.7475                   & 25.46 & 0.7010                  \\
                            &    & SepConv \cite{niklaus2017adsconv}                     &  EDVR \cite{wang2019edvr}                       & 21.31 & 0.4647           & 27.83 & 0.7909                 & 26.58 & 0.7466                   & 25.45 & 0.7001                  \\
                              &  & DAIN \cite{bao2019depth}                         & RBPN \cite{haris2019recurrent}             & \textcolor{blue}{21.37} & 0.4688           & 27.92 & \textcolor{blue}{0.7958}                 & \textcolor{blue}{26.65} & 0.7510                   & \textcolor{blue}{25.49} & 0.7028                  \\
                              &  & DAIN \cite{bao2019depth}                         &  EDVR \cite{wang2019edvr}                       & 21.35 & 0.4679           & \textcolor{blue}{27.92} & 0.7955                 & 26.64 & 0.7504                   & 25.48 & 0.7020                  \\
                                \cline{2-12}
 & \multirow{4}{*}{RNAN~\cite{zhang2019residual}}           & SepConv \cite{niklaus2017adsconv}                     & RBPN \cite{haris2019recurrent}             &  20.41 & \textcolor{blue}{0.4864} & 23.80 & 0.7926 & 23.61 & 0.7607 & 22.83 & 0.7171  \\
                        &        & SepConv \cite{niklaus2017adsconv}                     &  EDVR \cite{wang2019edvr}                       &   20.40 & 0.4858 & 23.80 & 0.7925 & 23.60 & 0.7604 & 22.82 & 0.7165  \\
                         &       & DAIN \cite{bao2019depth}                         & RBPN \cite{haris2019recurrent}             &  20.41 & 0.4859 & 23.81 & 0.7936 & 23.60 & \textcolor{blue}{0.7610} & 22.82 & \textcolor{blue}{0.7172}  \\
                          &      & DAIN \cite{bao2019depth}                         &  EDVR \cite{wang2019edvr}                       & 20.41 & 0.4856 & 23.81 & 0.7936 & 23.59 & 0.7609 & 22.81 & 0.7167          \\ 
                                \cline{2-12}
 & \multicolumn{3}{c}{ZSM (Ours)}                                                         & \textcolor{red}{22.03} & \textcolor{red}{0.5216}           & \textcolor{red}{29.52} & \textcolor{red}{0.8367}                 & \textcolor{red}{28.13} & \textcolor{red}{0.8009}                   & \textcolor{red}{26.64} & \textcolor{red}{0.7532}                  \\ \hline
\hline
\multirow{9}{*}{20} & \multirow{4}{*}{DNCNN~\cite{zhang2017beyond}}      & SepConv \cite{niklaus2017adsconv}                     & RBPN \cite{haris2019recurrent}             & 22.10 & 0.5102           & 29.36 & 0.8232                 & 28.03 & 0.7834                   & 26.76 & 0.7396                  \\
                        &    & SepConv \cite{niklaus2017adsconv}                     &  EDVR \cite{wang2019edvr}                       & 22.09 & 0.5089           & 29.36 & 0.8225                 & 28.02 & 0.7825                   & 26.75 & 0.7386                  \\
                        &    & DAIN \cite{bao2019depth}                         & RBPN \cite{haris2019recurrent}             & \textcolor{blue}{22.14} & 0.5132           & 29.46 & \textcolor{blue}{0.8266}                 & \textcolor{blue}{28.10} & 0.7865                   & \textcolor{blue}{26.79} & 0.7411                  \\
                        &    & DAIN \cite{bao2019depth}                         &  EDVR \cite{wang2019edvr}                       & 22.13 & 0.5121           & \textcolor{blue}{29.46} & 0.8263                 & 28.09 & 0.7859                   & 26.78 & 0.7403                  \\
                            \cline{2-12}
& \multirow{4}{*}{RNAN~\cite{zhang2019residual}}       & SepConv \cite{niklaus2017adsconv}                     & RBPN \cite{haris2019recurrent}             & 20.91 & 0.5292 & 24.93 & 0.8197 & 24.76 & 0.7920 & 24.01 & 0.7527  \\
                         &   & SepConv \cite{niklaus2017adsconv}                     &  EDVR \cite{wang2019edvr}                       & 20.90 & 0.5285 & 24.92 & 0.8196 & 24.76 & 0.7917 & 24.01 & 0.7524  \\
                          &  & DAIN \cite{bao2019depth}                         & RBPN \cite{haris2019recurrent}             & 20.93 & \textcolor{blue}{0.5293} & 24.95 & 0.8213 & 24.77 & \textcolor{blue}{0.7927} & 24.00 & \textcolor{blue}{0.7530}  \\
                          &  & DAIN \cite{bao2019depth}                         &  EDVR \cite{wang2019edvr}                       & 20.93 & 0.5290 & 24.95 & 0.8214 & 24.76 & 0.7927 & 24.00 & 0.7528   \\
                            \cline{2-12}
& \multicolumn{3}{c}{ZSM (Ours)}                                                          & \textcolor{red}{22.78} & \textcolor{red}{0.5707}           & \textcolor{red}{31.00} & \textcolor{red}{0.8607}                 & \textcolor{red}{29.54} & \textcolor{red}{0.8300}                   & \textcolor{red}{27.95} & \textcolor{red}{0.7864}       \\ \hline
\hline
\multirow{9}{*}{30} & \multirow{4}{*}{DNCNN~\cite{zhang2017beyond}}      & SepConv \cite{niklaus2017adsconv}                     & RBPN \cite{haris2019recurrent}             & 22.52 & 0.5361           & 30.08 & 0.8377                 & 28.73 & 0.8009                   & 27.43 & 0.7593                  \\
                       &     & SepConv \cite{niklaus2017adsconv}                     &  EDVR \cite{wang2019edvr}                       & 22.51 & 0.5354           & 30.09 & 0.8372                 & 28.72 & 0.8000                   & 27.41 & 0.7583                  \\
                        &    & DAIN \cite{bao2019depth}                         & RBPN \cite{haris2019recurrent}             & \textcolor{blue}{22.57} & 0.5393           & 30.19 & \textcolor{blue}{0.8410}                 & \textcolor{blue}{28.81} & 0.8039                   & \textcolor{blue}{27.46} & 0.7608                  \\
                       &     & DAIN \cite{bao2019depth}                         &  EDVR \cite{wang2019edvr}                       & 22.56 & 0.5388           & \textcolor{blue}{30.20} & 0.8401                 & 28.80 & 0.8033                   & 27.44 & 0.7598                  \\
                            \cline{2-12}
& \multirow{4}{*}{RNAN~\cite{zhang2019residual}} & SepConv~\cite{niklaus2017adsconv} & RBPN~\cite{haris2019recurrent} & 21.41 & 0.5569 & 25.54 & 0.8345 & 25.35 & 0.8084 & 24.64 & 0.7717  \\
                                                      &          & SepConv~ \cite{niklaus2017adsconv} & EDVR \cite{wang2019edvr}       & 21.41 & 0.5567 & 25.54 & 0.8344 & 25.34 & 0.8082 & 24.63 & 0.7715  \\
                                                       &         & DAIN \cite{bao2019depth}           & RBPN \cite{haris2019recurrent} & 21.45 & 0.5576 & 25.58 & 0.8365 & 25.36 & 0.8095 & 24.63 & \textcolor{blue}{0.7721}  \\
                                                       &         & DAIN \cite{bao2019depth}           & EDVR \cite{wang2019edvr}       & 21.44 & \textcolor{blue}{0.5577} & 25.58 & 0.8367 & 25.36 & \textcolor{blue}{0.8095} & 24.63 & 0.7720     \\
                            \cline{2-12}
 & \multicolumn{3}{c}{ZSM (Ours)}                                                          & \textcolor{red}{23.25} & \textcolor{red}{0.6013}           & \textcolor{red}{31.81} & \textcolor{red}{0.8733}                 & \textcolor{red}{30.30} & \textcolor{red}{0.8082}                   & \textcolor{red}{28.65} & \textcolor{red}{0.8042}          \\ \hline
\hline
\multirow{9}{*}{40} &\multirow{4}{*}{DNCNN~\cite{zhang2017beyond}}      & SepConv \cite{niklaus2017adsconv}                     & RBPN \cite{haris2019recurrent}             & 22.83 & 0.5559           & 30.51 & 0.8464                 & 29.17 & 0.8120                   & 27.84 & 0.7720                  \\
                        &    & SepConv \cite{niklaus2017adsconv}                     &  EDVR \cite{wang2019edvr}                       & 22.83 & 0.5560           & 30.51 & 0.8458                 & 29.16 & 0.8110                   & 27.83 & 0.7710                  \\
                         &   & DAIN \cite{bao2019depth}                         & RBPN \cite{haris2019recurrent}             & 22.88 & 0.5590           & 30.63 & \textcolor{blue}{0.8497}                 & \textcolor{blue}{29.26} & 0.8149                   & \textcolor{blue}{27.88} & 0.7734                  \\
                         &   & DAIN \cite{bao2019depth}                         &  EDVR \cite{wang2019edvr}                       & \textcolor{blue}{22.88} & 0.5592           & \textcolor{blue}{30.64} & 0.8496                 & 29.25 & 0.8144                   & 27.86 & 0.7725                  \\
                            \cline{2-12}
 & \multirow{4}{*}{RNAN \cite{zhang2019residual}} & SepConv \cite{niklaus2017adsconv}                     & RBPN \cite{haris2019recurrent}   & 21.63 & 0.5764 & 25.87 & 0.8426 & 25.71 & 0.8190 & 25.01 & 0.7837  \\
                &      & SepConv \cite{niklaus2017adsconv} &  EDVR \cite{wang2019edvr} & 21.63 & 0.5764 & 25.87 & 0.8425 & 25.70 & 0.8187 & 25.01 & 0.7836  \\
                &      & DAIN \cite{bao2019depth}                         & RBPN \cite{haris2019recurrent} & 21.67 & 0.5774 & 25.92 & 0.8449 & 25.73 & 0.8202 & 25.01 & 0.7842  \\
                &      & DAIN \cite{bao2019depth}                         &  EDVR \cite{wang2019edvr} & 21.67 & \textcolor{blue}{0.5779} & 25.92 & 0.8451 & 25.73 & \textcolor{blue}{0.8203} & 25.01 & \textcolor{blue}{0.7842} \\
                            \cline{2-12}
 & \multicolumn{3}{c}{ZSM (Ours)}                                                         & \textcolor{red}{23.59} & \textcolor{red}{0.6226}           & \textcolor{red}{32.40} & \textcolor{red}{0.8818}                 & \textcolor{red}{30.85} & \textcolor{red}{0.8560}                   & \textcolor{red}{29.15} & \textcolor{red}{0.8164}                  \\
\hline
\end{tabular}
}
\label{tab:car}
\end{table*}
\subsection{Noisy Space-Time Video Super-resolution}
\label{subsec:NSTVSR}
Real-world videos are usually compressed and come with complicated noise. To further validate the robustness of the proposed space-time video super-resolution method on noisy data, we add random noise and JPEG compression artifacts into LR video frames~\cite{xiang2020boosting}, respectively. 

\subsubsection{Random Noise}
\label{subsubsec:RN}
In our experiments, we train our model on the Vimeo-90K dataset with a mixture of Gaussian noise and 10\% salt-and-pepper noise is added to input LR frames as in ToFlow~\cite{xue2019video}. We compare our method with four models: ToFlow+SepConv+RBPN, Toflow+SepConv+EDVR, Toflow+DAIN+RBPN, and Toflow+DAIN+EDVR on Vid4, Vimeo-fast, Vimeo-Medium, and Vimeo-Slow datasets. Here, Toflow is used for denoising in the compared methods. The quantitative and qualitative results are shown in Table~\ref{tbl:noise} and Figure~\ref{fig:noise}, respectively.

From Table~\ref{tbl:noise}, we can see that our one-stage Zooming SlowMo (ZSM) achieves the best performance among all compared approaches in terms of four evaluation datasets. Our method outperforms the best three-stage method by $1.11$ dB on Vid4, $2.41$ dB on Vimeo-Fast, $2.00$ dB on Vimeo-Medium, and $1.81$ dB on Vimeo-Slow in terms of PSNR.
This trend is more obvious in Visual results. From Figure~\ref{fig:noise}, we observe that all three methods can effectively remove severe noises in input LR frames while the reconstructed HR frames by our model have more visual details and fewer artifacts. The results demonstrate that our one-stage network is capable of handling STVSR even for noisy inputs.

\begin{figure*}[t!]
\captionsetup[subfigure]{labelformat=empty}
\begin{center}
  \begin{subfigure}[b]{0.195\linewidth}
     \includegraphics[width=\linewidth]{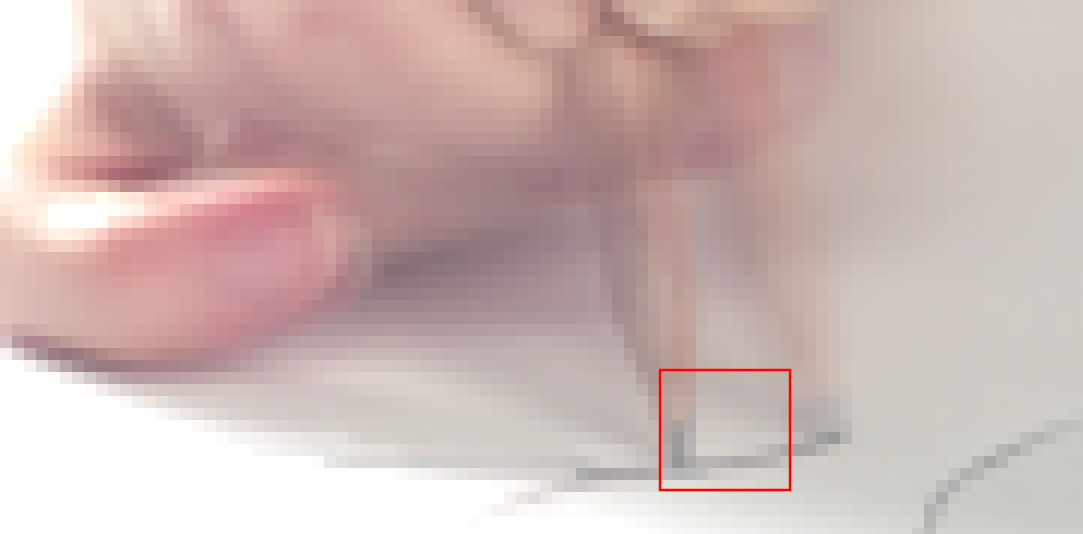}
  \end{subfigure}
  \begin{subfigure}[b]{0.195\linewidth}
  \includegraphics[width=\linewidth]{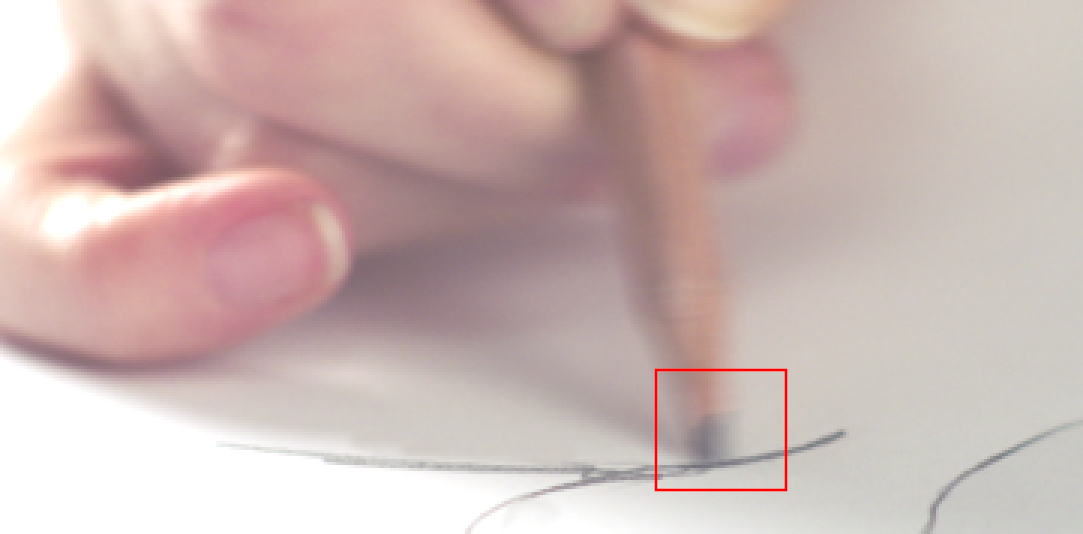}
  \end{subfigure}
  \begin{subfigure}[b]{0.195\linewidth}
     \includegraphics[width=\linewidth]{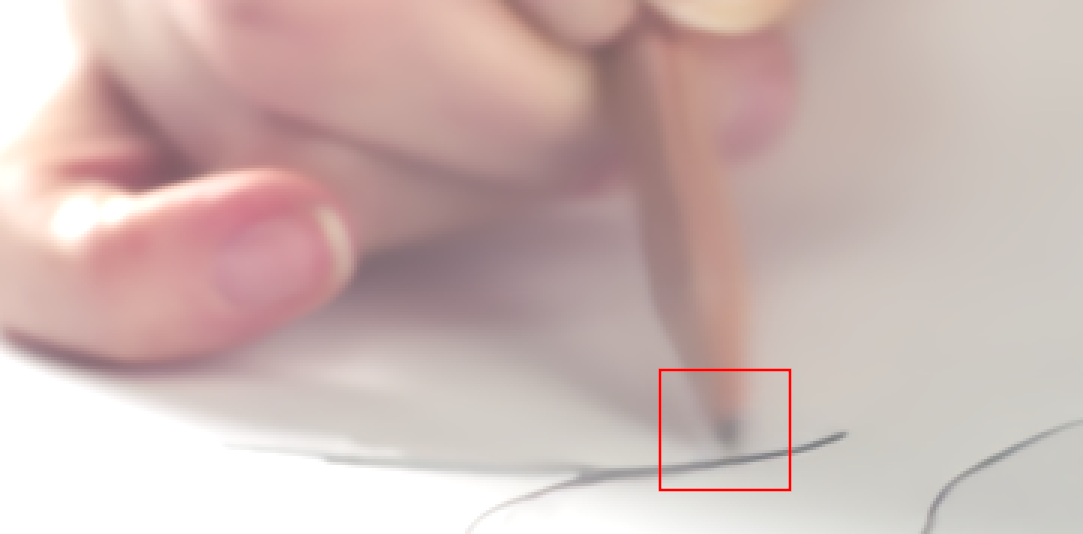}
  \end{subfigure}
  \begin{subfigure}[b]{0.195\linewidth}
  \includegraphics[width=\linewidth]{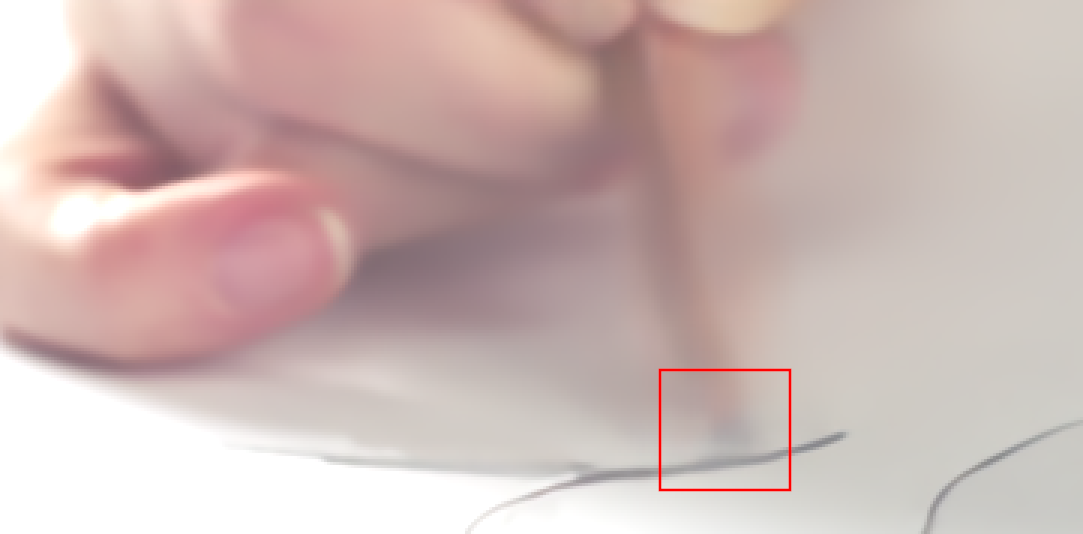}
  \end{subfigure}
  \begin{subfigure}[b]{0.195\linewidth}
  \includegraphics[width=\linewidth]{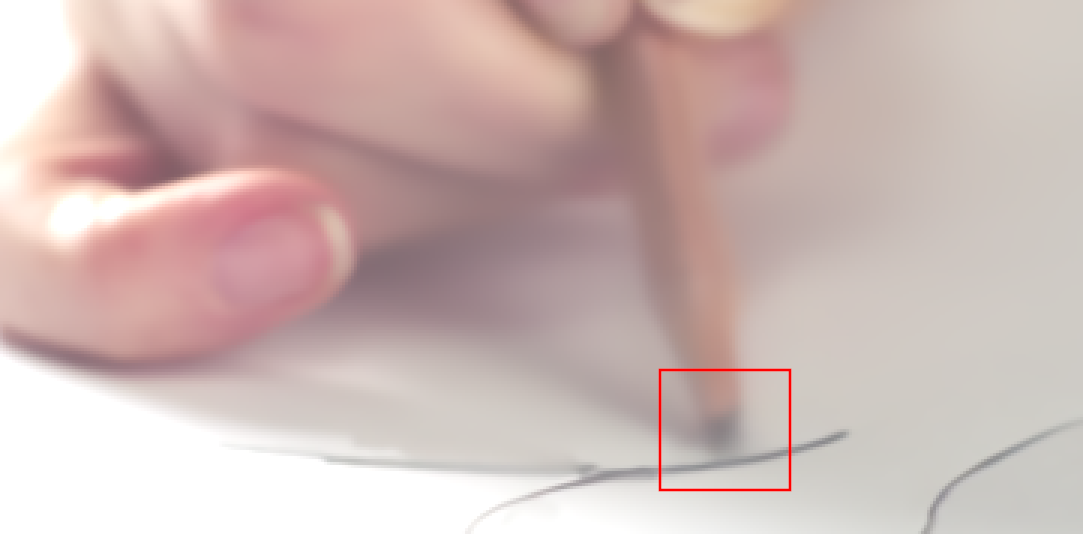}
  \end{subfigure}
  \vspace{1mm}
  
  \begin{subfigure}[b]{0.195\linewidth}
     \includegraphics[width=\linewidth]{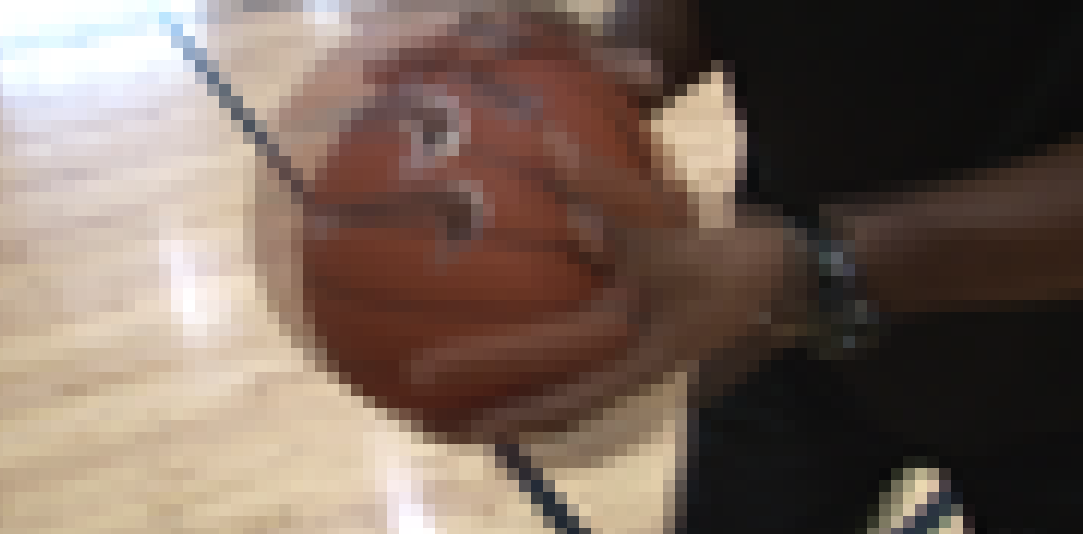}
     \subcaption{Overlayed LR}
  \end{subfigure}
  \begin{subfigure}[b]{0.195\linewidth}
  \includegraphics[width=\linewidth]{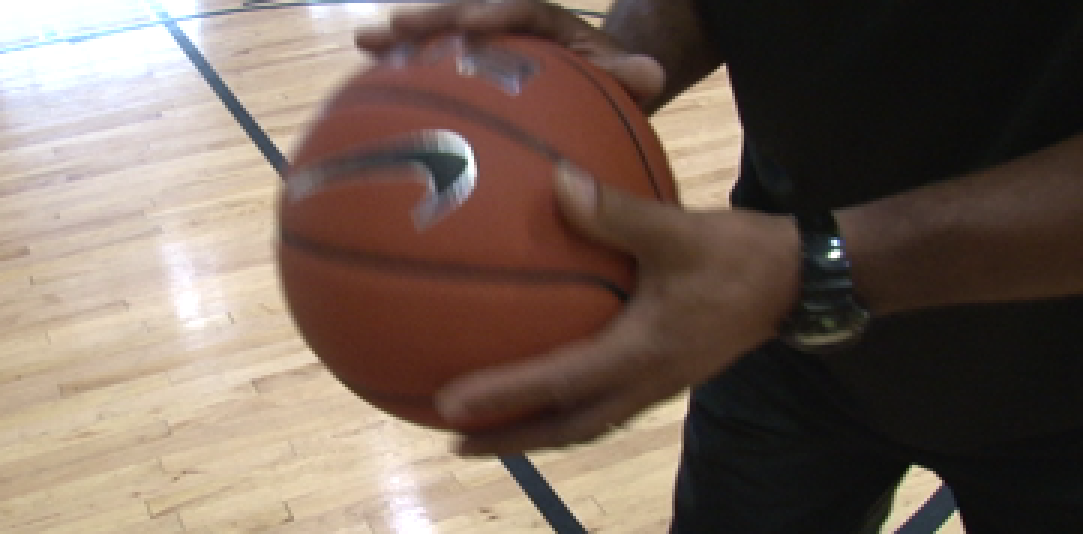}
  \subcaption{HR}
  \end{subfigure}
  \begin{subfigure}[b]{0.195\linewidth}
     \includegraphics[width=\linewidth]{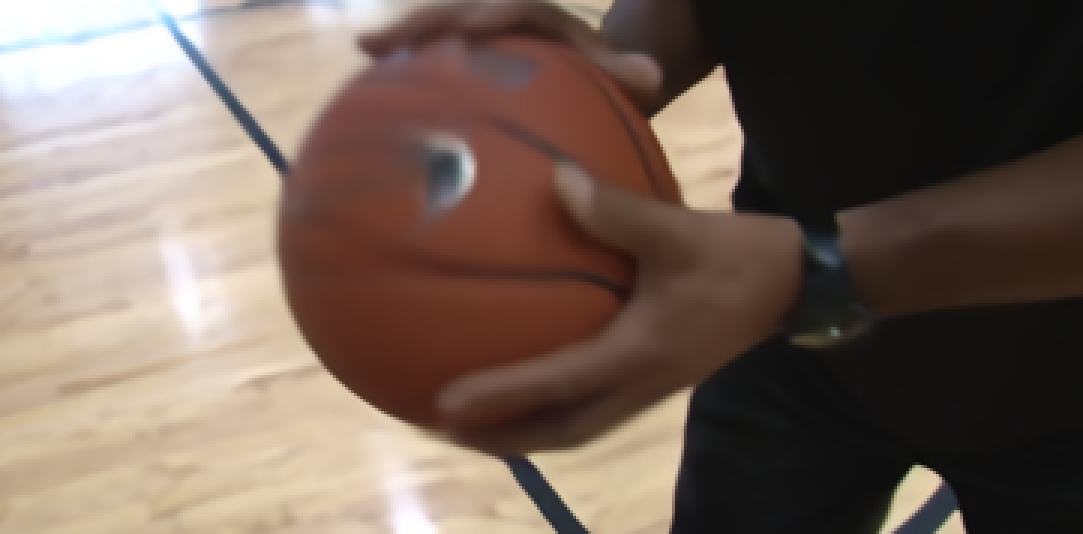}
     \subcaption{w/ DFI}
  \end{subfigure}
  \begin{subfigure}[b]{0.195\linewidth}
  \includegraphics[width=\linewidth]{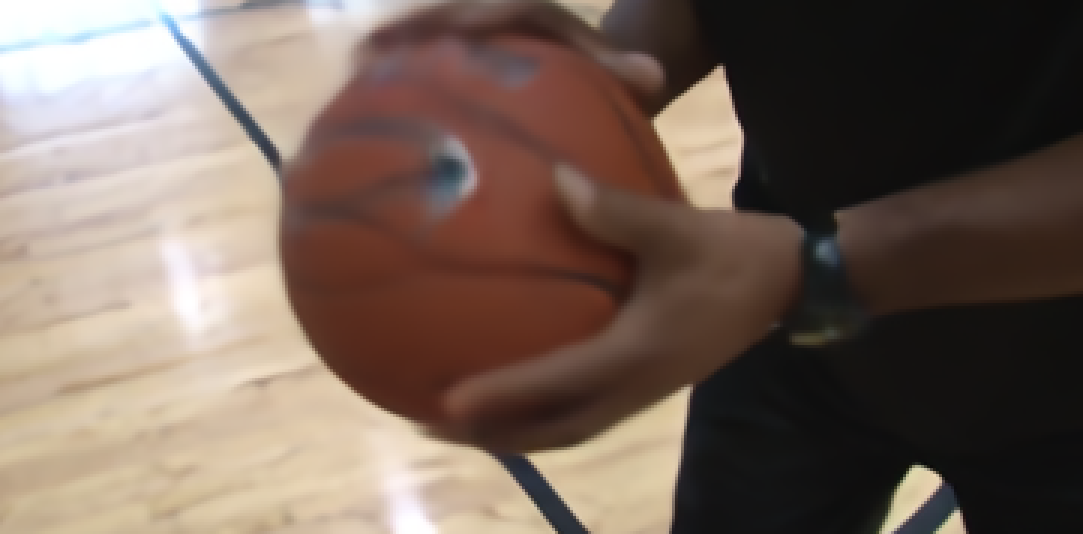}
  \subcaption{w/ DFI+ConvLSTM}
  \end{subfigure}
  \begin{subfigure}[b]{0.195\linewidth}
  \includegraphics[width=\linewidth]{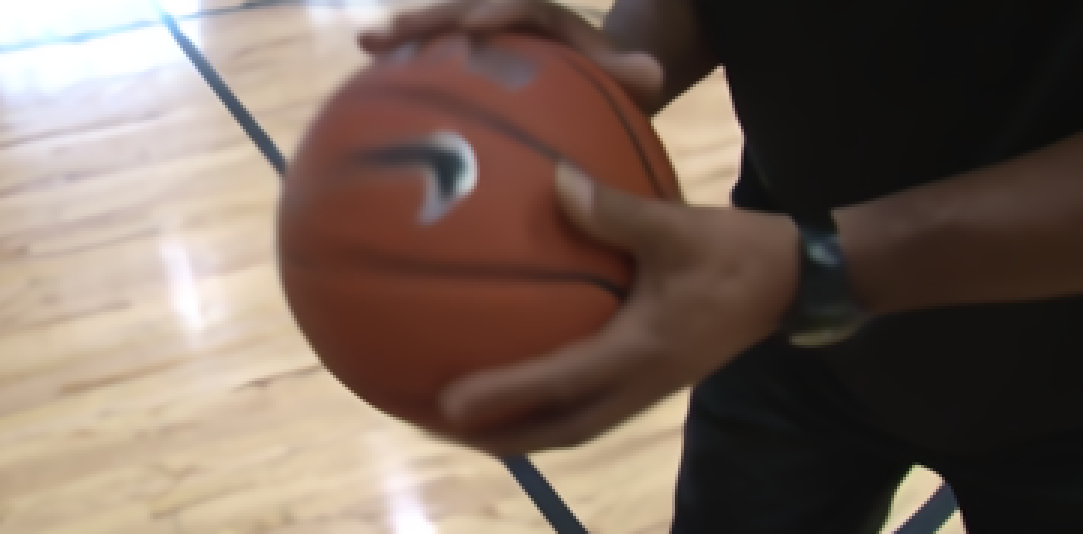}
  \subcaption{w/ DFI+DConvLSTM}
  \end{subfigure}
\end{center}
   \caption{Ablation study on Deformable ConvLSTM (DConvLSTM). Vanilla ConvLSTM will fail on videos with fast motions. Embedded with state updating cells, the proposed DConvLSTM is more capable of leveraging global temporal contexts for reconstructing more accurate content, even for fast-motion videos.
   }
 \label{fig:ablation_dconvlstm}
\vspace{-6mm}
\end{figure*}

\begin{figure}
\captionsetup[subfigure]{labelformat=empty}
\begin{center}
  \begin{subfigure}[b]{0.32\linewidth}
     \includegraphics[width=\linewidth]{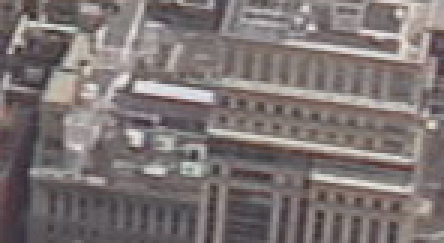}
     \subcaption{HR}
  \end{subfigure}
  \begin{subfigure}[b]{0.32\linewidth}
  \includegraphics[width=\linewidth]{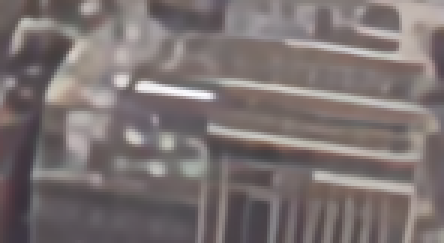}
  \subcaption{w/o bidirectional}
  \end{subfigure}
  \begin{subfigure}[b]{0.32\linewidth}
     \includegraphics[width=\linewidth]{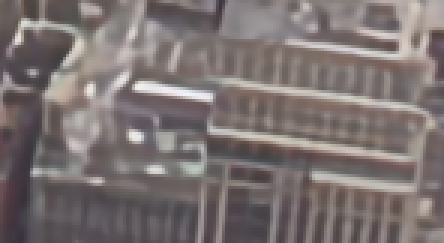}
     \subcaption{w/ bidirectional}
  \end{subfigure}
\end{center}
\vspace{-3mm}
   \caption{Ablation study on the bidirectional mechanism in DConvLSTM. By adding the bidirectional mechanism into DConvLSTM, our model can utilize both previous and future contexts, and therefore can reconstruct more visually appealing frames with finer details, especially for video frames at the first step, which cannot access any temporal information from preceding frames.}
 \label{fig:ablation_bidirectional}
\end{figure}

\begin{figure}
\captionsetup[subfigure]{labelformat=empty}
\begin{center}
  \begin{subfigure}[b]{0.42\linewidth}
     \includegraphics[width=\linewidth]{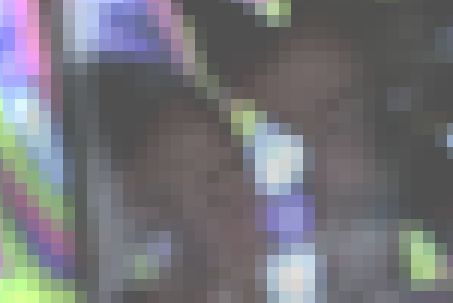}
     \subcaption{Overlayed LR}
  \end{subfigure}
  \begin{subfigure}[b]{0.42\linewidth}
  \includegraphics[width=\linewidth]{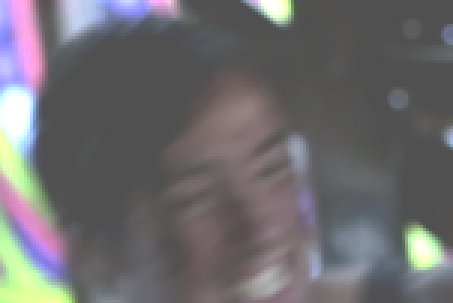}
  \subcaption{HR}
  \end{subfigure}
  
  \begin{subfigure}[b]{0.42\linewidth}
     \includegraphics[width=\linewidth]{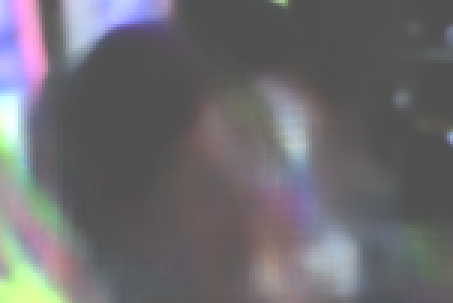}
     \subcaption{w/o DFI@model (a)}
  \end{subfigure}
  \begin{subfigure}[b]{0.42\linewidth}
  \includegraphics[width=\linewidth]{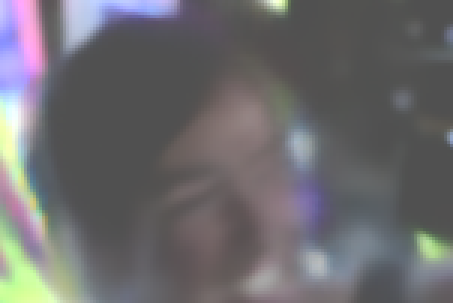}
  \subcaption{w/ DFI@model (b)}
  \end{subfigure}
\end{center}
\vspace{-3mm}
   \caption{Ablation study on feature interpolation. The naive feature interpolation model without deformable sampling will obtain overly smooth results for videos with fast motions. With the proposed deformable feature interpolation (DFI), our model can well exploit local contexts in adjacent frames, thus is more effective in handling large motions.}
 \label{fig:ablation_interp}
\vspace{-6mm}
\end{figure}
\subsubsection{JPEG Compression Artifact}
\label{subsubsec:CAR}
For this setting, we train our model on the Vimeo-90K dataset with JPEG compression of different quality factors (QF = 10, 20, 30, and 40). We compare our model with two compression artifact reduction (CAR) methods: DNCNN~\cite{zhang2017beyond} and RNAN \cite{zhang2019residual} +DAIN+RBPN on Vid4~\cite{liu2011bayesian} and three subsets of Vimeo90K's testset. The quantitative results and visual comparisons for each quality factor are shown in Table~\ref{tab:car}
and Figure~\ref{fig:car}, respectively.

It is shown in Table~\ref{tab:car} that our one-stage Zooming SlowMo (ZSM) outperforms the existing three-stage methods for all QFs significantly, and yielding the highest overall PNSR and SSIM across all testsets, especially on Vimeo-fast that is with large motions: our method exceeds the second-best result by $1.60$ dB for QF=10 test data, $1.54$ dB for QF=20, $1.61$ dB for QF=30 and $1.76$ dB for QF=40.
Figure~\ref{fig:car} provides some qualitative examples. We can assess the quality of the output images from the following perspectives: compression artifact reductions (e.g. blocking and ringing artifacts, color shift), and the reconstruction of details. According to Figure~\ref{fig:car}, even the other methods can reduce the compression artifacts in most cases, they suffer from over-smooth and lack of details. Compared with them, our output is with fewer ringing and blocking artifacts and sharper edges under different QFs. These results demonstrate that our one-stage framework can reach a better balance between artifact reduction and high-fidelity reconstruction. It is worth noting that our method is capable of restoring the color aberrations in the images compressed by low quality factors (see the top 2 rows of Figure~\ref{fig:car}). 

\subsection{Ablation Study}
In previous sections, we have already illustrated the superiority of our proposed one-stage framework. In this section, we make a comprehensive ablation study to further demonstrate the effectiveness of different modules in our network. 

\begin{table}
\caption{Ablation study on the proposed modules. While ConvLSTM performs worse for fast-motion videos, our deformable feature interpolation and deformable ConvLSTM can effectively handle motions and improve overall STVSR performance.}
\centering
\resizebox{0.49\textwidth}{!}{
\setlength{\tabcolsep}{0.7mm}
\begin{tabular}{l|cc|cccc}
\hline
Method                                                                        & (a) & (b) & (c)   & (d)     & (e) & (f) \\ \hline \hline
Naive feature interpolation                                                                  &     $\surd$  &      &    &                  &           &              \\
DFI     &      & $\surd$      & $\surd$      & $\surd$      & $\surd$    & $\surd$          \\ \hline
ConvLSTM                                                                      &            &          & $\surd$      &          &        &                \\
DConvLSTM                                                         &           &          &          & $\surd$      &       &           \\
Bidirectional DConvLSTM                                                             &      &          &          &          & $\surd$    &    $\surd$ \\
GFI                                                  &      &          &          &          &          &   $\surd$    \\
 \hline \hline
Vid4 (slow motion)                                                                    &   25.18        & 25.34 & 25.68 & 26.18 & 26.31   & 26.49      \\
\hline
Vimeo-Fast (fast motion)                                                                     &     34.93 &  35.66 &  35.39 & 36.56 & 36.81 & 36.96\\ 
\hline
\end{tabular}
}
\label{tab:ablation}
\end{table}




\subsubsection{Effectiveness of Deformable ConvLSTM} 
To investigate the proposed Deformable ConvLSTM (DConvLSTM) module, we take four different models for comparison: (b), (c), (d), and (e), where (c) adds a vanilla ConvLSTM into (b), (d) adds the proposed DConvLSTM, and (e) utilizes the DConvLSTM module in a bidirectional manner.  

From Table \ref{tab:ablation}, we can see that (c) outperforms (b) on Vid4 with slow motion while performs worse than (b) on Vimeo-Fast with fast motion videos. The results indicate that vanilla ConvLSTM can utilize global temporal contexts for slow motion videos, but cannot handle sequences with large motion. Furthermore, we observe that model (d) is significantly better than both (b) and (c), which demonstrates that our DConvLSTM can learn the temporal alignment between previous states and the current feature map. Therefore, it can better exploit global contexts for reconstructing visually appealing frames with more vivid details. Our findings are supported by qualitative results in Figure~\ref{fig:ablation_dconvlstm}. 

In addition, we verify the bidirectional mechanism in DConvLSTM by comparing (e) and (d) in Table \ref{tab:ablation} and Figure~\ref{fig:ablation_bidirectional}. From Table \ref{tab:ablation}, we observe that (e) with bidirectional DConvLSTM can further improve STVSR performance over (d) on both slow motion and fast motion testing sets. The visual results in Figure~\ref{fig:ablation_bidirectional} show that our full model with a bidirectional mechanism can restore more details by making better use of global temporal information for all input frames.


\subsubsection{Effectiveness of Deformable Feature Interpolation} To validate our deformable feature interpolation (DFI) module, we introduce two baselines for comparison: (a) and (b), where the model (a) only uses convolutions to blend LR features instead of deformable sampling functions as in model (b). Note that neither model (a) or (b) has ConvLSTM or DConvLSTM. 

From Table \ref{tab:ablation}, we observe that (b) outperforms (a) by 0.16dB on Vid4 and 0.73 dB on Vimeo-Fast dataset with fast motions in terms of PSNR. Figure~\ref{fig:ablation_interp} shows a qualitative comparison, where we can see that model (a) generates a face with severe motion blur, while the proposed deformable feature interpolation module can effectively address the large motion issue by exploiting local temporal contexts and help the model (b) generate more clear face structures and details. The superiority of the proposed DFI module demonstrates that the learned offsets in the deformable sampling functions can better exploit local temporal contexts and capture forward and backward motions in natural video sequences, even without any explicit supervision.

\subsubsection{Guided Feature Interpolation Critic} 
\begin{figure}[tb]
\captionsetup[subfigure]{labelformat=empty}
\begin{center}
  \begin{subfigure}[b]{0.32\linewidth}
     \includegraphics[width=\linewidth]{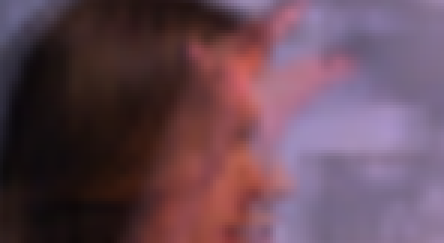}
     \subcaption{Overlayed LR}
  \end{subfigure}
  \begin{subfigure}[b]{0.32\linewidth}
  \includegraphics[width=\linewidth]{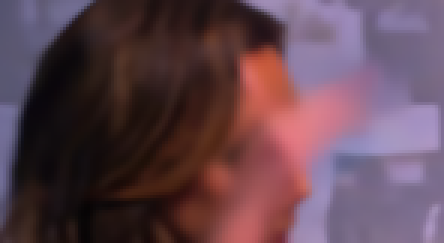}
  \subcaption{w/o guidance}
  \end{subfigure}
  \begin{subfigure}[b]{0.32\linewidth}
     \includegraphics[width=\linewidth]{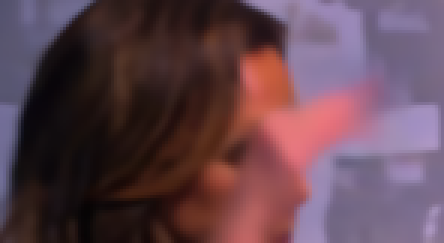}
     \subcaption{w/ guidance}
  \end{subfigure}
\end{center}
\vspace{-3mm}
   \caption{Ablation study on guided feature interpolation module. The additional guidance can help to strengthen the ability of the temporal feature interpolation network on handling motions. }
 \label{fig:ablation_ef}
\end{figure}

We first validate the strength of the guided feature interpolation (GFI) learning on STVSR task by comparing the model trained with it (model f) to the model without its supervision (model e) in Table \ref{tab:ablation}. Since the final performance is evaluated on the overall video sequence, optimizing the intermediate reward provided by LR frames can improve temporal consistency across frames. Visual results illustrated in Figure~\ref{fig:ablation_ef} can further demonstrate the superiority of the proposed GFI. 

Furthermore, we demonstrate the influence of the guided feature interpolation on noisy STVSR in Table \ref{tab:noisy_ablation}. We compare PSNR and SSIM of the Y channel of models trained with/without the guided loss on Vid4 dataset~\cite{liu2011bayesian}. For fair comparisons, both models are trained for the same number of iterations. We observe that the intermediate features guide actually decreases the performance in most cases when the input data is noisy.
\begin{table}
\caption{Ablation study of guided feature learning on noisy STVSR. We compare the PSNR and SSIM of the Y channel of models trained with/without the guided loss on Vid4 dataset~\cite{liu2011bayesian}.}
\centering
\resizebox{1.0\columnwidth}{!}{
\begin{tabular}{cc|cccc}
\hline
\multicolumn{2}{c|}{\multirow{2}{*}{Type}} & \multicolumn{2}{c}{w/o GFI} & \multicolumn{2}{c}{w/GFI}  \\
\multicolumn{2}{c|}{}                      & PSNR  & SSIM                & PSNR  & SSIM               \\ \hline
\hline 
\multicolumn{2}{c|}{Noise}                 & 23.91 & 0.6514              & 23.89 & 0.6510             \\
\hline
\multirow{4}{*}{Compression} & QF=10         & 22.03 & 0.5216              & 21.99 & 0.5204             \\
                             & QF=20         & 22.78 & 0.5696              & 22.78 & 0.5707             \\
                             & QF=30         & 23.25 & 0.6000              & 23.25 & 0.6013             \\
                             & QF=40         & 23.59 & 0.6226              & 23.55 & 0.6220            \\ \hline
\end{tabular}
}
\label{tab:noisy_ablation}
\vspace{-2mm}
\end{table}

\begin{figure}[t]
\captionsetup[subfigure]{labelformat=empty}
\begin{center}
  \begin{subfigure}[b]{0.32\linewidth}
     \includegraphics[width=\linewidth]{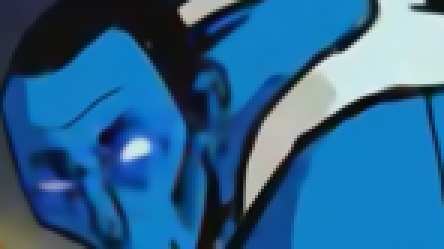}
     \subcaption{DAIN+EDVR: $t-1$}
  \end{subfigure}
  \begin{subfigure}[b]{0.32\linewidth}
  \includegraphics[width=\linewidth]{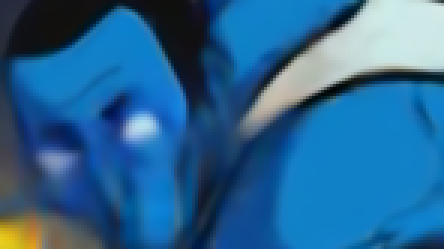}
  \subcaption{DAIN+EDVR: $t$}
  \end{subfigure}
  \begin{subfigure}[b]{0.32\linewidth}
     \includegraphics[width=\linewidth]{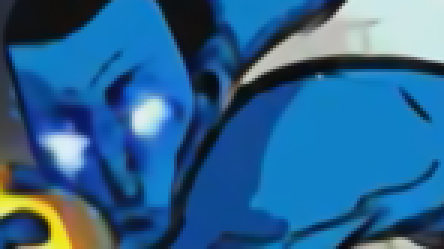}
     \subcaption{DAIN+EDVR: $t+1$}
  \end{subfigure}
  
  \begin{subfigure}[b]{0.32\linewidth}
     \includegraphics[width=\linewidth]{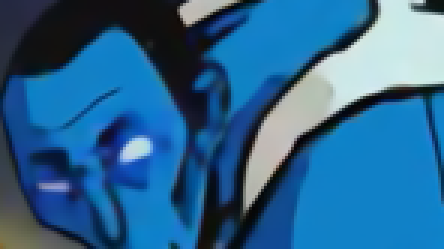}
      \subcaption{ZSM: $t-1$}
  \end{subfigure}
  \begin{subfigure}[b]{0.32\linewidth}
  \includegraphics[width=\linewidth]{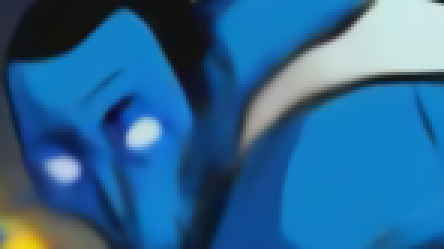}
\subcaption{ZSM: $t$}
  \end{subfigure}
  \begin{subfigure}[b]{0.32\linewidth}
     \includegraphics[width=\linewidth]{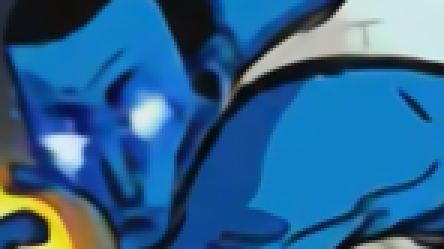}
    \subcaption{ZSM: $t+1$}
  \end{subfigure}
\end{center}
\vspace{-3mm}
  \caption{Temporal inconsistent issue in STVSR. It is more difficult to synthesis HR frame $t$ than HR frames: $t-1$ and $t+1$, since LR frames: $t-1$ and $t+1$ are available and LR frame $t$ is missing during testing. Synthesized HR frame at the time step $t$ is more blurry with fewer visual details than results at $t-1$ and $t+1$. }
 \label{fig:failure_incon}
\vspace{-2mm}
\end{figure}

\section{Discussion}
For space-time video super-resolution, the goal is to reconstruct HR frames for both missing intermediate and input LR frames. For example, given four input LR frames: $\{I_{1}^{L}, I_3^{L}, I_5^{L}, I_7^{L}\}$, we want to obtain the corresponding seven consecutive HR frames: $\{I_{1}^{H}, I_{2}^{H}, ..., I_{7}^{H}\}$. Since $I_{2}^{L}$, $I_{4}^{L}$, and $I_{6}^{L}$ are unavailable, and we need to hallucinate missing information for the frames. Therefore, it is more challenging to reconstruct $\{I_{2}^{H}, I_4^{H}, I_6^{H}\}$ than $\{I_{1}^{H}, I_3^{H}, I_5^{H}, I_7^{H}\}$. The imbalanced difficulty can lead to temporal inconsistent video results, and the potential video jittering becomes one of the most crucial issues that we need to consider when designing a space-time video super-resolution method. 
In our current framework, the proposed deformable ConvLSTM can implicitly enforce temporal coherence by handling visual motions and aggregating temporal contexts. However, results from our model still suffer from the temporal inconsistency issue due to the essential imbalanced difficulty (see Figure~\ref{fig:failure_incon}). To further alleviate the problem, we could consider devising new temporal consistency constraints to explicitly guide HR video frame reconstruction. 

\begin{figure}[t]
\captionsetup[subfigure]{labelformat=empty}
\begin{center}
  \begin{subfigure}[b]{0.32\linewidth}
     \includegraphics[width=\linewidth]{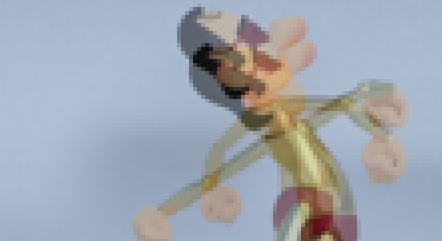}
     \subcaption{Overlayed LR}
  \end{subfigure}
  \begin{subfigure}[b]{0.32\linewidth}
  \includegraphics[width=\linewidth]{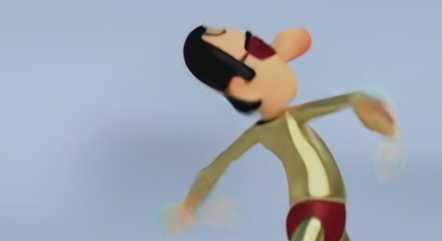}
  \subcaption{DAIN+EDVR}
  \end{subfigure}
  \begin{subfigure}[b]{0.32\linewidth}
     \includegraphics[width=\linewidth]{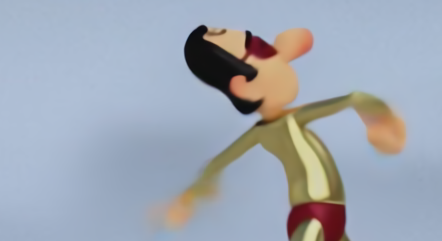}
     \subcaption{ZSM}
  \end{subfigure}
\end{center}
\vspace{-3mm}
  \caption{Failure example. Our model might fail to handle dynamic video objects with severe geometric deformations. }
 \label{fig:failure_deform}
 \vspace{-4mm}
\end{figure}

Videos might contain dramatically changing scenes or objects due to the existence of temporal motions and geometric deformations. Although our Zooming SlowMo is capable of handling large-motion videos, it might fail for certain dynamic objects with severe deformations. Multiple parts in a video object cross temporal frames might have different motion patterns, which leads to complex object deformations and introduces additional difficulties for space-time video super-resolution. We illustrate one failure example in Figure~\ref{fig:failure_deform}. Although the synthesized frame from our model contains fewer artifacts, the reconstructed \textit{hands} are pretty blurry due to large deformations in the animated character. To mitigate the issue, we desire a more deformation-robust temporal alignment approach. We believe it would be a promising direction.

\section{Conclusion}
In this paper, we propose a one-stage framework that directly reconstructs high-resolution and high frame rate videos without synthesizing intermediate low-resolution frames for space-time video super-resolution. To achieve this, we introduce a deformable feature interpolation network for feature-level temporal interpolation. Furthermore, we propose a deformable ConvLSTM for aggregating temporal information and handling motions. Our network can take advantage of the intra-relatedness between temporal interpolation and spatial super-resolution in the STVSR task with such a one-stage design. It enforces our model to adaptively learn to leverage useful local and global temporal contexts for alleviating large motion issues. 
Moreover, we introduce intermediate guidance to strengthen the temporal feature interpolation.
Extensive experimental results show that our one-stage framework is more effective yet efficient than existing two-stage methods; the proposed feature temporal interpolation network and deformable ConvLSTM are capable of handling very challenging fast motion videos; our network can successfully tackle more challenging noisy STVSR tasks.

\ifCLASSOPTIONcaptionsoff
  \newpage
\fi

{\small
\bibliographystyle{ieee_fullname}
\bibliography{egbib_journal}
}

\begin{IEEEbiography}[{\includegraphics[width=1in,height=1.25in,clip,keepaspectratio]{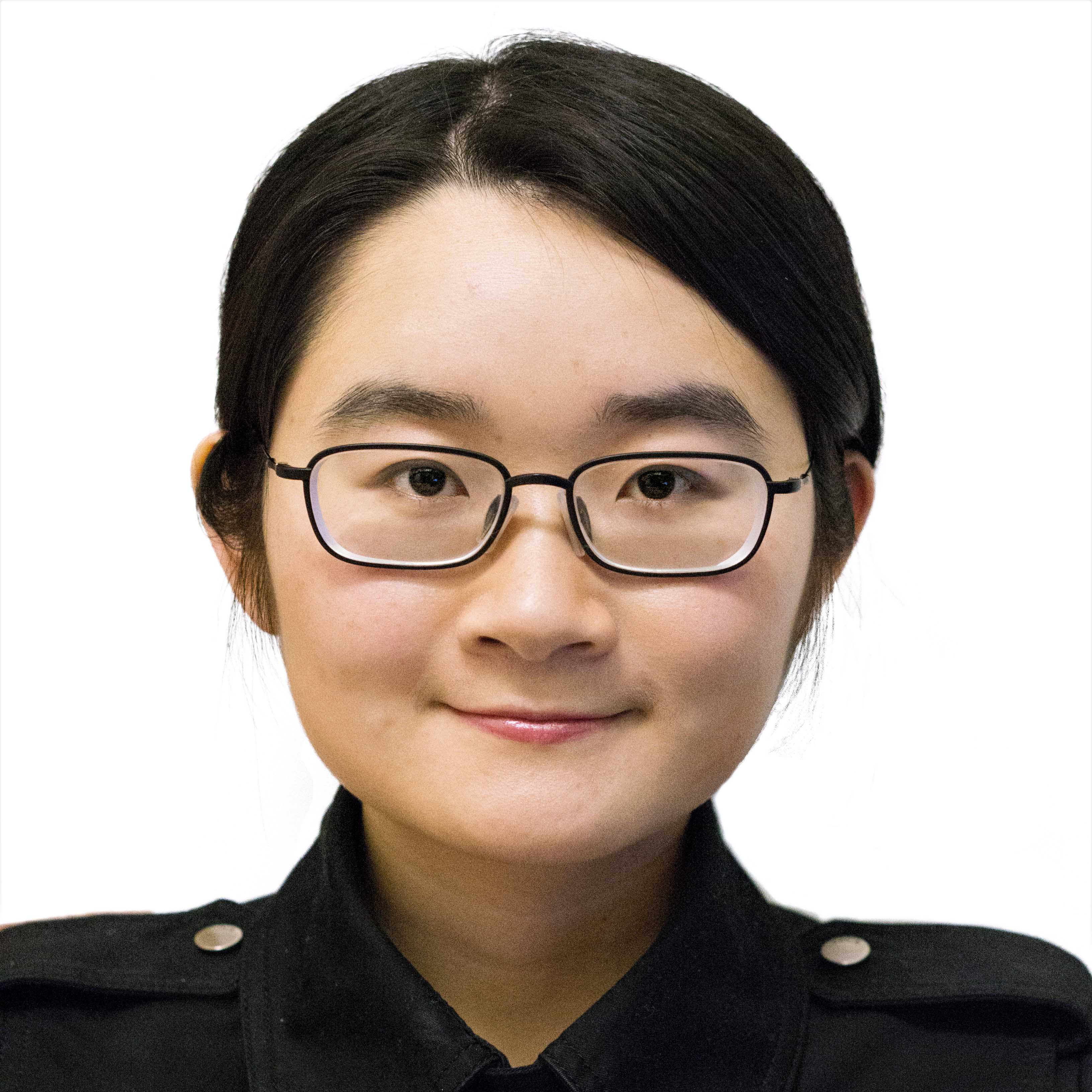}}]{Xiaoyu Xiang}
received the B.E. degree in engineering physics from Tsinghua University in 2015. She is currently pursuing a Ph.D. degree in the department of electrical and computer engineering at Purdue University. Her primary area of research has been image and video restoration.
\end{IEEEbiography}

\begin{IEEEbiography}[{\includegraphics[width=1in,height=1.25in,clip,keepaspectratio]{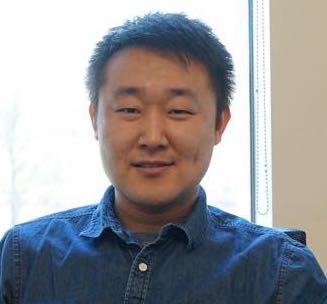}}]{Yapeng Tian}
received the B.E. degree in electronic engineering from Xidian University, Xian, China, in 2013, M.E. degree in electronic engineering at Tsinghua University, Beijing, China, in 2017, and is currently working toward the Ph.D. degree in the department of computer science at the University of Rochester, USA. His research interests include audio-visual scene understanding and low-level vision.
\end{IEEEbiography}

\begin{IEEEbiography}[{\includegraphics[width=1in,height=1.25in,clip,keepaspectratio]{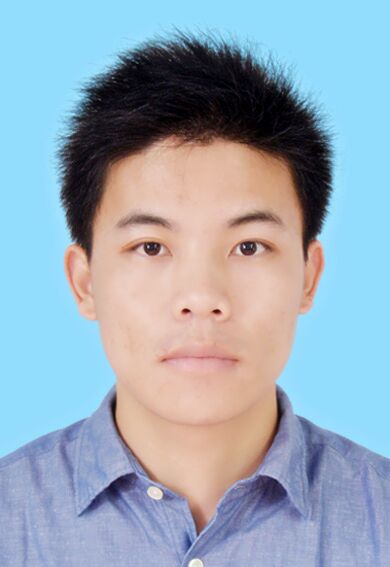}}]{Yulun Zhang} received B.E. degree from School of Electronic Engineering, Xidian University, China, in 2013 and M.E. degree from Department of Automation, Tsinghua University, China, in 2017. He is currently pursuing a Ph.D. degree with the Department of ECE, Northeastern University, USA. He was the receipt of the Best Student Paper Award at the IEEE International Conference on Visual Communication and Image Processing(VCIP) in 2015. He also won the Best Paper Award at IEEE International Conference on Computer Vision (ICCV) RLQ Workshop in 2019. His research interests include image restoration and deep learning.
\end{IEEEbiography}

\begin{IEEEbiography}[{\includegraphics[width=1in,height=1.25in,clip,keepaspectratio]{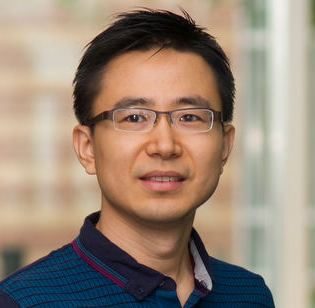}}]{Yun Fu (S'07-M'08-SM'11-F'19)} received the B.Eng. degree in information engineering and the M.Eng. degree in pattern recognition and intelligence systems from Xian Jiaotong University, China, respectively, and the M.S. degree in statistics and the Ph.D. degree in electrical and computer engineering from the University of Illinois at Urbana-Champaign, respectively. He is an interdisciplinary faculty member affiliated with College of Engineering and the Khoury College of Computer Sciences at Northeastern University since 2012. His research interests are Machine Learning, Computational Intelligence, Big Data Mining, Computer Vision, Pattern Recognition, and Cyber-Physical Systems. He has extensive publications in leading journals, books/book chapters and international conferences/workshops. He serves as associate editor, chairs, PC member and reviewer of many top journals and international conferences/workshops. He received seven Prestigious Young Investigator Awards from NAE, ONR, ARO, IEEE, INNS, UIUC, Grainger Foundation; eleven Best Paper Awards from IEEE, ACM, IAPR, SPIE, SIAM; many major Industrial Research Awards from Google, Samsung, and Adobe, etc. He is currently an Associate Editor of several journals, such as IEEE Transactions on Neural Networks and Leaning Systems (TNNLS), IEEE Transactions on Image Processing (TIP), and ACM Transactions on Knowledge Discovery from Data (TKDD). He is fellow of IEEE, IAPR, OSA, and SPIE, a Lifetime Distinguished Member of ACM, Lifetime Member of AAAI and Institute of Mathematical Statistics, member of ACM Future of Computing Academy, Global Young Academy, AAAS, INNS, and Beckman Graduate Fellow during 2007-2008.
\end{IEEEbiography}

\begin{IEEEbiography}[{\includegraphics[width=1in,height=1.29in,clip,keepaspectratio]{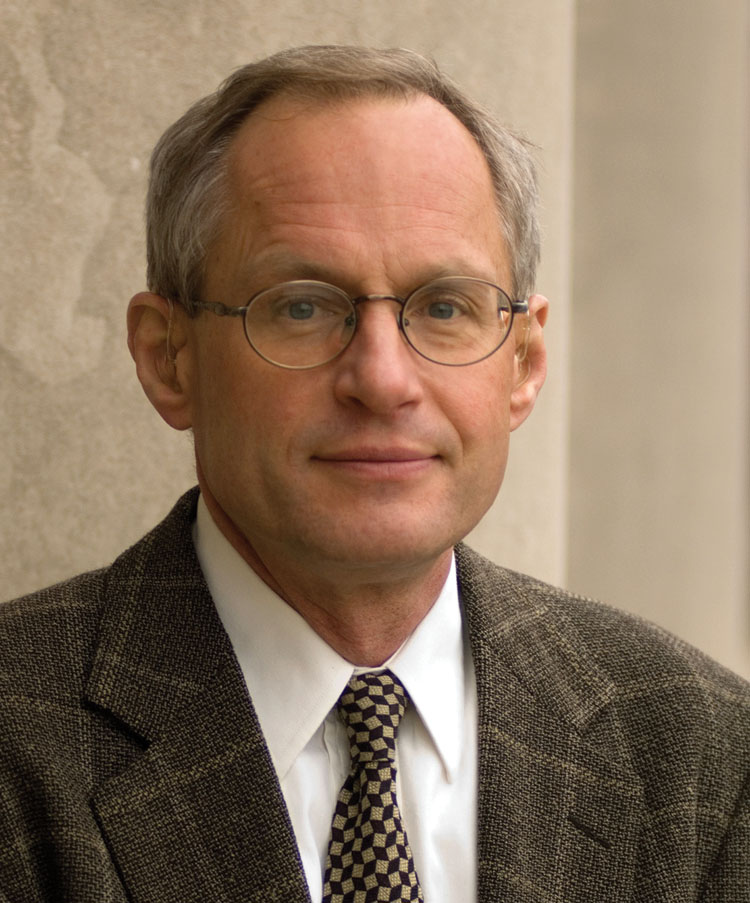}}]{Jan P. Allebach (LF'16)}
is currently a Hewlett-Packard Distinguished Professor of electrical and computer engineering at Purdue University. He is a fellow of the National Academy of Inventors, the Society for Imaging Science and Technology (IS\&T), and SPIE. He was named the Electronic Imaging Scientist of the Year by
IS\&T and SPIE, and was named the Honorary Member of IS\&T, the highest award that IS\&T bestows. He has received the IEEE Daniel E. Noble Award, the IS\&T/OSA Edwin Land Medal, and is a member of the National Academy of Engineering. He has served as an IEEE Signal Processing Society Distinguished Lecturer from 1994 to 1995, and again from 2016 to 2017.
\end{IEEEbiography}

\begin{IEEEbiography}[{\includegraphics[width=1in,height=1.25in,clip,keepaspectratio]{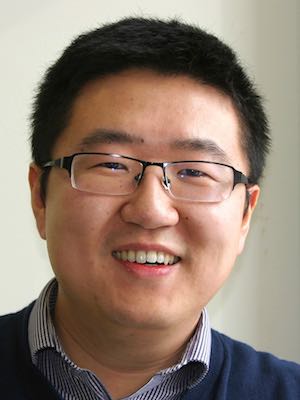}}]{Chenliang Xu}
is an Assistant Professor in the Department of Computer Science at the University of Rochester. He received his Ph.D. degree from the University of Michigan in 2016. His research interests range from solving core computer vision problems to pushing the limits of their understanding in the broader AI contexts, including multisensory perception, cognitive robotics, and data science. He was the recipient of multiple NSF awards, including BIGDATA and CDS\&E, the University of Rochester’s AR/VR Pilot Award, Tencent Rhino-Bird Award, a Best Paper Award at Sound and Music Computing Conference, a Best Paper Award at ACM SIGGRAPH VRCAI, and an Open Source Code Award in IEEE CVPR. He has authored more than 50 peer-reviewed papers in computer vision and artificial intelligence venues.
\end{IEEEbiography}







\end{document}